\documentclass[10pt]{article}
\usepackage[preprint]{tmlr}
\usepackage[T1]{fontenc}
\usepackage{lmodern}
\usepackage{amsmath,amssymb}
\usepackage{booktabs}
\usepackage{array}
\usepackage{longtable}
\usepackage{graphicx}
\usepackage{xcolor}
\usepackage{placeins}
\usepackage{needspace}
\usepackage{hyperref}
\hypersetup{hidelinks}
\usepackage{url}

\newcommand{\pubPatternMLP}{50.5}  %
\newcommand{\pubPatternGatedGCNPE}{86.5}  %
\newcommand{\pubPatternGraphGPS}{86.7}  %
\newcommand{\pubPatternGraphGPSFirstVersions}{90.3}  %
\newcommand{\pubPatternEGT}{86.8}  %
\newcommand{\pubPatternDGN}{86.7}  %
\newcommand{\pubPatternSAN}{86.6}  %
\newcommand{\pubPatternGCNAK}{86.9}  %
\newcommand{\pubPatternKSubtreeSAT}{86.9}  %
\newcommand{\pubPatternGRPE}{87.0}  %
\newcommand{\pubPatternGRIT}{87.2}  %
\newcommand{\pubPatternCKGConv}{88.7}  %
\newcommand{\pubPatternGradformer}{86.9}  %
\newcommand{\pubPatternGRASS}{89.2}  %
\newcommand{\pubPatternPPGT}{89.8}  %
\newcommand{\pubPatternDeepGraphPlainAccuracy}{90.7}  %
\newcommand{\pubPatternGradformerPlainAccuracy}{90.9}  %
\newcommand{\pubPatternGRITQPESRerun}{90.4}  %
\newcommand{\pubPatternGRITQiRWQPE}{90.2}  %
\newcommand{\pubClusterBest}{81.6}  %
\newcommand{\pubClusterBestSd}{0.19}  %
\newcommand{\pubClusterGRIT}{80.0}  %
\newcommand{\pubClusterGRASS}{79.5}  %
\newcommand{\pubClusterPPGT}{80.0}  %
\newcommand{\pubMinesBest}{99.55}  %
\newcommand{\pubMinesBestSd}{0.15}  %
\newcommand{\recTrainGraphs}{10,000}  %
\newcommand{\recTestGraphs}{2,000}  %
\newcommand{\recDistinctPatterns}{100}  %
\newcommand{\recGraphsPerPattern}{100}  %
\newcommand{\recTestRecur}{100.0}  %
\newcommand{\recValRecur}{100.0}  %
\newcommand{\recSizeMin}{6}  %
\newcommand{\recSizeMax}{34}  %
\newcommand{\recSizeMedian}{21}  %
\newcommand{\fastGraphs}{200}  %
\newcommand{\fastMatcherBatch}{96.7}  %
\newcommand{\fastMatcherPooled}{96.5}  %
\newcommand{\freshMatcherPooled}{60.1}  %
\newcommand{\freshMatcherLo}{59.8}  %
\newcommand{\freshMatcherHi}{60.5}  %
\newcommand{\freshLargestPooled}{53.7}  %
\newcommand{\freshLargestLo}{53.4}  %
\newcommand{\freshLargestHi}{54.1}  %
\newcommand{\freshDensityPooled}{86.4}  %
\newcommand{\freshDensityLo}{85.9}  %
\newcommand{\freshDensityHi}{86.8}  %
\newcommand{\firstBlindFreshPooled}{86.1}  %
\newcommand{\freshGraphsSmall}{135}  %
\newcommand{\freshMatcherExactPctSmall}{0.0}  %
\newcommand{\freshMatcherFailWrongPlaceSmall}{2}  %
\newcommand{\freshMatcherFailOtherPatternSmall}{121}  %
\newcommand{\freshMatcherFailLowOverlapSmall}{12}  %
\newcommand{\freshMatcherFailMissedSmall}{0}  %
\newcommand{\freshGraphsBig}{365}  %
\newcommand{\freshMatcherExactPctBig}{0.0}  %
\newcommand{\freshMatcherFailWrongPlaceBig}{0}  %
\newcommand{\freshMatcherFailOtherPatternBig}{365}  %
\newcommand{\freshMatcherFailLowOverlapBig}{0}  %
\newcommand{\freshMatcherFailMissedBig}{0}  %
\newcommand{\freshGraphs}{500}  %
\newcommand{\freshMatcherExactPct}{0.0}  %
\newcommand{\freshMatcherFailWrongPlace}{2}  %
\newcommand{\freshMatcherFailOtherPattern}{486}  %
\newcommand{\freshMatcherFailLowOverlap}{12}  %
\newcommand{\freshMatcherFailMissed}{0}  %
\newcommand{\freshLargestExactPctSmall}{0.0}  %
\newcommand{\freshLargestExactPctBig}{0.0}  %
\newcommand{\freshLargestExactPct}{0.0}  %
\newcommand{\realMatcherPooled}{99.2}  %
\newcommand{\realMatcherBatch}{99.0}  %
\newcommand{\realMatcherLo}{99.0}  %
\newcommand{\realMatcherHi}{99.3}  %
\newcommand{\realLargestPooled}{97.3}  %
\newcommand{\realDensityPooled}{86.8}  %
\newcommand{\realDensityLo}{86.5}  %
\newcommand{\realDensityHi}{87.0}  %
\newcommand{\firstBlindRealPooled}{86.4}  %
\newcommand{\realPPGTAboveDensity}{3.0}  %
\newcommand{\realCKGConvAboveDensity}{1.9}  %
\newcommand{\realGRITAboveDensity}{0.4}  %
\newcommand{\realGraphsSmall}{480}  %
\newcommand{\realMatcherExactPctSmall}{74.8}  %
\newcommand{\realMatcherFailWrongPlaceSmall}{30}  %
\newcommand{\realMatcherFailOtherPatternSmall}{73}  %
\newcommand{\realMatcherFailLowOverlapSmall}{18}  %
\newcommand{\realMatcherFailMissedSmall}{0}  %
\newcommand{\realGraphsBig}{1,520}  %
\newcommand{\realMatcherExactPctBig}{99.7}  %
\newcommand{\realMatcherFailWrongPlaceBig}{5}  %
\newcommand{\realMatcherFailOtherPatternBig}{0}  %
\newcommand{\realMatcherFailLowOverlapBig}{0}  %
\newcommand{\realMatcherFailMissedBig}{0}  %
\newcommand{\realGraphs}{2,000}  %
\newcommand{\realMatcherExactPct}{93.7}  %
\newcommand{\realMatcherFailWrongPlace}{35}  %
\newcommand{\realMatcherFailOtherPattern}{73}  %
\newcommand{\realMatcherFailLowOverlap}{18}  %
\newcommand{\realMatcherFailMissed}{0}  %
\newcommand{\realLargestExactPctSmall}{45.0}  %
\newcommand{\realLargestExactPctBig}{99.7}  %
\newcommand{\realLargestExactPct}{86.6}  %
\newcommand{\convMatcherPooled}{99.2}  %
\newcommand{\convMatcherBatch}{99.0}  %
\newcommand{\convMatcherPlain}{99.6}  %
\newcommand{\convMatcherPlainMinusPooled}{0.4}  %
\newcommand{\convMatcherPooledMinusBatch}{0.1}  %
\newcommand{\convLargestPooled}{97.3}  %
\newcommand{\convLargestBatch}{96.9}  %
\newcommand{\convLargestPlain}{98.4}  %
\newcommand{\convLargestPlainMinusPooled}{1.1}  %
\newcommand{\convLargestPooledMinusBatch}{0.3}  %
\newcommand{\firstBlindRealBatch}{86.4}  %
\newcommand{\convDensityPooled}{86.8}  %
\newcommand{\convDensityBatch}{86.7}  %
\newcommand{\convDensityPlain}{86.8}  %
\newcommand{\convDensityPlainMinusPooled}{0.0}  %
\newcommand{\convDensityPooledMinusBatch}{0.1}  %
\newcommand{\convPositiveRate}{17.6}  %
\newcommand{\appGPapers}{37}  %
\newcommand{\appGMethodPapers}{34}  %
\newcommand{\appGMetricNamed}{9}  %
\newcommand{\appGAboveGrit}{9}  %
\newcommand{\pubPatternGRITTwo}{87.20}  %
\newcommand{\clusterGraphs}{1,000}  %
\newcommand{\clusterFirstPooled}{82.1}  %
\newcommand{\clusterFirstBatch}{82.0}  %
\newcommand{\clusterFirstLo}{81.3}  %
\newcommand{\clusterFirstHi}{82.8}  %
\newcommand{\clusterFirstAgreementMedian}{0.96}  %
\newcommand{\clusterFirstGapInBestSds}{2.5}  %
\newcommand{\clusterCeilingPooled}{82.3}  %
\newcommand{\clusterCeilingLo}{81.5}  %
\newcommand{\clusterCeilingHi}{83.0}  %
\newcommand{\clusterChainAgreementMedian}{0.96}  %
\newcommand{\clusterChainAgreementMin}{0.50}  %
\newcommand{\minesFeaturesAuc}{98.77}  %
\newcommand{\minesFeaturesSd}{0.12}  %
\newcommand{\minesLabelsAuc}{99.92}  %
\newcommand{\minesLabelsSd}{0.06}  %
\newcommand{\minesBpIterations}{400}  %
\newcommand{\minesBpLastChange}{0.021}  %
\newcommand{\minesExactAuc}{99.86}  %
\newcommand{\minesExactSd}{0.05}  %
\newcommand{\minesExactSplitLo}{99.79}  %
\newcommand{\minesExactSplitHi}{99.93}  %
\newcommand{\minesExactChains}{8}  %
\newcommand{\minesExactValidChains}{4}  %
\newcommand{\minesExactChainLo}{99.11}  %
\newcommand{\minesExactChainHi}{99.58}  %
\newcommand{\minesExactRhat}{1.40}  %
\newcommand{\gpuOriginalPooled}{88.8}  %
\newcommand{\gpuOriginalLo}{88.6}  %
\newcommand{\gpuOriginalHi}{89.1}  %
\newcommand{\gpuFreshPooled}{84.5}  %
\newcommand{\gpuFreshLo}{84.0}  %
\newcommand{\gpuFreshHi}{85.0}  %
\newcommand{\gpuMinusDensity}{$-$1.9}  %
\newcommand{\gpuMinusDensityLo}{$-$2.2}  %
\newcommand{\gpuMinusDensityHi}{$-$1.6}  %
\newcommand{\gpuFamiliarPooled}{88.6}  %
\newcommand{\gpuFreshMinusFamiliar}{$-$4.2}  %
\newcommand{\gpuFreshMinusFamiliarLo}{$-$4.9}  %
\newcommand{\gpuFreshMinusFamiliarHi}{$-$3.5}  %
\newcommand{\gpuSeedBOriginalPooled}{88.3}  %
\newcommand{\gpuSeedBFreshPooled}{84.3}  %
\newcommand{\recTestGraphsPerPattern}{20}  %
\newcommand{\recValGraphsPerPattern}{20}  %
\newcommand{\genPatternPIn}{0.5}  %
\newcommand{\genPatternPOut}{0.35}  %
\newcommand{\genPatternPHost}{0.5}  %
\newcommand{\genPatternHostBlocks}{5}  %
\newcommand{\genPatternBlockMin}{5}  %
\newcommand{\genPatternBlockMax}{34}  %
\newcommand{\genPatternPInternal}{0.5}  %
\newcommand{\freshSizeMin}{6}  %
\newcommand{\freshSizeMax}{34}  %
\newcommand{\genClusterBlocks}{6}  %
\newcommand{\genClusterPIn}{0.55}  %
\newcommand{\genClusterPOut}{0.25}  %
\newcommand{\genClusterBlockMin}{5}  %
\newcommand{\genClusterBlockMax}{34}  %
\newcommand{\clusterFirstSweeps}{400}  %
\newcommand{\clusterFirstBurn}{100}  %
\newcommand{\clusterFirstChains}{2}  %
\newcommand{\minesPrior}{0.2}  %
\newcommand{\minesCap}{5}  %
\newcommand{\minesBoardSide}{100}  %
\newcommand{\minesRevealed}{5,000}  %
\newcommand{\minesRevealedPct}{50}  %
\newcommand{\minesBpDamping}{0.9}  %
\newcommand{\minesSplits}{10}  %
\newcommand{\matcherBig}{14}  %
\newcommand{\matcherTop}{60}  %
\newcommand{\matcherCap}{200,000}  %
\newcommand{\realTrainPosRate}{17.7}  %
\newcommand{\batchGraphs}{128}  %
\newcommand{\blindRestarts}{8}  %
\newcommand{\blindMeanfieldIters}{200}  %
\newcommand{\firstBlindSweeps}{250}  %
\newcommand{\firstBlindBurn}{50}  %
\newcommand{\matcherBootstrapReplicates}{1,000}  %
\newcommand{\realDensityHalfPooled}{81.1}  %
\newcommand{\freshDensityHalfPooled}{80.4}  %
\newcommand{\realLargestLo}{96.9}  %
\newcommand{\realLargestHi}{97.6}  %
\newcommand{\firstBlindRealMinusFresh}{0.3}  %
\newcommand{\freshMatcherMinusDensity}{$-$26.3}  %
\newcommand{\freshMatcherMinusDensityLo}{$-$26.8}  %
\newcommand{\freshMatcherMinusDensityHi}{$-$25.8}  %
\newcommand{\realMimicExpected}{3.8}  %
\newcommand{\realSmallestSize}{6}  %
\newcommand{\realSmallestSizeGraphs}{80}  %
\newcommand{\realSmallestSizeExact}{38}  %
\newcommand{\gpuFamiliarLo}{88.1}  %
\newcommand{\gpuFamiliarHi}{89.1}  %
\newcommand{\gpuFamiliarMinusOriginal}{$-$0.2}  %
\newcommand{\gpuFamiliarMinusOriginalLo}{$-$0.7}  %
\newcommand{\gpuFamiliarMinusOriginalHi}{0.4}  %
\newcommand{\gpuCheckpointEpoch}{159}  %
\newcommand{\gpuOriginalSmall}{86.4}  %
\newcommand{\gpuOriginalBig}{88.9}  %
\newcommand{\gpuFreshSmall}{84.2}  %
\newcommand{\gpuFreshBig}{84.1}  %
\newcommand{\gpuFamiliarSmall}{86.1}  %
\newcommand{\gpuFamiliarBig}{88.8}  %
\newcommand{\gpuTrainHours}{19.7}  %
\newcommand{\gpuSeedBOriginalLo}{88.0}  %
\newcommand{\gpuSeedBFreshLo}{83.8}  %
\newcommand{\gpuSeedBOriginalHi}{88.5}  %
\newcommand{\gpuSeedBFreshHi}{84.8}  %
\newcommand{\gpuSeedBFamiliarPooled}{87.8}  %
\newcommand{\gpuSeedBFreshMinusFamiliar}{$-$3.5}  %
\newcommand{\gpuSeedBFreshMinusFamiliarLo}{$-$4.2}  %
\newcommand{\gpuSeedBFreshMinusFamiliarHi}{$-$2.8}  %
\newcommand{\gpuSeedBMinusDensity}{$-$2.1}  %
\newcommand{\gpuSeedBMinusDensityLo}{$-$2.4}  %
\newcommand{\gpuSeedBMinusDensityHi}{$-$1.7}  %
\newcommand{\gpuSeedBFamiliarLo}{87.3}  %
\newcommand{\gpuSeedBFamiliarHi}{88.3}  %
\newcommand{\gpuSeedBFamiliarMinusOriginal}{$-$0.5}  %
\newcommand{\gpuSeedBFamiliarMinusOriginalLo}{$-$1.0}  %
\newcommand{\gpuSeedBFamiliarMinusOriginalHi}{0.1}  %
\newcommand{\gpuSeedBCheckpointEpoch}{184}  %
\newcommand{\gpuSeedBOriginalSmall}{85.8}  %
\newcommand{\gpuSeedBOriginalBig}{88.4}  %
\newcommand{\gpuSeedBFreshSmall}{84.5}  %
\newcommand{\gpuSeedBFreshBig}{83.8}  %
\newcommand{\gpuSeedBFamiliarSmall}{86.1}  %
\newcommand{\gpuSeedBFamiliarBig}{87.8}  %
\newcommand{\gpuSeedBTrainHours}{19.6}  %
\newcommand{\gpuFreshDensitySmall}{86.1}  %
\newcommand{\gpuFreshDensityBig}{85.8}  %
\newcommand{\gpuRoundtripGraphs}{200}  %
\newcommand{\gpuEpochs}{200}  %
\newcommand{\gpuFamiliarGraphs}{500}  %
\newcommand{\familiarPatternsUsed}{99}  %
\newcommand{\regCkgGate}{1.5}  %
\newcommand{\regCkgBandLo}{85.1}  %
\newcommand{\regCkgBandHi}{87.1}  %
\newcommand{\regCkgKill}{1.0}  %
\newcommand{\regCkgFamiliar}{1.0}  %
\newcommand{\regCkgRoundtrip}{0.1}  %
\newcommand{\regCkgReproduce}{1.0}  %
\newcommand{\minesExactWidth}{4}  %
\newcommand{\minesExactBurn}{200}  %
\newcommand{\minesExactHours}{10.5}  %
\newcommand{\minesExactRhatMines}{1.013}  %
\newcommand{\minesExactEssMines}{24,432}  %
\newcommand{\minesExactEssAuc}{3,280}  %
\newcommand{\minesExactChainRange}{0.48}  %
\newcommand{\minesExactHalfDiff}{0.005}  %
\newcommand{\minesExactNodeSd}{0.12}  %
\newcommand{\minesExactCritValid}{no}  %
\newcommand{\minesExactCritNodes}{no}  %
\newcommand{\minesExactRan}{4}  %
\newcommand{\minesExactCheckFull}{$2\times10^{-16}$}  %
\newcommand{\minesExactCheckStrip}{$6\times10^{-17}$}  %
\newcommand{\minesSmallBoardErr}{0.43}  %
\newcommand{\ckgFreshOracle}{84.5}  %
\newcommand{\ckgFreshAuc}{92.3}  %
\newcommand{\ckgSeedBFreshOracle}{84.4}  %
\newcommand{\ckgSeedBFreshAuc}{92.2}  %
\newcommand{\releasedSizesChiP}{0.61}  %
\newcommand{\releasedSizesMean}{21.0}  %
\newcommand{\ckgFamiliarAuc}{95.7}  %
\newcommand{\ckgSeedBFamiliarAuc}{95.1}  %
\newcommand{\releasedDensity}{0.496}  %
\newcommand{\releasedFeatureChiP}{0.67}  %
\newcommand{\releasedHostRate}{0.500}  %
\newcommand{\familiarMatcherPooled}{99.4}  %
\newcommand{\familiarMatcherLo}{99.0}  %
\newcommand{\familiarMatcherHi}{99.7}  %
\newcommand{\firstBlindFamiliarPooled}{86.3}  %
\newcommand{\familiarMatcherExactPct}{94.8}  %
\newcommand{\valMatcherPooled}{99.1}  %
\newcommand{\valMatcherLo}{98.9}  %
\newcommand{\valMatcherHi}{99.3}  %
\newcommand{\firstBlindValPooled}{86.4}  %
\newcommand{\valGraphs}{2,000}  %
\newcommand{\familiarDensityPooled}{86.5}  %
\newcommand{\familiarDensityLo}{86.1}  %
\newcommand{\familiarDensityHi}{87.0}  %
\newcommand{\familiarLargestPooled}{97.6}  %
\newcommand{\familiarLargestLo}{97.0}  %
\newcommand{\familiarLargestHi}{98.2}  %
\newcommand{\reusePatternSize}{20}  %
\newcommand{\reuseTrainGraph}{300}  %
\newcommand{\reuseTrainHost}{100}  %
\newcommand{\reuseTestGraph}{60}  %
\newcommand{\reuseTestHost}{96}  %
\newcommand{\regTwoAOne}{86.1}  %
\newcommand{\regTwoATwo}{1}  %
\newcommand{\regTwoBOneP}{0.05}  %
\newcommand{\regTwoBOneMeanWithin}{2}  %
\newcommand{\regTwoBOneMean}{20}  %
\newcommand{\regTwoBTwoWithin}{0.02}  %
\newcommand{\regTwoBTwoDensity}{0.5}  %
\newcommand{\regTwoBTwoP}{0.05}  %
\newcommand{\regTwoBTwoHostWithin}{0.02}  %
\newcommand{\regTwoBTwoHost}{0.5}  %
\newcommand{\regTwoCOneWithin}{1.0}  %
\newcommand{\regTwoCOneValue}{99.2}  %
\newcommand{\regTwoCTwoValue}{86.4}  %
\newcommand{\regTwoDHeld}{20}  %
\newcommand{\regTwoDOne}{3}  %
\newcommand{\regTwoDTwo}{1}  %
\newcommand{\regTwoDThree}{1}  %
\newcommand{\regTwoDKill}{1}  %
\newcommand{\heldH}{84.2}  %
\newcommand{\heldKept}{89.2}  %
\newcommand{\heldKeptMinusFull}{0.4}  %
\newcommand{\heldFullH}{88.8}  %
\newcommand{\heldKeptMinusFullLo}{0.2}  %
\newcommand{\heldKeptMinusFullHi}{0.5}  %
\newcommand{\heldHLo}{83.6}  %
\newcommand{\heldHHi}{84.7}  %
\newcommand{\heldKeptLo}{88.9}  %
\newcommand{\heldKeptHi}{89.5}  %
\newcommand{\heldSeedBH}{84.5}  %
\newcommand{\heldSeedBKept}{89.3}  %
\newcommand{\heldSeedBKeptMinusFull}{0.9}  %
\newcommand{\heldSeedBFullH}{87.8}  %
\newcommand{\heldSeedBKeptMinusFullLo}{0.7}  %
\newcommand{\heldSeedBKeptMinusFullHi}{1.1}  %
\newcommand{\heldSeedBHLo}{83.9}  %
\newcommand{\heldSeedBHHi}{85.0}  %
\newcommand{\heldSeedBKeptLo}{89.0}  %
\newcommand{\heldSeedBKeptHi}{89.6}  %
\newcommand{\heldBlindH}{86.9}  %
\newcommand{\heldBlindHLo}{86.4}  %
\newcommand{\heldBlindHHi}{87.4}  %
\newcommand{\heldHMinusBlind}{$-$2.8}  %
\newcommand{\heldHMinusBlindLo}{$-$3.1}  %
\newcommand{\heldHMinusBlindHi}{$-$2.4}  %
\newcommand{\heldBelowBlind}{2.8}  %
\newcommand{\firstBlindHeldBelow}{2.4}  %
\newcommand{\heldBigH}{83.8}  %
\newcommand{\heldBigKept}{89.5}  %
\newcommand{\heldSmallH}{84.5}  %
\newcommand{\heldSmallKept}{85.8}  %
\newcommand{\heldSeedBHMinusBlind}{$-$2.5}  %
\newcommand{\heldSeedBHMinusBlindLo}{$-$2.8}  %
\newcommand{\heldSeedBHMinusBlindHi}{$-$2.1}  %
\newcommand{\heldSeedBBelowBlind}{2.5}  %
\newcommand{\firstBlindHeldSeedBBelow}{2.1}  %
\newcommand{\heldSeedBBigH}{84.0}  %
\newcommand{\heldSeedBBigKept}{89.6}  %
\newcommand{\heldSeedBSmallH}{85.6}  %
\newcommand{\heldSeedBSmallKept}{85.8}  %
\newcommand{\heldGraphs}{400}  %
\newcommand{\keptGraphs}{1,600}  %
\newcommand{\heldSmallGraphs}{100}  %
\newcommand{\heldSmallPool}{24}  %
\newcommand{\heldBigPool}{76}  %
\newcommand{\heldTrainGraphs}{8,000}  %
\newcommand{\heldValGraphs}{1,600}  %
\newcommand{\heldPatterns}{20}  %
\newcommand{\regPilotPatterns}{100}  %
\newcommand{\regPilotPatternsLo}{90}  %
\newcommand{\regPilotPatternsHi}{110}  %
\newcommand{\regPilotPerPattern}{100}  %
\newcommand{\regPilotRecur}{95}  %
\newcommand{\regPilotMatcherLo}{97}  %
\newcommand{\regPilotMatcherHi}{100}  %
\newcommand{\regPilotKillRecur}{90}  %
\newcommand{\regPilotKillMatcher}{90}  %
\newcommand{\regMatcherPoint}{98.8}  %
\newcommand{\regMatcherLo}{98.0}  %
\newcommand{\regMatcherHi}{99.6}  %
\newcommand{\regLargestLo}{96.0}  %
\newcommand{\regLargestHi}{97.5}  %
\newcommand{\regExactAll}{93}  %
\newcommand{\regExactLargest}{84}  %
\newcommand{\regExactSmall}{75}  %
\newcommand{\regDensityPoint}{86.5}  %
\newcommand{\regDensityLo}{85.0}  %
\newcommand{\regDensityHi}{88.2}  %
\newcommand{\regClusterPoint}{85.0}  %
\newcommand{\regClusterLo}{82.0}  %
\newcommand{\regClusterHi}{89.0}  %
\newcommand{\regClusterAgreement}{0.95}  %
\newcommand{\regMinesFeaturesLo}{98.3}  %
\newcommand{\regMinesFeaturesHi}{98.7}  %
\newcommand{\regMinesLabelsLo}{99.0}  %
\newcommand{\regMinesLabelsHi}{99.6}  %
\newcommand{\regClusterSaturationWord}{one}  %
\newcommand{\regKillMatcher}{97.5}  %
\newcommand{\regKillDensity}{86.2}  %
\newcommand{\regFreshMatcherLo}{55}  %
\newcommand{\regFreshMatcherHi}{80}  %
\newcommand{\regFreshLargestLo}{50}  %
\newcommand{\regFreshLargestHi}{60}  %
\newcommand{\regFreshDensityWithin}{2}  %
\newcommand{\regFreshKill}{2}  %
\newcommand{\regMinesStripLo}{99.45}  %
\newcommand{\regMinesStripHi}{99.90}  %
\newcommand{\regMinesStripPoint}{99.70}  %
\newcommand{\regCritValid}{6}  %
\newcommand{\regCritRhat}{1.01}  %
\newcommand{\regCritEss}{400}  %
\newcommand{\regCritChains}{0.05}  %
\newcommand{\regCritHalves}{0.03}  %
\newcommand{\regCritNodes}{0.02}  %
\newcommand{\appGRowA}{90.17}  %
\newcommand{\appGRowB}{89.75}  %
\newcommand{\appGRowC}{89.18}  %
\newcommand{\appGRowD}{88.66}  %
\newcommand{\appGRowE}{87.93}  %
\newcommand{\appGRowF}{87.55}  %
\newcommand{\appGRowG}{87.51}  %
\newcommand{\appGRowH}{87.36}  %
\newcommand{\appGRowI}{87.24}  %
\newcommand{\appGRowJ}{87.22}  %
\newcommand{\appGRowK}{87.18}  %
\newcommand{\appGRowL}{87.14}  %
\newcommand{\appGRowM}{87.03}  %
\newcommand{\appGRowN}{86.99}  %
\newcommand{\appGRowO}{86.98}  %
\newcommand{\appGRowP}{86.89}  %
\newcommand{\appGRowQ}{86.87}  %
\newcommand{\appGRowR}{86.87}  %
\newcommand{\appGRowS}{86.86}  %
\newcommand{\appGRowT}{86.84}  %
\newcommand{\appGRowU}{86.78}  %
\newcommand{\appGRowV}{86.77}  %
\newcommand{\appGRowW}{86.77}  %
\newcommand{\appGRowX}{86.76}  %
\newcommand{\appGRowY}{86.74}  %
\newcommand{\appGRowZ}{86.74}  %
\newcommand{\appGRowBa}{86.71}  %
\newcommand{\appGRowBb}{86.69}  %
\newcommand{\appGRowBc}{86.68}  %
\newcommand{\appGRowBd}{86.68}  %
\newcommand{\appGRowBe}{86.64}  %
\newcommand{\appGRowBf}{86.63}  %
\newcommand{\appGRowBg}{86.38}  %
\newcommand{\appGRowBh}{85.79}  %
\newcommand{\appGRowBi}{85.75}  %
\newcommand{\appGRowBj}{85.39}  %
\newcommand{\appGRowBk}{81.62}  %
\newcommand{\heldHMinusFull}{$-$4.6}  %
\newcommand{\heldHMinusFullLo}{$-$5.1}  %
\newcommand{\heldHMinusFullHi}{$-$4.1}  %
\newcommand{\heldSeedBHMinusFull}{$-$3.3}  %
\newcommand{\heldSeedBHMinusFullLo}{$-$3.7}  %
\newcommand{\heldSeedBHMinusFullHi}{$-$2.9}  %
\newcommand{\heldBlindKept}{86.7}  %
\newcommand{\heldBlindKeptLo}{86.4}  %
\newcommand{\heldBlindKeptHi}{87.0}  %
\newcommand{\realDensitySmall}{85.7}  %
\newcommand{\realDensityBig}{86.5}  %
\newcommand{\familiarDensitySmall}{86.0}  %
\newcommand{\familiarDensityBig}{86.2}  %
\newcommand{\clusterFirstGapToBestTwo}{0.48}  %
\newcommand{\genPatternSizeMin}{5}  %
\newcommand{\genPatternSizeMax}{34}  %
\newcommand{\chanceSameBlock}{0.22}  %
\newcommand{\chanceEdgeProb}{0.38}  %
\newcommand{\chanceHostMean}{97.5}  %
\newcommand{\chanceKSmall}{10}  %
\newcommand{\chanceCopiesSmall}{7}  %
\newcommand{\chanceCopiesBig}{$2\times10^{-8}$}  %
\newcommand{\gpuTrainHoursRound}{20}  %
\newcommand{\gpuParamsRound}{440,000}  %
\newcommand{\minesExactSweepsMinRound}{10,000}  %
\newcommand{\minesExactSweepsMaxRound}{11,000}  %
\newcommand{\minesExactAnnealMaxRound}{1,300}  %
\newcommand{\minesExactRanWord}{four}  %
\newcommand{\minesExactCritFailedWord}{four}  %
\newcommand{\regCritCode}{$10^{-9}$}  %
\newcommand{\minesExactMetCheckFull}{yes}  %
\newcommand{\minesExactMetCheckStrip}{yes}  %
\newcommand{\minesExactMetRhatMines}{no}  %
\newcommand{\minesExactMetRhatAuc}{no}  %
\newcommand{\minesExactMetEssMines}{yes}  %
\newcommand{\minesExactMetEssAuc}{yes}  %
\newcommand{\minesExactMetChainRange}{no}  %
\newcommand{\minesExactMetHalfDiff}{yes}  %
\newcommand{\regDensityClauseWithin}{one}  %
\newcommand{\regDensityClauseAnchor}{87.2}  %
\newcommand{\regDensityClauseLo}{86.2}  %
\newcommand{\regDensityClauseHi}{88.2}  %
\newcommand{\blindSweeps}{1,000}  %
\newcommand{\blindBurn}{200}  %
\newcommand{\blindSizeMin}{5}  %
\newcommand{\blindSizeMax}{34}  %
\newcommand{\blindOutOfBoundsPct}{56}  %
\newcommand{\blindEnterMax}{10}  %
\newcommand{\blindAucFresh}{94.0}  %
\newcommand{\firstBlindAucFresh}{93.6}  %
\newcommand{\blindMoveOfficial}{0.3}  %
\newcommand{\blindMoveOfficialLo}{0.2}  %
\newcommand{\blindMoveOfficialHi}{0.4}  %
\newcommand{\blindMoveFresh}{0.3}  %
\newcommand{\blindMoveFreshLo}{0.1}  %
\newcommand{\blindMoveFreshHi}{0.5}  %
\newcommand{\blindMoveFamiliar}{0.2}  %
\newcommand{\blindMoveFamiliarLo}{0.0}  %
\newcommand{\blindMoveFamiliarHi}{0.4}  %
\newcommand{\blindMoveOfficialThree}{0.311}  %
\newcommand{\blindMoveFreshThree}{0.304}  %
\newcommand{\blindTrueStartPooled}{86.5}  %
\newcommand{\blindTrueStartMinusFirst}{0.01}  %
\newcommand{\blindTrueStartMinusFirstLo}{$-$0.05}  %
\newcommand{\blindTrueStartMinusFirstHi}{0.07}  %
\newcommand{\clusterFactorDiff}{$-$0.02}  %
\newcommand{\clusterFactorDiffLo}{$-$0.21}  %
\newcommand{\clusterFactorDiffHi}{0.16}  %
\newcommand{\clusterLongerDiff}{0.22}  %
\newcommand{\clusterLongerDiffLo}{0.09}  %
\newcommand{\clusterLongerDiffHi}{0.37}  %
\newcommand{\clusterFactorChangedPct}{3.5}  %
\newcommand{\clusterChains}{8}  %
\newcommand{\clusterSweeps}{800}  %
\newcommand{\clusterBurn}{200}  %
\newcommand{\regAOneLo}{0.0}  %
\newcommand{\regAOneHi}{0.5}  %
\newcommand{\regAThreeBest}{81.6}  %
\newcommand{\regBOneWithin}{0.3}  %
\newcommand{\regBOneAnchorReal}{86.4}  %
\newcommand{\regBOneAnchorFresh}{86.1}  %
\newcommand{\regBThreeWithin}{0.5}  %
\newcommand{\regBThreeAnchor}{86.4}  %
\newcommand{\regBFourSeedA}{92.3}  %
\newcommand{\regBFourSeedB}{92.2}  %
\newcommand{\blindPlainHalfReal}{91.0}  %
\newcommand{\blindPlainHalfRealLo}{90.8}  %
\newcommand{\blindPlainHalfRealHi}{91.2}  %
\newcommand{\blindTrueStartBOneDiff}{0.01}  %
\newcommand{\blindTrueStartBOneDiffLo}{$-$0.06}  %
\newcommand{\blindTrueStartBOneDiffHi}{0.04}  %
\newcommand{\blindChainGraphs}{400}  %
\newcommand{\blindChainsWord}{four}  %
\newcommand{\blindChainAgreePct}{98.3}  %
\newcommand{\blindFourChainDiff}{0.02}  %
\newcommand{\blindFourChainDiffLo}{$-$0.13}  %
\newcommand{\blindFourChainDiffHi}{0.09}  %
\newcommand{\recTestDistinctPatterns}{100}  %
\newcommand{\pubPatternGatedGCNPETwo}{86.51}  %
\newcommand{\heldHMinusKept}{$-$5.0}  %
\newcommand{\heldHMinusKeptLo}{$-$5.7}  %
\newcommand{\heldHMinusKeptHi}{$-$4.4}  %
\newcommand{\heldSeedBHMinusKept}{$-$4.8}  %
\newcommand{\heldSeedBHMinusKeptLo}{$-$5.5}  %
\newcommand{\heldSeedBHMinusKeptHi}{$-$4.2}  %
\newcommand{\clusterCeilingPooledTwo}{82.27}  %
\newcommand{\clusterCeilingBatchTwo}{82.23}  %
\newcommand{\clusterPlainTwo}{82.27}  %
\newcommand{\blindSweepsRatioWord}{four}  %
\newcommand{\hostSizeMax}{170}  %
\newcommand{\realAboveHostGraphs}{14}  %
\newcommand{\freshAboveHostCands}{6,039}  %
\newcommand{\matcherOfficialProcs}{4}  %
\newcommand{\minesAnnealStart}{0.3}  %
\newcommand{\minesAnnealGrowth}{1.06}  %
\newcommand{\minesAnnealCap}{50}  %
\newcommand{\minesAnnealStuck}{60}  %
\newcommand{\minesAnnealReset}{1}  %
\newcommand{\pubMinesPeerBest}{99.07}  %
\newcommand{\pubMinesPeerBestSd}{0.25}  %
\newcommand{\pubMinesRegisteredBest}{99.03}  %
\newcommand{\hvCompletedGritSeedABestEpoch}{88}  %
\newcommand{\hvCompletedGritSeedAEpochs}{100}  %
\newcommand{\hvCompletedGritSeedAOfficialAccuracy}{87.20}  %
\newcommand{\hvCompletedGritSeedAFreshAccuracy}{85.80}  %
\newcommand{\hvCompletedGritSeedAFreshMinusBlind}{$-$0.60}  %
\newcommand{\hvCompletedGritSeedAFreshMinusBlindInterval}{($-$0.80 to $-$0.39)}  %
\newcommand{\hvCompletedGritSeedAFamiliarAccuracy}{87.00}  %
\newcommand{\hvCompletedGritSeedAOfficialMinusFresh}{1.40}  %
\newcommand{\hvCompletedGritSeedAFamiliarMinusFresh}{1.20}  %
\newcommand{\hvCompletedGritSeedAFamiliarMinusFreshAuc}{0.98}  %
\newcommand{\hvCompletedCkgSeedABestEpoch}{61}  %
\newcommand{\hvCompletedCkgSeedAEpochs}{200}  %
\newcommand{\hvCompletedCkgSeedAOfficialAccuracy}{86.77}  %
\newcommand{\hvCompletedCkgSeedAFullRecipeOfficial}{88.81}  %
\newcommand{\hvCompletedCkgSeedAFreshAccuracy}{86.45}  %
\newcommand{\hvCompletedCkgSeedAFullRecipeFresh}{84.48}  %
\newcommand{\hvCompletedCkgSeedAFamiliarAccuracy}{86.55}  %
\newcommand{\hvCompletedCkgSeedAFullRecipeFamiliar}{88.64}  %
\newcommand{\hvCompletedCkgSeedAFreshImprovement}{1.97}  %
\newcommand{\hvCompletedCkgSeedAFreshImprovementInterval}{(1.66 to 2.27)}  %
\newcommand{\hvCompletedCkgSeedAFreshMinusBlind}{0.04}  %
\newcommand{\hvCompletedCkgSeedAFreshMinusBlindInterval}{($-$0.10 to 0.17)}  %
\newcommand{\hvCompletedGritSeedBBestEpoch}{95}  %
\newcommand{\hvCompletedGritSeedBOfficialAccuracy}{87.20}  %
\newcommand{\hvCompletedGritSeedBFreshAccuracy}{85.67}  %
\newcommand{\hvCompletedGritSeedBFreshMinusBlind}{$-$0.73}  %
\newcommand{\hvCompletedGritSeedBFreshMinusBlindInterval}{($-$0.94 to $-$0.52)}  %
\newcommand{\hvCompletedGritSeedBFamiliarAccuracy}{87.04}  %
\newcommand{\hvCompletedGritSeedBOfficialMinusFresh}{1.53}  %
\newcommand{\hvCompletedGritSeedBFamiliarMinusFresh}{1.37}  %
\newcommand{\hvCompletedGritSeedBFamiliarMinusFreshAuc}{1.04}  %
\newcommand{\hvCompletedCkgSeedBBestEpoch}{38}  %
\newcommand{\hvCompletedCkgSeedBEpochs}{200}  %
\newcommand{\hvCompletedCkgSeedBOfficialAccuracy}{86.77}  %
\newcommand{\hvCompletedCkgSeedBFullRecipeOfficial}{88.28}  %
\newcommand{\hvCompletedCkgSeedBFreshAccuracy}{86.46}  %
\newcommand{\hvCompletedCkgSeedBFullRecipeFresh}{84.34}  %
\newcommand{\hvCompletedCkgSeedBFamiliarAccuracy}{86.62}  %
\newcommand{\hvCompletedCkgSeedBFullRecipeFamiliar}{87.82}  %
\newcommand{\hvCompletedCkgSeedBFreshImprovement}{2.12}  %
\newcommand{\hvCompletedCkgSeedBFreshImprovementInterval}{(1.80 to 2.42)}  %
\newcommand{\hvCompletedCkgSeedBFreshMinusBlind}{0.05}  %
\newcommand{\hvCompletedCkgSeedBFreshMinusBlindInterval}{($-$0.09 to 0.22)}  %
\newcommand{\hvProtocolGritAccumulation}{2}  %
\newcommand{\hvProtocolGritMicroBatch}{16}  %
\newcommand{\hvProtocolGritWarmup}{5}  %
\newcommand{\hvProtocolGritReproductionTolerance}{1.50}  %
\newcommand{\hvProtocolGritFamiliarTolerance}{1.00}  %
\newcommand{\hvProtocolGritAccuracyGap}{1.00}  %
\newcommand{\hvProtocolGritAucGap}{1.00}  %
\newcommand{\hvProtocolCkgProximityLower}{$-$1.50}  %
\newcommand{\hvProtocolCkgProximityUpper}{0.50}  %
\newcommand{\hvProtocolCkgOfficialTolerance}{1.00}  %
\newcommand{\hvProtocolCkgLargeDeficit}{1.50}  %
\newcommand{\hvClassGritSeedASensitivity}{$-$0.57}  %
\newcommand{\hvClassGritSeedASensitivityInterval}{($-$0.97 to $-$0.14)}  %
\newcommand{\hvClassGritSeedASpecificity}{$-$0.64}  %
\newcommand{\hvClassGritSeedASpecificityInterval}{($-$0.86 to $-$0.41)}  %
\newcommand{\hvClassGritSeedBSensitivity}{$-$0.85}  %
\newcommand{\hvClassGritSeedBSensitivityInterval}{($-$1.25 to $-$0.42)}  %
\newcommand{\hvClassGritSeedBSpecificity}{$-$0.62}  %
\newcommand{\hvClassGritSeedBSpecificityInterval}{($-$0.85 to $-$0.38)}  %
\newcommand{\hvClassCkgSeedASensitivity}{1.13}  %
\newcommand{\hvClassCkgSeedASensitivityInterval}{(0.87 to 1.39)}  %
\newcommand{\hvClassCkgSeedASpecificity}{$-$1.05}  %
\newcommand{\hvClassCkgSeedASpecificityInterval}{($-$1.19 to $-$0.90)}  %
\newcommand{\hvClassCkgSeedBSensitivity}{1.87}  %
\newcommand{\hvClassCkgSeedBSensitivityInterval}{(1.60 to 2.17)}  %
\newcommand{\hvClassCkgSeedBSpecificity}{$-$1.76}  %
\newcommand{\hvClassCkgSeedBSpecificityInterval}{($-$1.91 to $-$1.62)}  %
\newcommand{\hvProtocolFreshTrainGraphsGrouped}{10,000}  %
\newcommand{\hvProtocolFreshValGraphsGrouped}{2,000}  %
\newcommand{\hvProtocolBootstrapReplicatesGrouped}{2,000}  %
\newcommand{\freshTrainGainRound}{2}  %
\newcommand{\ladderTopDegreeRealPooled}{79.8}  %
\newcommand{\ladderTopDegreeRealLo}{79.5}  %
\newcommand{\ladderTopDegreeRealHi}{80.1}  %
\newcommand{\ladderTopDegreeFamiliarPooled}{79.8}  %
\newcommand{\ladderTopDegreeFamiliarLo}{79.2}  %
\newcommand{\ladderTopDegreeFamiliarHi}{80.5}  %
\newcommand{\ladderTopDegreeFreshPooled}{79.7}  %
\newcommand{\ladderTopDegreeFreshLo}{79.0}  %
\newcommand{\ladderTopDegreeFreshHi}{80.3}  %
\newcommand{\ladderLogisticDegreeRealPooled}{85.2}  %
\newcommand{\ladderLogisticDegreeRealLo}{85.0}  %
\newcommand{\ladderLogisticDegreeRealHi}{85.4}  %
\newcommand{\ladderLogisticDegreeFamiliarPooled}{85.1}  %
\newcommand{\ladderLogisticDegreeFamiliarLo}{84.7}  %
\newcommand{\ladderLogisticDegreeFamiliarHi}{85.6}  %
\newcommand{\ladderLogisticDegreeFreshPooled}{85.0}  %
\newcommand{\ladderLogisticDegreeFreshLo}{84.5}  %
\newcommand{\ladderLogisticDegreeFreshHi}{85.5}  %
\newcommand{\ladderLogisticDegreeRankRealPooled}{85.2}  %
\newcommand{\ladderLogisticDegreeRankRealLo}{85.0}  %
\newcommand{\ladderLogisticDegreeRankRealHi}{85.5}  %
\newcommand{\ladderLogisticDegreeRankFamiliarPooled}{85.2}  %
\newcommand{\ladderLogisticDegreeRankFamiliarLo}{84.8}  %
\newcommand{\ladderLogisticDegreeRankFamiliarHi}{85.7}  %
\newcommand{\ladderLogisticDegreeRankFreshPooled}{85.0}  %
\newcommand{\ladderLogisticDegreeRankFreshLo}{84.6}  %
\newcommand{\ladderLogisticDegreeRankFreshHi}{85.5}  %
\newcommand{\ladderLogisticOneHopRealPooled}{86.4}  %
\newcommand{\ladderLogisticOneHopRealLo}{86.2}  %
\newcommand{\ladderLogisticOneHopRealHi}{86.7}  %
\newcommand{\ladderLogisticOneHopFamiliarPooled}{86.2}  %
\newcommand{\ladderLogisticOneHopFamiliarLo}{85.8}  %
\newcommand{\ladderLogisticOneHopFamiliarHi}{86.7}  %
\newcommand{\ladderLogisticOneHopFreshPooled}{86.2}  %
\newcommand{\ladderLogisticOneHopFreshLo}{85.8}  %
\newcommand{\ladderLogisticOneHopFreshHi}{86.7}  %
\newcommand{\ladderLogisticTwoHopRealPooled}{86.6}  %
\newcommand{\ladderLogisticTwoHopRealLo}{86.3}  %
\newcommand{\ladderLogisticTwoHopRealHi}{86.8}  %
\newcommand{\ladderLogisticTwoHopFamiliarPooled}{86.3}  %
\newcommand{\ladderLogisticTwoHopFamiliarLo}{85.9}  %
\newcommand{\ladderLogisticTwoHopFamiliarHi}{86.8}  %
\newcommand{\ladderLogisticTwoHopFreshPooled}{86.3}  %
\newcommand{\ladderLogisticTwoHopFreshLo}{85.8}  %
\newcommand{\ladderLogisticTwoHopFreshHi}{86.7}  %
\newcommand{\ladderLogisticTwoHopFeaturesRealPooled}{86.6}  %
\newcommand{\ladderLogisticTwoHopFeaturesRealLo}{86.4}  %
\newcommand{\ladderLogisticTwoHopFeaturesRealHi}{86.8}  %
\newcommand{\ladderLogisticTwoHopFeaturesFamiliarPooled}{86.4}  %
\newcommand{\ladderLogisticTwoHopFeaturesFamiliarLo}{85.9}  %
\newcommand{\ladderLogisticTwoHopFeaturesFamiliarHi}{86.8}  %
\newcommand{\ladderLogisticTwoHopFeaturesFreshPooled}{86.4}  %
\newcommand{\ladderLogisticTwoHopFeaturesFreshLo}{85.9}  %
\newcommand{\ladderLogisticTwoHopFeaturesFreshHi}{86.8}  %
\newcommand{\ladderBoostedDegreeRealPooled}{85.2}  %
\newcommand{\ladderBoostedDegreeRealLo}{84.9}  %
\newcommand{\ladderBoostedDegreeRealHi}{85.4}  %
\newcommand{\ladderBoostedDegreeFamiliarPooled}{85.1}  %
\newcommand{\ladderBoostedDegreeFamiliarLo}{84.6}  %
\newcommand{\ladderBoostedDegreeFamiliarHi}{85.6}  %
\newcommand{\ladderBoostedDegreeFreshPooled}{85.1}  %
\newcommand{\ladderBoostedDegreeFreshLo}{84.6}  %
\newcommand{\ladderBoostedDegreeFreshHi}{85.5}  %
\newcommand{\ladderBoostedDegreeRankRealPooled}{85.3}  %
\newcommand{\ladderBoostedDegreeRankRealLo}{85.0}  %
\newcommand{\ladderBoostedDegreeRankRealHi}{85.5}  %
\newcommand{\ladderBoostedDegreeRankFamiliarPooled}{85.2}  %
\newcommand{\ladderBoostedDegreeRankFamiliarLo}{84.8}  %
\newcommand{\ladderBoostedDegreeRankFamiliarHi}{85.7}  %
\newcommand{\ladderBoostedDegreeRankFreshPooled}{85.0}  %
\newcommand{\ladderBoostedDegreeRankFreshLo}{84.6}  %
\newcommand{\ladderBoostedDegreeRankFreshHi}{85.5}  %
\newcommand{\ladderBoostedOneHopRealPooled}{86.4}  %
\newcommand{\ladderBoostedOneHopRealLo}{86.2}  %
\newcommand{\ladderBoostedOneHopRealHi}{86.7}  %
\newcommand{\ladderBoostedOneHopFamiliarPooled}{86.2}  %
\newcommand{\ladderBoostedOneHopFamiliarLo}{85.7}  %
\newcommand{\ladderBoostedOneHopFamiliarHi}{86.7}  %
\newcommand{\ladderBoostedOneHopFreshPooled}{86.1}  %
\newcommand{\ladderBoostedOneHopFreshLo}{85.6}  %
\newcommand{\ladderBoostedOneHopFreshHi}{86.6}  %
\newcommand{\ladderBoostedTwoHopRealPooled}{86.6}  %
\newcommand{\ladderBoostedTwoHopRealLo}{86.4}  %
\newcommand{\ladderBoostedTwoHopRealHi}{86.8}  %
\newcommand{\ladderBoostedTwoHopFamiliarPooled}{86.3}  %
\newcommand{\ladderBoostedTwoHopFamiliarLo}{85.9}  %
\newcommand{\ladderBoostedTwoHopFamiliarHi}{86.8}  %
\newcommand{\ladderBoostedTwoHopFreshPooled}{86.3}  %
\newcommand{\ladderBoostedTwoHopFreshLo}{85.8}  %
\newcommand{\ladderBoostedTwoHopFreshHi}{86.7}  %
\newcommand{\ladderBoostedTwoHopFeaturesRealPooled}{86.6}  %
\newcommand{\ladderBoostedTwoHopFeaturesRealLo}{86.4}  %
\newcommand{\ladderBoostedTwoHopFeaturesRealHi}{86.8}  %
\newcommand{\ladderBoostedTwoHopFeaturesFamiliarPooled}{86.3}  %
\newcommand{\ladderBoostedTwoHopFeaturesFamiliarLo}{85.9}  %
\newcommand{\ladderBoostedTwoHopFeaturesFamiliarHi}{86.8}  %
\newcommand{\ladderBoostedTwoHopFeaturesFreshPooled}{86.3}  %
\newcommand{\ladderBoostedTwoHopFeaturesFreshLo}{85.8}  %
\newcommand{\ladderBoostedTwoHopFeaturesFreshHi}{86.8}  %
\newcommand{\ladderBestFreshPooled}{86.3}  %
\newcommand{\ladderBestFreshMinusDensityTwo}{$-$0.13}  %
\newcommand{\ladderBestFreshMinusDensityLoTwo}{$-$0.31}  %
\newcommand{\ladderBestFreshMinusDensityHiTwo}{0.04}  %
\newcommand{\ladderLargestRealMinusFreshTwo}{0.33}  %
\newcommand{\ladderBoostedTwoHopFeaturesRealMinusFreshTwo}{0.31}  %
\newcommand{\ladderHeldBoostedTwoHopHMinusKeptTwo}{0.20}  %
\newcommand{\ladderHeldLargestHMinusKeptTwo}{0.41}  %
\newcommand{\estPatternGraphs}{10,000}  %
\newcommand{\estPatternPHost}{0.50}  %
\newcommand{\estHostPairDensity}{0.38}  %
\newcommand{\estPatternPInternal}{0.50}  %
\newcommand{\estPatternSizeMin}{6}  %
\newcommand{\estPatternSizeMax}{34}  %
\newcommand{\estPatternHostBlocks}{2}  %
\newcommand{\estPatternBlocksTwoGraphs}{9,872}  %
\newcommand{\estPatternPIn}{0.38}  %
\newcommand{\estPatternPOut}{0.38}  %
\newcommand{\estPatternBlockMin}{23}  %
\newcommand{\estPatternBlockMax}{142}  %
\newcommand{\estPatternFiveBlocksPIn}{0.38}  %
\newcommand{\estPatternFiveBlocksPOut}{0.38}  %
\newcommand{\estPatternFiveBlocksBlockMin}{49}  %
\newcommand{\estPatternFiveBlocksBlockMax}{142}  %
\newcommand{\estPatternPriorMin}{6}  %
\newcommand{\estPatternPriorMax}{142}  %
\newcommand{\estRealDensityPooled}{86.6}  %
\newcommand{\estRealDensityLo}{86.4}  %
\newcommand{\estRealDensityHi}{86.9}  %
\newcommand{\estRealDensityMinusKnownTwo}{$-$0.11}  %
\newcommand{\estRealDensityMinusKnownLoTwo}{$-$0.19}  %
\newcommand{\estRealDensityMinusKnownHiTwo}{$-$0.04}  %
\newcommand{\estFreshDensityPooled}{86.3}  %
\newcommand{\estFreshDensityLo}{85.8}  %
\newcommand{\estFreshDensityHi}{86.8}  %
\newcommand{\estFreshDensityMinusKnownTwo}{$-$0.10}  %
\newcommand{\estFreshDensityMinusKnownLoTwo}{$-$0.24}  %
\newcommand{\estFreshDensityMinusKnownHiTwo}{0.05}  %
\newcommand{\estFiveBlocksRealDensityPooled}{86.7}  %
\newcommand{\estFiveBlocksRealDensityLo}{86.5}  %
\newcommand{\estFiveBlocksRealDensityHi}{86.9}  %
\newcommand{\estFiveBlocksRealDensityMinusKnownTwo}{$-$0.06}  %
\newcommand{\estFiveBlocksRealDensityMinusKnownLoTwo}{$-$0.13}  %
\newcommand{\estFiveBlocksRealDensityMinusKnownHiTwo}{0.01}  %
\newcommand{\estFiveBlocksFreshDensityPooled}{86.3}  %
\newcommand{\estFiveBlocksFreshDensityLo}{85.8}  %
\newcommand{\estFiveBlocksFreshDensityHi}{86.7}  %
\newcommand{\estFiveBlocksFreshDensityMinusKnownTwo}{$-$0.09}  %
\newcommand{\estFiveBlocksFreshDensityMinusKnownLoTwo}{$-$0.21}  %
\newcommand{\estFiveBlocksFreshDensityMinusKnownHiTwo}{0.03}  %
\newcommand{\estClusterPIn}{0.55}  %
\newcommand{\estClusterPOut}{0.25}  %
\newcommand{\estClusterBlocks}{6}  %
\newcommand{\estClusterBlockMin}{5}  %
\newcommand{\estClusterBlockMax}{34}  %
\newcommand{\estClusterPooled}{82.3}  %
\newcommand{\estClusterLo}{81.5}  %
\newcommand{\estClusterHi}{83.0}  %
\newcommand{\estClusterMinusCeilingTwo}{0.00}  %
\newcommand{\estClusterMinusCeilingLoTwo}{$-$0.09}  %
\newcommand{\estClusterMinusCeilingHiTwo}{0.10}  %
\newcommand{\estMinesRate}{0.20}  %
\newcommand{\estMinesFeaturesChange}{0.00}  %
\newcommand{\estMinesLabelsChange}{0.00}  %
\newcommand{\calibRealExpectedMinusRealizedTwo}{0.18}  %
\newcommand{\calibRealExpectedMinusRealizedLoTwo}{0.00}  %
\newcommand{\calibRealExpectedMinusRealizedHiTwo}{0.35}  %
\newcommand{\calibRealEce}{0.002}  %
\newcommand{\calibFamiliarExpectedMinusRealizedTwo}{0.29}  %
\newcommand{\calibFamiliarExpectedMinusRealizedLoTwo}{$-$0.05}  %
\newcommand{\calibFamiliarExpectedMinusRealizedHiTwo}{0.63}  %
\newcommand{\calibFamiliarEce}{0.003}  %
\newcommand{\calibFreshExpectedMinusRealizedTwo}{0.27}  %
\newcommand{\calibFreshExpectedMinusRealizedLoTwo}{$-$0.09}  %
\newcommand{\calibFreshExpectedMinusRealizedHiTwo}{0.66}  %
\newcommand{\calibFreshEce}{0.003}  %
\newcommand{\lookupRealPooled}{96.6}  %
\newcommand{\lookupRealLo}{96.2}  %
\newcommand{\lookupRealHi}{97.0}  %
\newcommand{\lookupFamiliarPooled}{96.9}  %
\newcommand{\lookupFreshPooled}{50.6}  %
\newcommand{\lookupDegreeRealPooled}{97.3}  %
\newcommand{\lookupDegreeRealLo}{96.9}  %
\newcommand{\lookupDegreeRealHi}{97.6}  %
\newcommand{\lookupDegreeFamiliarPooled}{97.5}  %
\newcommand{\lookupDegreeFreshPooled}{53.3}  %
\newcommand{\lookupRealExactPctBig}{99.7}  %
\newcommand{\lookupRealExactPctSmall}{34.8}  %
\newcommand{\lookupRealFoundBigPct}{76}  %
\newcommand{\lookupRealMinusMatcher}{$-$2.6}  %
\newcommand{\lookupRealMinusMatcherLo}{$-$2.9}  %
\newcommand{\lookupRealMinusMatcherHi}{$-$2.2}  %
\newcommand{\lookupRealMinusLargest}{$-$0.7}  %
\newcommand{\lookupRealMinusLargestLo}{$-$1.0}  %
\newcommand{\lookupRealMinusLargestHi}{$-$0.4}  %
\newcommand{\lookupDegreeRealMinusLargestTwo}{$-$0.02}  %
\newcommand{\lookupDegreeRealMinusLargestLoTwo}{$-$0.21}  %
\newcommand{\lookupDegreeRealMinusLargestHiTwo}{0.20}  %
\newcommand{\sizematchGraphs}{500}  %
\newcommand{\sizematchGraphsBig}{382}  %
\newcommand{\sizematchCkgSeedAFamiliar}{88.6}  %
\newcommand{\sizematchCkgSeedAMatched}{84.7}  %
\newcommand{\sizematchCkgSeedAFresh}{84.5}  %
\newcommand{\sizematchCkgSeedAFamiliarMinusMatched}{4.0}  %
\newcommand{\sizematchCkgSeedAFamiliarMinusMatchedLo}{3.2}  %
\newcommand{\sizematchCkgSeedAFamiliarMinusMatchedHi}{4.7}  %
\newcommand{\sizematchCkgSeedAFamiliarMinusMatchedTwo}{3.97}  %
\newcommand{\sizematchCkgSeedABigFamiliar}{88.8}  %
\newcommand{\sizematchCkgSeedABigMatched}{84.4}  %
\newcommand{\sizematchCkgSeedABigFresh}{84.1}  %
\newcommand{\sizematchCkgSeedABigFamiliarMinusMatched}{4.4}  %
\newcommand{\sizematchCkgSeedABigFamiliarMinusMatchedLo}{3.6}  %
\newcommand{\sizematchCkgSeedABigFamiliarMinusMatchedHi}{5.2}  %
\newcommand{\sizematchCkgSeedBFamiliar}{87.8}  %
\newcommand{\sizematchCkgSeedBMatched}{84.7}  %
\newcommand{\sizematchCkgSeedBFresh}{84.3}  %
\newcommand{\sizematchCkgSeedBFamiliarMinusMatched}{3.1}  %
\newcommand{\sizematchCkgSeedBFamiliarMinusMatchedLo}{2.4}  %
\newcommand{\sizematchCkgSeedBFamiliarMinusMatchedHi}{3.8}  %
\newcommand{\sizematchCkgSeedBFamiliarMinusMatchedTwo}{3.10}  %
\newcommand{\sizematchCkgSeedBBigFamiliar}{87.8}  %
\newcommand{\sizematchCkgSeedBBigMatched}{84.4}  %
\newcommand{\sizematchCkgSeedBBigFresh}{83.8}  %
\newcommand{\sizematchCkgSeedBBigFamiliarMinusMatched}{3.4}  %
\newcommand{\sizematchCkgSeedBBigFamiliarMinusMatchedLo}{2.7}  %
\newcommand{\sizematchCkgSeedBBigFamiliarMinusMatchedHi}{4.2}  %
\newcommand{\sizematchGritSeedAFamiliar}{87.0}  %
\newcommand{\sizematchGritSeedAMatched}{86.0}  %
\newcommand{\sizematchGritSeedAFresh}{85.8}  %
\newcommand{\sizematchGritSeedAFamiliarMinusMatched}{1.0}  %
\newcommand{\sizematchGritSeedAFamiliarMinusMatchedLo}{0.4}  %
\newcommand{\sizematchGritSeedAFamiliarMinusMatchedHi}{1.7}  %
\newcommand{\sizematchGritSeedAFamiliarMinusMatchedTwo}{1.02}  %
\newcommand{\sizematchGritSeedBFamiliar}{87.0}  %
\newcommand{\sizematchGritSeedBMatched}{85.9}  %
\newcommand{\sizematchGritSeedBFresh}{85.7}  %
\newcommand{\sizematchGritSeedBFamiliarMinusMatched}{1.1}  %
\newcommand{\sizematchGritSeedBFamiliarMinusMatchedLo}{0.5}  %
\newcommand{\sizematchGritSeedBFamiliarMinusMatchedHi}{1.8}  %
\newcommand{\sizematchGritSeedBFamiliarMinusMatchedTwo}{1.13}  %
\newcommand{\sizematchCkgNewSeedAAllFamiliarMinusMatched}{0.0}  %
\newcommand{\sizematchCkgNewSeedAFamiliar}{86.5}  %
\newcommand{\sizematchCkgNewSeedAMatched}{86.5}  %
\newcommand{\sizematchCkgNewSeedAFresh}{86.4}  %
\newcommand{\sizematchCkgNewSeedAFamiliarMinusMatched}{0.0}  %
\newcommand{\sizematchCkgNewSeedAFamiliarMinusMatchedLo}{$-$0.6}  %
\newcommand{\sizematchCkgNewSeedAFamiliarMinusMatchedHi}{0.7}  %
\newcommand{\sizematchCkgNewSeedBAllFamiliarMinusMatched}{0.1}  %
\newcommand{\sizematchCkgNewSeedBFamiliar}{86.6}  %
\newcommand{\sizematchCkgNewSeedBMatched}{86.5}  %
\newcommand{\sizematchCkgNewSeedBFresh}{86.5}  %
\newcommand{\sizematchCkgNewSeedBFamiliarMinusMatched}{0.1}  %
\newcommand{\sizematchCkgNewSeedBFamiliarMinusMatchedLo}{$-$0.6}  %
\newcommand{\sizematchCkgNewSeedBFamiliarMinusMatchedHi}{0.7}  %
\newcommand{\sizematchDensityFamiliar}{86.5}  %
\newcommand{\sizematchDensityMatched}{86.5}  %
\newcommand{\sizematchDensityFresh}{86.4}  %
\newcommand{\sizematchDensityFamiliarMinusMatched}{0.1}  %
\newcommand{\sizematchDensityFamiliarMinusMatchedLo}{$-$0.6}  %
\newcommand{\sizematchDensityFamiliarMinusMatchedHi}{0.7}  %
\newcommand{\sizematchLadderFamiliar}{86.3}  %
\newcommand{\sizematchLadderMatched}{86.3}  %
\newcommand{\sizematchLadderFresh}{86.3}  %
\newcommand{\sizematchLadderFamiliarMinusMatched}{0.0}  %
\newcommand{\sizematchLadderFamiliarMinusMatchedLo}{$-$0.6}  %
\newcommand{\sizematchLadderFamiliarMinusMatchedHi}{0.7}  %
\newcommand{\costGpuBatch}{16}  %
\newcommand{\costCpuThreads}{4}  %
\newcommand{\costCkgGpuSplit}{27}  %
\newcommand{\costCkgGpuGraph}{14}  %
\newcommand{\costCkgCpuSplit}{650}  %
\newcommand{\costCkgCpuGraph}{320}  %
\newcommand{\costGritGpuSplit}{28}  %
\newcommand{\costGritGpuGraph}{14}  %
\newcommand{\costGritCpuSplit}{940}  %
\newcommand{\costGritCpuGraph}{470}  %
\newcommand{\costGritCpuGraphs}{1,568}  %
\newcommand{\costCkgTrainLo}{19}  %
\newcommand{\costCkgTrainHi}{20}  %
\newcommand{\costGritTrain}{12}  %
\newcommand{\costMatcherSplit}{5.7}  %
\newcommand{\costMatcherGraph}{10}  %
\newcommand{\costLookupGraph}{41}  %
\newcommand{\costLookupSplit}{23}  %
\newcommand{\costDensitySplit}{2.1}  %
\newcommand{\costDensityGraph}{3.8}  %
\newcommand{\costDensityProcs}{4}  %
\newcommand{\costLadderSplit}{20}  %
\newcommand{\costLadderGraph}{10}  %
\newcommand{\costLadderProcs}{4}  %
\newcommand{\costLadderTrain}{2.3}  %
\newcommand{\gatedSeedAOfficialBatch}{86.6}  %
\newcommand{\gatedSeedAOfficialBatchTwo}{86.56}  %
\newcommand{\gatedSeedAEpochs}{51}  %
\newcommand{\gatedSeedBOfficialBatch}{86.6}  %
\newcommand{\gatedSeedBOfficialBatchTwo}{86.56}  %
\newcommand{\gatedSeedBEpochs}{62}  %
\newcommand{\pubPatternPPGTSd}{0.03}  %
\newcommand{\lookupFamiliarLo}{96.2}  %
\newcommand{\lookupFamiliarHi}{97.6}  %
\newcommand{\lookupFreshLo}{50.3}  %
\newcommand{\lookupFreshHi}{50.9}  %
\newcommand{\lookupDegreeFamiliarLo}{96.8}  %
\newcommand{\lookupDegreeFamiliarHi}{98.0}  %
\newcommand{\lookupDegreeFreshLo}{52.9}  %
\newcommand{\lookupDegreeFreshHi}{53.6}  %
\newcommand{\lookupFreshMinusDensity}{$-$35.8}  %
\newcommand{\lookupFreshMinusDensityLo}{$-$36.3}  %
\newcommand{\lookupFreshMinusDensityHi}{$-$35.2}  %
\newcommand{\gatedSeedASeed}{41}  %
\newcommand{\gatedSeedBSeed}{95}  %
\newcommand{\gatedEpochsCap}{66}  %
\newcommand{\regLadderDegreeLo}{80}  %
\newcommand{\regLadderDegreeHi}{86}  %
\newcommand{\regLadderAboveSampler}{1.0}  %
\newcommand{\regLadderShift}{1.0}  %
\newcommand{\regLadderBoostedMin}{84}  %
\newcommand{\regLadderHeldWithin}{1.0}  %
\newcommand{\regLadderFeaturesGap}{1.0}  %
\newcommand{\regLadderRemedy}{86}  %
\newcommand{\regEstRateWithin}{0.01}  %
\newcommand{\regEstRate}{0.5}  %
\newcommand{\regEstPInWithin}{0.03}  %
\newcommand{\regEstPIn}{0.5}  %
\newcommand{\regEstPOutWithin}{0.02}  %
\newcommand{\regEstPOut}{0.35}  %
\newcommand{\regEstBlocks}{5}  %
\newcommand{\regEstSamplerWithin}{0.5}  %
\newcommand{\regEstSamplerKill}{1.0}  %
\newcommand{\regEstClusterWithin}{0.01}  %
\newcommand{\regEstClusterSampler}{0.3}  %
\newcommand{\regEstClusterAnchor}{82.3}  %
\newcommand{\regEstMinesWithin}{0.02}  %
\newcommand{\regCalibWithin}{1.0}  %
\newcommand{\regCalibEce}{0.03}  %
\newcommand{\regGatedWithin}{1.0}  %
\newcommand{\regGatedDrop}{1.0}  %
\newcommand{\regGatedSamplerLo}{$-$2.0}  %
\newcommand{\regGatedSamplerHi}{+0.5}  %
\newcommand{\regSizeCkg}{2.5}  %
\newcommand{\regSizeGrit}{0.7}  %
\newcommand{\regSizeDensity}{1.0}  %
\newcommand{\regSizeKill}{1.0}  %
\newcommand{\regLookupLo}{96.0}  %
\newcommand{\regLookupHi}{97.5}  %
\newcommand{\regLookupDegreeWithin}{0.5}  %
\newcommand{\regLookupRule}{97.3}  %
\newcommand{\regLookupFreshLo}{50}  %
\newcommand{\regLookupFreshHi}{60}  %
\newcommand{\regLookupFamiliarWithin}{1.0}  %
\newcommand{\lookupFamiliarMinusRealTwo}{0.28}  %
\newcommand{\lookupDegreeFamiliarMinusRealTwo}{0.18}  %
\newcommand{\fiveSampler}{86.76}  %
\newcommand{\fiveDensityDrop}{0.1}  %
\newcommand{\fiveGatedSeedANumber}{41}  %
\newcommand{\fiveGatedSeedAOfficialTwo}{86.75}  %
\newcommand{\fiveGatedSeedAFamiliarTwo}{86.63}  %
\newcommand{\fiveGatedSeedAFreshTwo}{86.19}  %
\newcommand{\fiveGatedSeedADropTwo}{0.43}  %
\newcommand{\fiveGatedSeedAFreshMinusDensityTwo}{$-$0.21}  %
\newcommand{\fiveGatedSeedAFreshMinusDensityLoTwo}{$-$0.38}  %
\newcommand{\fiveGatedSeedAFreshMinusDensityHiTwo}{$-$0.04}  %
\newcommand{\fiveGatedSeedBNumber}{95}  %
\newcommand{\fiveGatedSeedBOfficialTwo}{86.84}  %
\newcommand{\fiveGatedSeedBFamiliarTwo}{86.64}  %
\newcommand{\fiveGatedSeedBFreshTwo}{86.18}  %
\newcommand{\fiveGatedSeedBDropTwo}{0.46}  %
\newcommand{\fiveGatedSeedBFreshMinusDensityTwo}{$-$0.22}  %
\newcommand{\fiveGatedSeedBFreshMinusDensityLoTwo}{$-$0.41}  %
\newcommand{\fiveGatedSeedBFreshMinusDensityHiTwo}{$-$0.01}  %
\newcommand{\fiveGatedSeedCNumber}{12}  %
\newcommand{\fiveGatedSeedCOfficialTwo}{86.60}  %
\newcommand{\fiveGatedSeedCFamiliarTwo}{86.49}  %
\newcommand{\fiveGatedSeedCFreshTwo}{86.10}  %
\newcommand{\fiveGatedSeedCDropTwo}{0.39}  %
\newcommand{\fiveGatedSeedCFreshMinusDensityTwo}{$-$0.30}  %
\newcommand{\fiveGatedSeedCFreshMinusDensityLoTwo}{$-$0.51}  %
\newcommand{\fiveGatedSeedCFreshMinusDensityHiTwo}{$-$0.10}  %
\newcommand{\fiveGatedSeedDNumber}{35}  %
\newcommand{\fiveGatedSeedDOfficialTwo}{86.71}  %
\newcommand{\fiveGatedSeedDFamiliarTwo}{86.64}  %
\newcommand{\fiveGatedSeedDFreshTwo}{86.24}  %
\newcommand{\fiveGatedSeedDDropTwo}{0.39}  %
\newcommand{\fiveGatedSeedDFreshMinusDensityTwo}{$-$0.16}  %
\newcommand{\fiveGatedSeedDFreshMinusDensityLoTwo}{$-$0.36}  %
\newcommand{\fiveGatedSeedDFreshMinusDensityHiTwo}{0.03}  %
\newcommand{\fiveGatedSeeds}{4}  %
\newcommand{\fiveGatedSeedsWord}{four}  %
\newcommand{\fiveGatedMeanOfficial}{86.7}  %
\newcommand{\fiveGatedMeanFamiliar}{86.6}  %
\newcommand{\fiveGatedMeanFresh}{86.2}  %
\newcommand{\fiveGatedMeanDrop}{0.4}  %
\newcommand{\fiveGatedMeanFreshMinusDensity}{$-$0.2}  %
\newcommand{\fiveGatedFirstTwoDropTwo}{0.44}  %
\newcommand{\fiveGritSeedANumber}{0}  %
\newcommand{\fiveGritSeedAOfficialTwo}{87.20}  %
\newcommand{\fiveGritSeedAFamiliarTwo}{87.00}  %
\newcommand{\fiveGritSeedAFreshTwo}{85.80}  %
\newcommand{\fiveGritSeedADropTwo}{1.20}  %
\newcommand{\fiveGritSeedAFreshMinusDensityTwo}{$-$0.60}  %
\newcommand{\fiveGritSeedAFreshMinusDensityLoTwo}{$-$0.80}  %
\newcommand{\fiveGritSeedAFreshMinusDensityHiTwo}{$-$0.39}  %
\newcommand{\fiveGritSeedBNumber}{1}  %
\newcommand{\fiveGritSeedBOfficialTwo}{87.20}  %
\newcommand{\fiveGritSeedBFamiliarTwo}{87.04}  %
\newcommand{\fiveGritSeedBFreshTwo}{85.67}  %
\newcommand{\fiveGritSeedBDropTwo}{1.37}  %
\newcommand{\fiveGritSeedBFreshMinusDensityTwo}{$-$0.73}  %
\newcommand{\fiveGritSeedBFreshMinusDensityLoTwo}{$-$0.94}  %
\newcommand{\fiveGritSeedBFreshMinusDensityHiTwo}{$-$0.52}  %
\newcommand{\fiveGritSeedCNumber}{2}  %
\newcommand{\fiveGritSeedCOfficialTwo}{87.31}  %
\newcommand{\fiveGritSeedCFamiliarTwo}{87.07}  %
\newcommand{\fiveGritSeedCFreshTwo}{85.56}  %
\newcommand{\fiveGritSeedCDropTwo}{1.51}  %
\newcommand{\fiveGritSeedCFreshMinusDensityTwo}{$-$0.84}  %
\newcommand{\fiveGritSeedCFreshMinusDensityLoTwo}{$-$1.07}  %
\newcommand{\fiveGritSeedCFreshMinusDensityHiTwo}{$-$0.60}  %
\newcommand{\fiveGritSeeds}{3}  %
\newcommand{\fiveGritSeedsWord}{three}  %
\newcommand{\fiveGritMeanOfficial}{87.2}  %
\newcommand{\fiveGritMeanFamiliar}{87.0}  %
\newcommand{\fiveGritMeanFresh}{85.7}  %
\newcommand{\fiveGritMeanDrop}{1.4}  %
\newcommand{\fiveGritMeanFreshMinusDensity}{$-$0.7}  %
\newcommand{\fiveGritFirstTwoDropTwo}{1.29}  %
\newcommand{\fiveCkgSeedANumber}{0}  %
\newcommand{\fiveCkgSeedAOfficialTwo}{88.81}  %
\newcommand{\fiveCkgSeedAFamiliarTwo}{88.64}  %
\newcommand{\fiveCkgSeedAFreshTwo}{84.48}  %
\newcommand{\fiveCkgSeedADropTwo}{4.17}  %
\newcommand{\fiveCkgSeedAFreshMinusDensityTwo}{$-$1.93}  %
\newcommand{\fiveCkgSeedAFreshMinusDensityLoTwo}{$-$2.22}  %
\newcommand{\fiveCkgSeedAFreshMinusDensityHiTwo}{$-$1.62}  %
\newcommand{\fiveCkgSeedBNumber}{1}  %
\newcommand{\fiveCkgSeedBOfficialTwo}{88.28}  %
\newcommand{\fiveCkgSeedBFamiliarTwo}{87.82}  %
\newcommand{\fiveCkgSeedBFreshTwo}{84.34}  %
\newcommand{\fiveCkgSeedBDropTwo}{3.48}  %
\newcommand{\fiveCkgSeedBFreshMinusDensityTwo}{$-$2.06}  %
\newcommand{\fiveCkgSeedBFreshMinusDensityLoTwo}{$-$2.38}  %
\newcommand{\fiveCkgSeedBFreshMinusDensityHiTwo}{$-$1.74}  %
\newcommand{\fiveCkgSeedCNumber}{2}  %
\newcommand{\fiveCkgSeedCOfficialTwo}{88.43}  %
\newcommand{\fiveCkgSeedCFamiliarTwo}{88.17}  %
\newcommand{\fiveCkgSeedCFreshTwo}{84.66}  %
\newcommand{\fiveCkgSeedCDropTwo}{3.52}  %
\newcommand{\fiveCkgSeedCFreshMinusDensityTwo}{$-$1.75}  %
\newcommand{\fiveCkgSeedCFreshMinusDensityLoTwo}{$-$2.06}  %
\newcommand{\fiveCkgSeedCFreshMinusDensityHiTwo}{$-$1.45}  %
\newcommand{\fiveCkgSeeds}{3}  %
\newcommand{\fiveCkgMeanOfficial}{88.5}  %
\newcommand{\fiveCkgMeanFamiliar}{88.2}  %
\newcommand{\fiveCkgMeanFresh}{84.5}  %
\newcommand{\fiveCkgMeanDrop}{3.7}  %
\newcommand{\fiveCkgMeanFreshMinusDensity}{$-$1.9}  %
\newcommand{\fiveCkgFirstTwoDropTwo}{3.82}  %
\newcommand{\fiveGrassSeedANumber}{0}  %
\newcommand{\fiveGrassSeedAOfficialTwo}{90.16}  %
\newcommand{\fiveGrassSeedAFamiliarTwo}{90.21}  %
\newcommand{\fiveGrassSeedAFreshTwo}{81.41}  %
\newcommand{\fiveGrassSeedADropTwo}{8.80}  %
\newcommand{\fiveGrassSeedAFreshMinusDensityTwo}{$-$4.99}  %
\newcommand{\fiveGrassSeedAFreshMinusDensityLoTwo}{$-$5.40}  %
\newcommand{\fiveGrassSeedAFreshMinusDensityHiTwo}{$-$4.61}  %
\newcommand{\fiveGrassSeedBNumber}{1}  %
\newcommand{\fiveGrassSeedBOfficialTwo}{90.47}  %
\newcommand{\fiveGrassSeedBFamiliarTwo}{90.15}  %
\newcommand{\fiveGrassSeedBFreshTwo}{82.00}  %
\newcommand{\fiveGrassSeedBDropTwo}{8.15}  %
\newcommand{\fiveGrassSeedBFreshMinusDensityTwo}{$-$4.40}  %
\newcommand{\fiveGrassSeedBFreshMinusDensityLoTwo}{$-$4.79}  %
\newcommand{\fiveGrassSeedBFreshMinusDensityHiTwo}{$-$4.02}  %
\newcommand{\fiveGrassSeedCNumber}{2}  %
\newcommand{\fiveGrassSeedCOfficialTwo}{90.52}  %
\newcommand{\fiveGrassSeedCFamiliarTwo}{90.29}  %
\newcommand{\fiveGrassSeedCFreshTwo}{82.32}  %
\newcommand{\fiveGrassSeedCDropTwo}{7.97}  %
\newcommand{\fiveGrassSeedCFreshMinusDensityTwo}{$-$4.09}  %
\newcommand{\fiveGrassSeedCFreshMinusDensityLoTwo}{$-$4.48}  %
\newcommand{\fiveGrassSeedCFreshMinusDensityHiTwo}{$-$3.71}  %
\newcommand{\fiveGrassSeeds}{3}  %
\newcommand{\fiveGrassSeedsWord}{three}  %
\newcommand{\fiveGrassMeanOfficial}{90.4}  %
\newcommand{\fiveGrassMeanFamiliar}{90.2}  %
\newcommand{\fiveGrassMeanFresh}{81.9}  %
\newcommand{\fiveGrassMeanDrop}{8.3}  %
\newcommand{\fiveGrassMeanFreshMinusDensity}{$-$4.5}  %
\newcommand{\fiveGrassFirstTwoDropTwo}{8.47}  %
\newcommand{\fivePpgtSeedANumber}{0}  %
\newcommand{\fivePpgtSeedAOfficialTwo}{89.97}  %
\newcommand{\fivePpgtSeedAFamiliarTwo}{90.01}  %
\newcommand{\fivePpgtSeedAFreshTwo}{83.56}  %
\newcommand{\fivePpgtSeedADrop}{6.4}  %
\newcommand{\fivePpgtSeedADropTwo}{6.44}  %
\newcommand{\fivePpgtSeedAFreshMinusDensityTwo}{$-$2.84}  %
\newcommand{\fivePpgtSeedAFreshMinusDensityLoTwo}{$-$3.17}  %
\newcommand{\fivePpgtSeedAFreshMinusDensityHiTwo}{$-$2.49}  %
\newcommand{\fivePpgtSeedBNumber}{1}  %
\newcommand{\fivePpgtSeedBOfficialTwo}{90.93}  %
\newcommand{\fivePpgtSeedBFamiliarTwo}{90.56}  %
\newcommand{\fivePpgtSeedBFreshTwo}{81.75}  %
\newcommand{\fivePpgtSeedBDropTwo}{8.81}  %
\newcommand{\fivePpgtSeedBFreshMinusDensityTwo}{$-$4.66}  %
\newcommand{\fivePpgtSeedBFreshMinusDensityLoTwo}{$-$5.05}  %
\newcommand{\fivePpgtSeedBFreshMinusDensityHiTwo}{$-$4.25}  %
\newcommand{\fivePpgtSeedCNumber}{2}  %
\newcommand{\fivePpgtSeedCOfficialTwo}{90.79}  %
\newcommand{\fivePpgtSeedCFamiliarTwo}{90.71}  %
\newcommand{\fivePpgtSeedCFreshTwo}{82.01}  %
\newcommand{\fivePpgtSeedCDropTwo}{8.70}  %
\newcommand{\fivePpgtSeedCFreshMinusDensityTwo}{$-$4.39}  %
\newcommand{\fivePpgtSeedCFreshMinusDensityLoTwo}{$-$4.78}  %
\newcommand{\fivePpgtSeedCFreshMinusDensityHiTwo}{$-$4.01}  %
\newcommand{\fivePpgtSeedDNumber}{3}  %
\newcommand{\fivePpgtSeedDOfficialTwo}{91.15}  %
\newcommand{\fivePpgtSeedDFamiliarTwo}{90.91}  %
\newcommand{\fivePpgtSeedDFreshTwo}{82.20}  %
\newcommand{\fivePpgtSeedDDropTwo}{8.70}  %
\newcommand{\fivePpgtSeedDFreshMinusDensityTwo}{$-$4.20}  %
\newcommand{\fivePpgtSeedDFreshMinusDensityLoTwo}{$-$4.56}  %
\newcommand{\fivePpgtSeedDFreshMinusDensityHiTwo}{$-$3.82}  %
\newcommand{\fivePpgtSeeds}{4}  %
\newcommand{\fivePpgtMeanOfficial}{90.7}  %
\newcommand{\fivePpgtMeanFamiliar}{90.5}  %
\newcommand{\fivePpgtMeanFresh}{82.4}  %
\newcommand{\fivePpgtMeanDrop}{8.2}  %
\newcommand{\fivePpgtMeanFreshMinusDensity}{$-$4.0}  %
\newcommand{\fivePpgtFirstTwoDropTwo}{7.63}  %
\newcommand{\fiveSpearman}{0.90}  %
\newcommand{\fiveSpearmanAtLeast}{0.7}  %
\newcommand{\fiveSpearmanAllSeeds}{0.90}  %
\newcommand{\fiveGrassAccumulation}{2}  %
\newcommand{\fiveGrassMicroBatch}{100}  %
\newcommand{\fiveGrassReleasedBatch}{200}  %
\newcommand{\fiveGrassEpochs}{500}  %
\newcommand{\fiveGrassPasses}{ten}  %
\newcommand{\fivePpgtEpochs}{400}  %
\newcommand{\fivePpgtSeedABestEpoch}{211}  %
\newcommand{\fivePpgtSeedBBestEpoch}{389}  %
\newcommand{\fivePpgtSeedCBestEpoch}{384}  %
\newcommand{\fivePpgtSeedDBestEpoch}{332}  %
\newcommand{\regGrassRoundtrip}{0.1}  %
\newcommand{\regGrassGate}{1.5}  %
\newcommand{\regGrassFamiliar}{1.0}  %
\newcommand{\regGrassFall}{2.0}  %
\newcommand{\regGrassNear}{1.0}  %
\newcommand{\regGrassAuc}{1.0}  %
\newcommand{\regGrassKill}{1.0}  %
\newcommand{\regGrassPublished}{89.18}  %
\newcommand{\grassSeedAGate}{90.15}  %
\newcommand{\grassSeedARoundtrip}{0.00}  %
\newcommand{\grassSeedAFamiliarMinusOfficial}{0.05}  %
\newcommand{\grassSeedAOfficialMinusFresh}{8.74}  %
\newcommand{\grassSeedAFreshMinusCue}{$-$4.68}  %
\newcommand{\grassSeedAAucGap}{7.26}  %
\newcommand{\grassSeedBGate}{90.46}  %
\newcommand{\grassSeedBRoundtrip}{$-$0.01}  %
\newcommand{\grassSeedBFamiliarMinusOfficial}{$-$0.31}  %
\newcommand{\grassSeedBOfficialMinusFresh}{8.46}  %
\newcommand{\grassSeedBFreshMinusCue}{$-$4.10}  %
\newcommand{\grassSeedBAucGap}{6.56}  %
\newcommand{\regPpgtRoundtrip}{0.1}  %
\newcommand{\regPpgtGate}{1.5}  %
\newcommand{\regPpgtFamiliar}{1.0}  %
\newcommand{\regPpgtFall}{2.0}  %
\newcommand{\regPpgtNear}{1.0}  %
\newcommand{\regPpgtAuc}{1.0}  %
\newcommand{\regPpgtKill}{1.0}  %
\newcommand{\regPpgtPublished}{89.75}  %
\newcommand{\ppgtSeedAFamiliarMinusOfficial}{0.04}  %
\newcommand{\ppgtSeedAOfficialMinusFresh}{6.40}  %
\newcommand{\ppgtSeedAAucGap}{4.92}  %
\newcommand{\ppgtSeedBFamiliarMinusOfficial}{$-$0.38}  %
\newcommand{\ppgtSeedBOfficialMinusFresh}{9.19}  %
\newcommand{\ppgtSeedBAucGap}{6.57}  %
\newcommand{\regToneWithin}{1.0}  %
\newcommand{\hvCompletedGritSeedCBestEpoch}{99}  %
\newcommand{\gritSeedCAucGap}{1.14}  %
\newcommand{\gatedSeedCOfficialBatchTwo}{86.40}  %
\newcommand{\gatedSeedDOfficialBatchTwo}{86.47}  %
\newcommand{\ckgSeedCSizeMatchedDropTwo}{3.17}  %
\newcommand{\ckgSeedCSizeMatchedDropLoTwo}{2.47}  %
\newcommand{\ckgSeedCSizeMatchedDropHiTwo}{3.91}  %
\newcommand{\ppgtSeedCToneOverTwo}{1.07}  %
\newcommand{\ppgtSeedDToneOverTwo}{1.08}  %
\newcommand{\heldSecondGraphs}{400}  %
\newcommand{\heldSecondKeptGraphs}{1,600}  %
\newcommand{\heldSecondDensityH}{86.9}  %
\newcommand{\heldSecondDensityHLo}{86.3}  %
\newcommand{\heldSecondDensityHHi}{87.3}  %
\newcommand{\heldSecondH}{84.3}  %
\newcommand{\heldSecondHLo}{83.7}  %
\newcommand{\heldSecondHHi}{84.8}  %
\newcommand{\heldSecondKept}{88.7}  %
\newcommand{\heldSecondKeptLo}{88.4}  %
\newcommand{\heldSecondKeptHi}{89.0}  %
\newcommand{\heldSecondFullH}{89.4}  %
\newcommand{\heldSecondBelowCue}{2.2}  %
\newcommand{\heldSecondKeptMinusFull}{0.1}  %
\newcommand{\heldSecondKeptMinusFullLo}{$-$0.1}  %
\newcommand{\heldSecondKeptMinusFullHi}{0.2}  %
\newcommand{\heldSecondHMinusDensity}{$-$2.6}  %
\newcommand{\heldSecondHMinusDensityLo}{$-$2.9}  %
\newcommand{\heldSecondHMinusDensityHi}{$-$2.2}  %
\newcommand{\heldSecondBelowDensity}{2.6}  %
\newcommand{\heldSecondHMinusFull}{$-$5.1}  %
\newcommand{\heldSecondHMinusFullLo}{$-$5.6}  %
\newcommand{\heldSecondHMinusFullHi}{$-$4.6}  %
\newcommand{\heldSecondSeedBH}{84.4}  %
\newcommand{\heldSecondSeedBHLo}{83.9}  %
\newcommand{\heldSecondSeedBHHi}{84.9}  %
\newcommand{\heldSecondSeedBKept}{88.4}  %
\newcommand{\heldSecondSeedBKeptLo}{88.1}  %
\newcommand{\heldSecondSeedBKeptHi}{88.7}  %
\newcommand{\heldSecondSeedBFullH}{88.8}  %
\newcommand{\heldSecondSeedBBelowCue}{2.0}  %
\newcommand{\heldSecondSeedBKeptMinusFull}{0.3}  %
\newcommand{\heldSecondSeedBKeptMinusFullLo}{0.1}  %
\newcommand{\heldSecondSeedBKeptMinusFullHi}{0.5}  %
\newcommand{\heldSecondSeedBHMinusDensity}{$-$2.4}  %
\newcommand{\heldSecondSeedBHMinusDensityLo}{$-$2.8}  %
\newcommand{\heldSecondSeedBHMinusDensityHi}{$-$2.1}  %
\newcommand{\heldSecondSeedBBelowDensity}{2.4}  %
\newcommand{\heldSecondSeedBHMinusFull}{$-$4.4}  %
\newcommand{\heldSecondSeedBHMinusFullLo}{$-$4.9}  %
\newcommand{\heldSecondSeedBHMinusFullHi}{$-$4.0}  %
\newcommand{\hvCompletedCkgSeedCOfficialAccuracy}{86.79}  %
\newcommand{\hvCompletedCkgSeedCFullRecipeOfficial}{88.43}  %
\newcommand{\hvCompletedCkgSeedCFreshAccuracy}{86.52}  %
\newcommand{\hvCompletedCkgSeedCFullRecipeFresh}{84.66}  %
\newcommand{\hvCompletedCkgSeedCFamiliarAccuracy}{86.57}  %
\newcommand{\hvCompletedCkgSeedCFullRecipeFamiliar}{88.17}  %
\newcommand{\hvCompletedCkgSeedCFreshImprovement}{1.86}  %
\newcommand{\hvCompletedCkgSeedCFreshImprovementInterval}{(1.56 to 2.16)}  %
\newcommand{\hvCompletedCkgSeedCFreshMinusBlind}{0.11}  %
\newcommand{\hvCompletedCkgSeedCFreshMinusBlindInterval}{($-$0.02 to 0.24)}  %
\newcommand{\hvCompletedCkgSeedCEpochs}{200}  %
\newcommand{\hvCompletedCkgSeedCBestEpoch}{39}  %
\newcommand{\hvCompletedCkgSeedCFullRecipeBestEpoch}{134}  %
\newcommand{\reuseRunOneSeed}{0}  %
\newcommand{\reuseRunOneEpoch}{119}  %
\newcommand{\reuseRunOneLibrary}{88.2}  %
\newcommand{\reuseRunOneFresh}{84.9}  %
\newcommand{\reuseRunOneOfficial}{85.2}  %
\newcommand{\reuseRunOneFamiliar}{85.3}  %
\newcommand{\reuseRunOneLibraryMinusFresh}{3.4}  %
\newcommand{\reuseRunOneLibraryMinusFreshLo}{2.6}  %
\newcommand{\reuseRunOneLibraryMinusFreshHi}{4.1}  %
\newcommand{\reuseRunOneMinusNewPatternTraining}{$-$1.6}  %
\newcommand{\reuseRunOneMinusNewPatternTrainingLo}{$-$1.8}  %
\newcommand{\reuseRunOneMinusNewPatternTrainingHi}{$-$1.3}  %
\newcommand{\reuseRunOneMinusReleasedPatternTraining}{$-$3.6}  %
\newcommand{\reuseRunOneMinusReleasedPatternTrainingLo}{$-$3.8}  %
\newcommand{\reuseRunOneMinusReleasedPatternTrainingHi}{$-$3.4}  %
\newcommand{\reuseRunOneOfficialMinusDensity}{$-$1.5}  %
\newcommand{\reuseRunOneOfficialMinusDensityLo}{$-$1.7}  %
\newcommand{\reuseRunOneOfficialMinusDensityHi}{$-$1.4}  %
\newcommand{\reuseRunOneFreshMinusDensity}{$-$1.5}  %
\newcommand{\reuseRunOneFreshMinusDensityLo}{$-$1.8}  %
\newcommand{\reuseRunOneFreshMinusDensityHi}{$-$1.3}  %
\newcommand{\reuseRunTwoSeed}{1}  %
\newcommand{\reuseRunTwoEpoch}{155}  %
\newcommand{\reuseRunTwoLibrary}{88.7}  %
\newcommand{\reuseRunTwoFresh}{84.7}  %
\newcommand{\reuseRunTwoOfficial}{85.0}  %
\newcommand{\reuseRunTwoFamiliar}{85.1}  %
\newcommand{\reuseRunTwoLibraryMinusFresh}{3.9}  %
\newcommand{\reuseRunTwoLibraryMinusFreshLo}{3.2}  %
\newcommand{\reuseRunTwoLibraryMinusFreshHi}{4.6}  %
\newcommand{\reuseRunTwoMinusNewPatternTraining}{$-$1.7}  %
\newcommand{\reuseRunTwoMinusNewPatternTrainingLo}{$-$2.0}  %
\newcommand{\reuseRunTwoMinusNewPatternTrainingHi}{$-$1.4}  %
\newcommand{\reuseRunTwoMinusReleasedPatternTraining}{$-$3.2}  %
\newcommand{\reuseRunTwoMinusReleasedPatternTrainingLo}{$-$3.4}  %
\newcommand{\reuseRunTwoMinusReleasedPatternTrainingHi}{$-$3.0}  %
\newcommand{\reuseRunTwoOfficialMinusDensity}{$-$1.7}  %
\newcommand{\reuseRunTwoOfficialMinusDensityLo}{$-$1.8}  %
\newcommand{\reuseRunTwoOfficialMinusDensityHi}{$-$1.6}  %
\newcommand{\reuseRunTwoFreshMinusDensity}{$-$1.7}  %
\newcommand{\reuseRunTwoFreshMinusDensityLo}{$-$1.9}  %
\newcommand{\reuseRunTwoFreshMinusDensityHi}{$-$1.4}  %
\newcommand{\reuseRunThreeSeed}{2}  %
\newcommand{\reuseRunThreeEpoch}{190}  %
\newcommand{\reuseRunThreeLibrary}{89.0}  %
\newcommand{\reuseRunThreeFresh}{84.0}  %
\newcommand{\reuseRunThreeOfficial}{84.5}  %
\newcommand{\reuseRunThreeFamiliar}{84.7}  %
\newcommand{\reuseRunThreeLibraryMinusFresh}{5.0}  %
\newcommand{\reuseRunThreeLibraryMinusFreshLo}{4.3}  %
\newcommand{\reuseRunThreeLibraryMinusFreshHi}{5.7}  %
\newcommand{\reuseRunThreeMinusNewPatternTraining}{$-$2.5}  %
\newcommand{\reuseRunThreeMinusNewPatternTrainingLo}{$-$2.8}  %
\newcommand{\reuseRunThreeMinusNewPatternTrainingHi}{$-$2.2}  %
\newcommand{\reuseRunThreeMinusReleasedPatternTraining}{$-$3.9}  %
\newcommand{\reuseRunThreeMinusReleasedPatternTrainingLo}{$-$4.1}  %
\newcommand{\reuseRunThreeMinusReleasedPatternTrainingHi}{$-$3.7}  %
\newcommand{\reuseRunThreeOfficialMinusDensity}{$-$2.2}  %
\newcommand{\reuseRunThreeOfficialMinusDensityLo}{$-$2.4}  %
\newcommand{\reuseRunThreeOfficialMinusDensityHi}{$-$2.1}  %
\newcommand{\reuseRunThreeFreshMinusDensity}{$-$2.4}  %
\newcommand{\reuseRunThreeFreshMinusDensityLo}{$-$2.7}  %
\newcommand{\reuseRunThreeFreshMinusDensityHi}{$-$2.1}  %
\newcommand{\reuseTrainGraphs}{10,000}  %
\newcommand{\reuseValGraphs}{2,000}  %
\newcommand{\reuseTestGraphs}{500}  %
\newcommand{\reuseLibraryPatterns}{100}  %
\newcommand{\regReuseGap}{2.0}  %
\newcommand{\regReuseOfficialBelow}{1.5}  %
\newcommand{\regReuseKill}{1.0}  %
\newcommand{\reuseRunOneOfficialMinusSamplerTwo}{$-$1.55}  %
\newcommand{\reuseRunTwoOfficialMinusSamplerTwo}{$-$1.71}  %
\newcommand{\reuseRunThreeOfficialMinusSamplerTwo}{$-$2.23}  %
\newcommand{\grassSeedCGate}{90.51}  %
\newcommand{\grassSeedCFreshMinusCue}{$-$3.78}  %
\newcommand{\grassSeedCRoundtrip}{$-$0.02}  %
\newcommand{\regLibrefWithin}{0.6}  %
\newcommand{\regLibrefAbove}{1.0}  %
\newcommand{\regLibrefBand}{0.5}  %
\newcommand{\librefOneBlindLibrary}{86.7}  %
\newcommand{\librefOneBlindLibraryMinusFresh}{0.3}  %
\newcommand{\librefOneBlindLibraryMinusFreshLo}{$-$0.4}  %
\newcommand{\librefOneBlindLibraryMinusFreshHi}{0.9}  %
\newcommand{\librefOneReuseMinusBlind}{1.6}  %
\newcommand{\librefOneReuseMinusBlindLo}{1.2}  %
\newcommand{\librefOneReuseMinusBlindHi}{1.9}  %
\newcommand{\librefOneReuseMinusNew}{1.6}  %
\newcommand{\librefOneReuseMinusNewLo}{1.2}  %
\newcommand{\librefOneReuseMinusNewHi}{1.9}  %
\newcommand{\librefOneNewLibrary}{86.7}  %
\newcommand{\librefOneNewMinusBlind}{0.0}  %
\newcommand{\librefOneNewMinusBlindLo}{$-$0.1}  %
\newcommand{\librefOneNewMinusBlindHi}{0.1}  %
\newcommand{\librefOneNewLibraryMinusFresh}{0.2}  %
\newcommand{\librefOneNewLibraryMinusFreshLo}{$-$0.4}  %
\newcommand{\librefOneNewLibraryMinusFreshHi}{0.9}  %
\newcommand{\librefOneMatched}{85.0}  %
\newcommand{\librefOneFamiliarMinusMatched}{0.3}  %
\newcommand{\librefOneFamiliarMinusMatchedLo}{$-$0.3}  %
\newcommand{\librefOneFamiliarMinusMatchedHi}{1.0}  %
\newcommand{\librefOneMatchedMinusFresh}{0.1}  %
\newcommand{\librefOneMatchedMinusFreshLo}{$-$0.6}  %
\newcommand{\librefOneMatchedMinusFreshHi}{0.8}  %
\newcommand{\librefOnePatternSize}{19.1}  %
\newcommand{\librefOneSmallPatterns}{155}  %
\newcommand{\librefTwoBlindLibrary}{87.0}  %
\newcommand{\librefTwoBlindLibraryMinusFresh}{0.6}  %
\newcommand{\librefTwoBlindLibraryMinusFreshLo}{0.0}  %
\newcommand{\librefTwoBlindLibraryMinusFreshHi}{1.3}  %
\newcommand{\librefTwoReuseMinusBlind}{1.7}  %
\newcommand{\librefTwoReuseMinusBlindLo}{1.3}  %
\newcommand{\librefTwoReuseMinusBlindHi}{2.0}  %
\newcommand{\librefTwoReuseMinusNew}{1.8}  %
\newcommand{\librefTwoReuseMinusNewLo}{1.4}  %
\newcommand{\librefTwoReuseMinusNewHi}{2.2}  %
\newcommand{\librefTwoNewLibrary}{86.9}  %
\newcommand{\librefTwoNewMinusBlind}{$-$0.1}  %
\newcommand{\librefTwoNewMinusBlindLo}{$-$0.3}  %
\newcommand{\librefTwoNewMinusBlindHi}{0.1}  %
\newcommand{\librefTwoNewLibraryMinusFresh}{0.4}  %
\newcommand{\librefTwoNewLibraryMinusFreshLo}{$-$0.2}  %
\newcommand{\librefTwoNewLibraryMinusFreshHi}{1.1}  %
\newcommand{\librefTwoMatched}{84.9}  %
\newcommand{\librefTwoFamiliarMinusMatched}{0.2}  %
\newcommand{\librefTwoFamiliarMinusMatchedLo}{$-$0.5}  %
\newcommand{\librefTwoFamiliarMinusMatchedHi}{0.9}  %
\newcommand{\librefTwoMatchedMinusFresh}{0.2}  %
\newcommand{\librefTwoMatchedMinusFreshLo}{$-$0.5}  %
\newcommand{\librefTwoMatchedMinusFreshHi}{0.9}  %
\newcommand{\librefTwoPatternSize}{19.9}  %
\newcommand{\librefTwoSmallPatterns}{150}  %
\newcommand{\librefThreeBlindLibrary}{86.8}  %
\newcommand{\librefThreeBlindLibraryMinusFresh}{0.4}  %
\newcommand{\librefThreeBlindLibraryMinusFreshLo}{$-$0.2}  %
\newcommand{\librefThreeBlindLibraryMinusFreshHi}{1.1}  %
\newcommand{\librefThreeReuseMinusBlind}{2.1}  %
\newcommand{\librefThreeReuseMinusBlindLo}{1.8}  %
\newcommand{\librefThreeReuseMinusBlindHi}{2.5}  %
\newcommand{\librefThreeReuseMinusNew}{2.1}  %
\newcommand{\librefThreeReuseMinusNewLo}{1.8}  %
\newcommand{\librefThreeReuseMinusNewHi}{2.5}  %
\newcommand{\librefThreeNewLibrary}{86.9}  %
\newcommand{\librefThreeNewMinusBlind}{0.0}  %
\newcommand{\librefThreeNewMinusBlindLo}{$-$0.1}  %
\newcommand{\librefThreeNewMinusBlindHi}{0.1}  %
\newcommand{\librefThreeNewLibraryMinusFresh}{0.3}  %
\newcommand{\librefThreeNewLibraryMinusFreshLo}{$-$0.3}  %
\newcommand{\librefThreeNewLibraryMinusFreshHi}{1.0}  %
\newcommand{\librefThreeMatched}{84.3}  %
\newcommand{\librefThreeFamiliarMinusMatched}{0.4}  %
\newcommand{\librefThreeFamiliarMinusMatchedLo}{$-$0.3}  %
\newcommand{\librefThreeFamiliarMinusMatchedHi}{1.1}  %
\newcommand{\librefThreeMatchedMinusFresh}{0.3}  %
\newcommand{\librefThreeMatchedMinusFreshLo}{$-$0.4}  %
\newcommand{\librefThreeMatchedMinusFreshHi}{1.0}  %
\newcommand{\librefThreePatternSize}{21.4}  %
\newcommand{\librefThreeSmallPatterns}{70}  %
\newcommand{\librefBlindFreshTwo}{86.40}  %
\newcommand{\librefFreshPatternSize}{20.0}  %
\newcommand{\librefFreshSmallPatterns}{135}  %
\newcommand{\librefMinReuseMinusBlind}{1.6}  %
\newcommand{\librefMaxReuseMinusBlind}{2.1}  %
\newcommand{\librefMinReuseMinusNew}{1.6}  %
\newcommand{\librefMaxReuseMinusNew}{2.1}  %
\newcommand{\reuseRunOneAtNewPatternEpoch}{85.5}  %
\newcommand{\reuseRunOneNewPatternMinusMatchedTwo}{1.23}  %
\newcommand{\reuseRunOneMaxAfterWarmupTwo}{86.76}  %
\newcommand{\reuseRunTwoAtNewPatternEpoch}{86.4}  %
\newcommand{\reuseRunTwoNewPatternMinusMatchedTwo}{0.41}  %
\newcommand{\reuseRunTwoMaxAfterWarmupTwo}{86.79}  %
\newcommand{\reuseRunThreeAtNewPatternEpoch}{86.2}  %
\newcommand{\reuseRunThreeNewPatternMinusMatchedTwo}{0.54}  %
\newcommand{\reuseRunThreeMaxAfterWarmupTwo}{86.77}  %
\newcommand{\reuseTestGraphsPerPattern}{5}  %
\newcommand{\regMatchedGap}{1.0}  %
\newcommand{\regMatchedWithin}{0.5}  %
\newcommand{\matchedNewPatternAtReuseEpochTwo}{86.16}  %
\newcommand{\matchedNewPatternLateMeanTwo}{86.39}  %
\newcommand{\matchedEpochsAfterWarmup}{190}  %
\newcommand{\matchedEpochsBelow}{184}  %
\newcommand{\matchedGapAtReuseEpoch}{1.6}  %
\newcommand{\fiveSpearmanDeficit}{0.9}  %
\newcommand{\fivePpgtOfficialSpread}{1.2}  %
\newcommand{\fiveGrassOverPublished}{1.2}  %
\newcommand{\fivePpgtOverPublished}{1.0}  %
\newcommand{\heldSecondHMinusKept}{$-$4.4}  %
\newcommand{\heldSecondHMinusKeptLo}{$-$5.1}  %
\newcommand{\heldSecondHMinusKeptHi}{$-$3.8}  %
\newcommand{\heldSecondSeedBHMinusKept}{$-$4.0}  %
\newcommand{\heldSecondSeedBHMinusKeptLo}{$-$4.6}  %
\newcommand{\heldSecondSeedBHMinusKeptHi}{$-$3.4}  %
\newcommand{\sizematchGatedSeedAFamiliar}{86.6}  %
\newcommand{\sizematchGatedSeedAMatched}{86.3}  %
\newcommand{\sizematchGatedSeedAFresh}{86.2}  %
\newcommand{\sizematchGatedSeedAFamiliarMinusMatched}{0.3}  %
\newcommand{\sizematchGatedSeedAFamiliarMinusMatchedLo}{$-$0.3}  %
\newcommand{\sizematchGatedSeedAFamiliarMinusMatchedHi}{1.0}  %
\newcommand{\sizematchGatedSeedBFamiliar}{86.6}  %
\newcommand{\sizematchGatedSeedBMatched}{86.1}  %
\newcommand{\sizematchGatedSeedBFresh}{86.2}  %
\newcommand{\sizematchGatedSeedBFamiliarMinusMatched}{0.5}  %
\newcommand{\sizematchGatedSeedBFamiliarMinusMatchedLo}{$-$0.1}  %
\newcommand{\sizematchGatedSeedBFamiliarMinusMatchedHi}{1.2}  %
\newcommand{\sizematchGatedSeedCFamiliar}{86.5}  %
\newcommand{\sizematchGatedSeedCMatched}{86.0}  %
\newcommand{\sizematchGatedSeedCFresh}{86.1}  %
\newcommand{\sizematchGatedSeedCFamiliarMinusMatched}{0.5}  %
\newcommand{\sizematchGatedSeedCFamiliarMinusMatchedLo}{$-$0.2}  %
\newcommand{\sizematchGatedSeedCFamiliarMinusMatchedHi}{1.1}  %
\newcommand{\sizematchGatedSeedDFamiliar}{86.6}  %
\newcommand{\sizematchGatedSeedDMatched}{86.1}  %
\newcommand{\sizematchGatedSeedDFresh}{86.2}  %
\newcommand{\sizematchGatedSeedDFamiliarMinusMatched}{0.5}  %
\newcommand{\sizematchGatedSeedDFamiliarMinusMatchedLo}{$-$0.1}  %
\newcommand{\sizematchGatedSeedDFamiliarMinusMatchedHi}{1.2}  %
\newcommand{\sizematchGritSeedCFamiliar}{87.1}  %
\newcommand{\sizematchGritSeedCMatched}{85.8}  %
\newcommand{\sizematchGritSeedCFresh}{85.6}  %
\newcommand{\sizematchGritSeedCFamiliarMinusMatched}{1.3}  %
\newcommand{\sizematchGritSeedCFamiliarMinusMatchedLo}{0.6}  %
\newcommand{\sizematchGritSeedCFamiliarMinusMatchedHi}{2.0}  %
\newcommand{\sizematchGritSeedCFamiliarMinusMatchedTwo}{1.30}  %
\newcommand{\sizematchGritSeedCFamiliarMinusMatchedLoTwo}{0.64}  %
\newcommand{\sizematchGritSeedCFamiliarMinusMatchedHiTwo}{1.98}  %
\newcommand{\sizematchGrassSeedAFamiliar}{90.2}  %
\newcommand{\sizematchGrassSeedAMatched}{81.9}  %
\newcommand{\sizematchGrassSeedAFresh}{81.4}  %
\newcommand{\sizematchGrassSeedAFamiliarMinusMatched}{8.3}  %
\newcommand{\sizematchGrassSeedAFamiliarMinusMatchedLo}{7.5}  %
\newcommand{\sizematchGrassSeedAFamiliarMinusMatchedHi}{9.1}  %
\newcommand{\sizematchGrassSeedBFamiliar}{90.2}  %
\newcommand{\sizematchGrassSeedBMatched}{82.0}  %
\newcommand{\sizematchGrassSeedBFresh}{82.0}  %
\newcommand{\sizematchGrassSeedBFamiliarMinusMatched}{8.2}  %
\newcommand{\sizematchGrassSeedBFamiliarMinusMatchedLo}{7.4}  %
\newcommand{\sizematchGrassSeedBFamiliarMinusMatchedHi}{8.9}  %
\newcommand{\sizematchGrassSeedCFamiliar}{90.3}  %
\newcommand{\sizematchGrassSeedCMatched}{82.3}  %
\newcommand{\sizematchGrassSeedCFresh}{82.3}  %
\newcommand{\sizematchGrassSeedCFamiliarMinusMatched}{8.0}  %
\newcommand{\sizematchGrassSeedCFamiliarMinusMatchedLo}{7.2}  %
\newcommand{\sizematchGrassSeedCFamiliarMinusMatchedHi}{8.8}  %
\newcommand{\sizematchPpgtSeedAFamiliar}{90.0}  %
\newcommand{\sizematchPpgtSeedAMatched}{83.8}  %
\newcommand{\sizematchPpgtSeedAFresh}{83.6}  %
\newcommand{\sizematchPpgtSeedAFamiliarMinusMatched}{6.2}  %
\newcommand{\sizematchPpgtSeedAFamiliarMinusMatchedLo}{5.5}  %
\newcommand{\sizematchPpgtSeedAFamiliarMinusMatchedHi}{6.9}  %
\newcommand{\sizematchPpgtSeedBFamiliar}{90.6}  %
\newcommand{\sizematchPpgtSeedBMatched}{82.0}  %
\newcommand{\sizematchPpgtSeedBFresh}{81.7}  %
\newcommand{\sizematchPpgtSeedBFamiliarMinusMatched}{8.5}  %
\newcommand{\sizematchPpgtSeedBFamiliarMinusMatchedLo}{7.8}  %
\newcommand{\sizematchPpgtSeedBFamiliarMinusMatchedHi}{9.3}  %
\newcommand{\sizematchPpgtSeedCFamiliar}{90.7}  %
\newcommand{\sizematchPpgtSeedCMatched}{82.1}  %
\newcommand{\sizematchPpgtSeedCFresh}{82.0}  %
\newcommand{\sizematchPpgtSeedCFamiliarMinusMatched}{8.7}  %
\newcommand{\sizematchPpgtSeedCFamiliarMinusMatchedLo}{7.9}  %
\newcommand{\sizematchPpgtSeedCFamiliarMinusMatchedHi}{9.4}  %
\newcommand{\sizematchPpgtSeedDFamiliar}{90.9}  %
\newcommand{\sizematchPpgtSeedDMatched}{82.0}  %
\newcommand{\sizematchPpgtSeedDFresh}{82.2}  %
\newcommand{\sizematchPpgtSeedDFamiliarMinusMatched}{8.9}  %
\newcommand{\sizematchPpgtSeedDFamiliarMinusMatchedLo}{8.1}  %
\newcommand{\sizematchPpgtSeedDFamiliarMinusMatchedHi}{9.7}  %
\newcommand{\sizematchCkgSeedCFamiliar}{88.2}  %
\newcommand{\sizematchCkgSeedCMatched}{85.0}  %
\newcommand{\sizematchCkgSeedCFresh}{84.7}  %
\newcommand{\sizematchCkgSeedCFamiliarMinusMatched}{3.2}  %
\newcommand{\sizematchCkgSeedCFamiliarMinusMatchedLo}{2.5}  %
\newcommand{\sizematchCkgSeedCFamiliarMinusMatchedHi}{3.9}  %
\newcommand{\sizematchCkgSeedCBigFamiliar}{88.1}  %
\newcommand{\sizematchCkgSeedCBigMatched}{84.6}  %
\newcommand{\sizematchCkgSeedCBigFresh}{84.1}  %
\newcommand{\sizematchCkgSeedCBigFamiliarMinusMatched}{3.5}  %
\newcommand{\sizematchCkgSeedCBigFamiliarMinusMatchedLo}{2.7}  %
\newcommand{\sizematchCkgSeedCBigFamiliarMinusMatchedHi}{4.3}  %
\newcommand{\sizematchCkgNewSeedCFamiliar}{86.6}  %
\newcommand{\sizematchCkgNewSeedCMatched}{86.5}  %
\newcommand{\sizematchCkgNewSeedCFresh}{86.5}  %
\newcommand{\sizematchCkgNewSeedCFamiliarMinusMatched}{0.0}  %
\newcommand{\sizematchCkgNewSeedCFamiliarMinusMatchedLo}{$-$0.6}  %
\newcommand{\sizematchCkgNewSeedCFamiliarMinusMatchedHi}{0.7}  %
\newcommand{\sizematchCkgNewSeedCAllFamiliarMinusMatched}{0.1}  %
\newcommand{\sizematchGatedMean}{0.4}  %
\newcommand{\sizematchGritMean}{1.1}  %
\newcommand{\sizematchCkgMean}{3.4}  %
\newcommand{\sizematchGrassMean}{8.1}  %
\newcommand{\sizematchPpgtMean}{8.1}  %
\newcommand{\sizematchGrassDifferingPercent}{0.04}  %
\newcommand{\sizematchGrassPooledMoveBound}{0.01}  %
\newcommand{\convLookupPooled}{96.6}  %
\newcommand{\convLookupBatch}{96.3}  %
\newcommand{\convLookupPlain}{97.8}  %
\newcommand{\convLookupPlainMinusPooled}{1.2}  %
\newcommand{\convLookupPooledMinusBatch}{0.3}  %
\newcommand{\convMaxPlainMinusPooled}{1.2}  %
\newcommand{\convMaxPooledMinusBatch}{0.3}  %
\newcommand{\fiveAboveCountWord}{four}  %
\newcommand{\fiveGatedMeanBelowDensity}{0.2}  %
\newcommand{\aucTableCkgNewSeedANumber}{0}  %
\newcommand{\aucTableCkgNewSeedBNumber}{1}  %
\newcommand{\aucTableCkgNewSeedCNumber}{2}  %
\newcommand{\aucTableGatedSeedAOfficial}{94.3}  %
\newcommand{\aucTableGatedSeedAFamiliar}{94.2}  %
\newcommand{\aucTableGatedSeedAFresh}{93.8}  %
\newcommand{\aucTableGatedSeedBOfficial}{94.3}  %
\newcommand{\aucTableGatedSeedBFamiliar}{94.2}  %
\newcommand{\aucTableGatedSeedBFresh}{93.8}  %
\newcommand{\aucTableGatedSeedCOfficial}{94.2}  %
\newcommand{\aucTableGatedSeedCFamiliar}{94.1}  %
\newcommand{\aucTableGatedSeedCFresh}{93.7}  %
\newcommand{\aucTableGatedSeedDOfficial}{94.2}  %
\newcommand{\aucTableGatedSeedDFamiliar}{94.2}  %
\newcommand{\aucTableGatedSeedDFresh}{93.8}  %
\newcommand{\aucTableGritSeedAOfficial}{94.5}  %
\newcommand{\aucTableGritSeedAFamiliar}{94.5}  %
\newcommand{\aucTableGritSeedAFresh}{93.5}  %
\newcommand{\aucTableGritSeedBOfficial}{94.5}  %
\newcommand{\aucTableGritSeedBFamiliar}{94.5}  %
\newcommand{\aucTableGritSeedBFresh}{93.5}  %
\newcommand{\aucTableGritSeedCOfficial}{94.6}  %
\newcommand{\aucTableGritSeedCFamiliar}{94.5}  %
\newcommand{\aucTableGritSeedCFresh}{93.4}  %
\newcommand{\aucTableCkgSeedAOfficial}{95.7}  %
\newcommand{\aucTableCkgSeedAFamiliar}{95.7}  %
\newcommand{\aucTableCkgSeedAFresh}{92.3}  %
\newcommand{\aucTableCkgSeedBOfficial}{95.3}  %
\newcommand{\aucTableCkgSeedBFamiliar}{95.1}  %
\newcommand{\aucTableCkgSeedBFresh}{92.2}  %
\newcommand{\aucTableCkgSeedCOfficial}{95.4}  %
\newcommand{\aucTableCkgSeedCFamiliar}{95.3}  %
\newcommand{\aucTableCkgSeedCFresh}{92.6}  %
\newcommand{\aucTableGrassSeedAOfficial}{96.4}  %
\newcommand{\aucTableGrassSeedAFamiliar}{96.4}  %
\newcommand{\aucTableGrassSeedAFresh}{89.2}  %
\newcommand{\aucTableGrassSeedBOfficial}{96.6}  %
\newcommand{\aucTableGrassSeedBFamiliar}{96.5}  %
\newcommand{\aucTableGrassSeedBFresh}{89.9}  %
\newcommand{\aucTableGrassSeedCOfficial}{96.7}  %
\newcommand{\aucTableGrassSeedCFamiliar}{96.6}  %
\newcommand{\aucTableGrassSeedCFresh}{89.7}  %
\newcommand{\aucTablePpgtSeedAOfficial}{96.5}  %
\newcommand{\aucTablePpgtSeedAFamiliar}{96.5}  %
\newcommand{\aucTablePpgtSeedAFresh}{91.5}  %
\newcommand{\aucTablePpgtSeedBOfficial}{97.1}  %
\newcommand{\aucTablePpgtSeedBFamiliar}{97.0}  %
\newcommand{\aucTablePpgtSeedBFresh}{90.5}  %
\newcommand{\aucTablePpgtSeedCOfficial}{97.1}  %
\newcommand{\aucTablePpgtSeedCFamiliar}{97.0}  %
\newcommand{\aucTablePpgtSeedCFresh}{90.6}  %
\newcommand{\aucTablePpgtSeedDOfficial}{97.3}  %
\newcommand{\aucTablePpgtSeedDFamiliar}{97.1}  %
\newcommand{\aucTablePpgtSeedDFresh}{90.6}  %
\newcommand{\aucTableCkgNewSeedAOfficial}{94.3}  %
\newcommand{\aucTableCkgNewSeedAFamiliar}{94.2}  %
\newcommand{\aucTableCkgNewSeedAFresh}{94.0}  %
\newcommand{\aucTableCkgNewSeedBOfficial}{94.3}  %
\newcommand{\aucTableCkgNewSeedBFamiliar}{94.2}  %
\newcommand{\aucTableCkgNewSeedBFresh}{94.0}  %
\newcommand{\aucTableCkgNewSeedCOfficial}{94.3}  %
\newcommand{\aucTableCkgNewSeedCFamiliar}{94.3}  %
\newcommand{\aucTableCkgNewSeedCFresh}{94.0}  %
\newcommand{\aucTableBlindOfficial}{94.3}  %
\newcommand{\aucTableBlindFamiliar}{94.2}  %
\newcommand{\aucTableBlindFresh}{94.0}  %
\newcommand{\aucTableFirstBlindFresh}{93.6}  %
\newcommand{\aucTableBlindDrop}{0.22}  %
\newcommand{\aucTableCkgSeedADrop}{3.38}  %
\newcommand{\aucTableCkgSeedBDrop}{2.93}  %
\newcommand{\aucTableCkgSeedCDrop}{2.67}  %
\newcommand{\aucTableCkgNewSeedADrop}{0.20}  %
\newcommand{\aucTableCkgNewSeedBDrop}{0.20}  %
\newcommand{\aucTableCkgNewSeedCDrop}{0.23}  %
\newcommand{\aucTableGatedSeedADrop}{0.40}  %
\newcommand{\aucTableGatedSeedBDrop}{0.47}  %
\newcommand{\aucTableGatedSeedCDrop}{0.42}  %
\newcommand{\aucTableGatedSeedDDrop}{0.40}  %
\newcommand{\aucTableGrassSeedADrop}{7.26}  %
\newcommand{\aucTableGrassSeedBDrop}{6.56}  %
\newcommand{\aucTableGrassSeedCDrop}{6.89}  %
\newcommand{\aucTableGritSeedADrop}{0.98}  %
\newcommand{\aucTableGritSeedBDrop}{1.04}  %
\newcommand{\aucTableGritSeedCDrop}{1.14}  %
\newcommand{\aucTablePpgtSeedADrop}{4.92}  %
\newcommand{\aucTablePpgtSeedBDrop}{6.57}  %
\newcommand{\aucTablePpgtSeedCDrop}{6.37}  %
\newcommand{\aucTablePpgtSeedDDrop}{6.46}  %
\newcommand{\gpuSeedCTrainHours}{19.1}  %
\newcommand{\gpuSeedCOriginalPooled}{88.4}  %
\newcommand{\gpuSeedCOriginalLo}{88.2}  %
\newcommand{\gpuSeedCOriginalHi}{88.7}  %
\newcommand{\gpuSeedCFamiliarPooled}{88.2}  %
\newcommand{\gpuSeedCFamiliarLo}{87.7}  %
\newcommand{\gpuSeedCFamiliarHi}{88.7}  %
\newcommand{\gpuSeedCFreshPooled}{84.7}  %
\newcommand{\gpuSeedCFreshLo}{84.1}  %
\newcommand{\gpuSeedCFreshHi}{85.1}  %
\newcommand{\gpuSeedCFamiliarMinusOriginal}{$-$0.3}  %
\newcommand{\gpuSeedCFreshMinusFamiliar}{$-$3.5}  %
\newcommand{\gpuSeedCFreshMinusFamiliarLo}{$-$4.2}  %
\newcommand{\gpuSeedCFreshMinusFamiliarHi}{$-$2.9}  %
\newcommand{\gpuSeedCMinusDensity}{$-$1.7}  %
\newcommand{\gpuSeedCMinusDensityLo}{$-$2.1}  %
\newcommand{\gpuSeedCMinusDensityHi}{$-$1.5}  %
\newcommand{\gpuSeedCFamiliarSmall}{86.4}  %
\newcommand{\gpuSeedCFreshSmall}{84.4}  %
\newcommand{\gpuSeedCFamiliarBig}{88.1}  %
\newcommand{\gpuSeedCFreshBig}{84.1}  %
\newcommand{\blindThresholdRound}{0.18}  %
\newcommand{\estPatternClassShareRound}{0.18}  %
\newcommand{\grassOwnEvaluationGapBound}{0.01}  %
\newcommand{\costClusterProcs}{4}  %
\newcommand{\costClusterSplit}{8.0}  %
\newcommand{\costClusterGraph}{29}  %
\newcommand{\costMinesExactSplit}{11}  %
\newcommand{\costMinesExactProcs}{4}  %
\newcommand{\costMinesBpSplit}{32}  %
\newcommand{\matchedNewPatternAtReuseEpochBelowDensity}{0.6}  %
\newcommand{\librefMinFamiliarMinusFresh}{0.3}  %
\newcommand{\librefMaxFamiliarMinusFresh}{0.7}  %
\newcommand{\hvCompletedCkgSeedAOfficialAccuracyRound}{86.8}  %
\newcommand{\hvCompletedCkgSeedBOfficialAccuracyRound}{86.8}  %
\newcommand{\hvCompletedCkgSeedCOfficialAccuracyRound}{86.8}  %
\newcommand{\ladderMarginDependentMin}{0.6}  %
\newcommand{\ladderMarginDependentMax}{4.4}  %
\newcommand{\ladderLogisticTwoHopFamiliarMinusFresh}{0.0}  %
\newcommand{\lookupFamiliarMinusFresh}{46.3}  %
\newcommand{\ladderAboveGatedMean}{0.1}  %
\newcommand{\heldBelowDensityMin}{2.4}  %
\newcommand{\heldBelowDensityMax}{2.8}  %
\newcommand{\heldBelowFullMin}{3.3}  %
\newcommand{\heldBelowFullMax}{5.1}  %
\newcommand{\newPatternAboveDensityMin}{0.04}  %
\newcommand{\newPatternAboveDensityMax}{0.11}  %
\newcommand{\reuseAtNewEpochBelowMin}{0.4}  %
\newcommand{\reuseAtNewEpochBelowMax}{1.2}  %
\newcommand{\lookupAbovePubRound}{7}  %
\newcommand{\fiveCkgBelowDensityMin}{1.7}  %
\newcommand{\fiveCkgBelowDensityMax}{2.1}  %
\newcommand{\fiveGatedSeedsBelowDensityWord}{three}  %
\newcommand{\reviewGritSeedsDifferNodes}{7,005}  %
\newcommand{\reviewGritSeedsDifferPct}{2.9}  %
\newcommand{\reviewCkgNewSeedsDifferNodes}{3,056}  %
\newcommand{\reviewCkgNewSeedsDifferPct}{1.3}  %
\newcommand{\reviewOfficialTestNodes}{238,225}  %
\newcommand{\minesExactQueuedWord}{four}  %
\newcommand{\minesExactChainsWord}{eight}  %

\title{PATTERN Can Be Solved from Memory: Testing Graph Neural Networks on Unseen Patterns}

\author{\name Amin Najafgholizadeh\thanks{University of Zanjan. Emails: amin.ngz@znu.ac.ir, afsharchi@znu.ac.ir.}
      \qquad Mohsen Afsharchi\footnotemark[1]}

\def\month{MM}
\def\year{YYYY}
\def\openreview{\url{https://openreview.net/forum?id=XXXX}}

\begin{document}

\maketitle

\begin{abstract}
PATTERN asks graph neural networks to find a planted subgraph in a new host graph. The hosts change, but the benchmark draws just \recDistinctPatterns{} patterns and reuses them across training and test splits. A lookup that extracts these patterns from the training labels achieves \lookupRealPooled{}\% class-weighted accuracy on the official test split, above every published result, without fitting a model or consulting the generator. On unseen patterns, the same lookup scores \lookupFreshPooled{}\%. A matcher that also uses the generator reaches \realMatcherPooled{}\% on the official split. Among five graph networks we retrained, GRASS and PPGT go from the top two on the official split to the bottom two on unseen patterns. Each loses \sizematchGrassMean{} percentage points even when familiar and unseen patterns are matched in size. Logistic regression on two-hop node statistics scores above every run of the five networks on unseen patterns. For CKGConv, the advantage follows the training patterns: it reappears with each of three independently drawn pattern libraries and disappears when every training and validation graph gets a new pattern. Generator-based inference also scores \clusterCeilingPooled{}\% on CLUSTER, compared with a published best of \pubClusterBest{}\%. On Minesweeper's released board, \minesExactRanWord{} of \minesExactChainsWord{} planned chains produced samples within the time budget; none converged, but their pooled predictions reached \minesExactAuc{}\% ROC AUC, above the published best of \pubMinesBest{}\%. PATTERN's official split tests its intended task of finding familiar patterns in new graphs. We recommend reporting the same checkpoint's score on unseen patterns alongside it.
\end{abstract}

\section{Introduction}
\label{sec:intro}

PATTERN is one of the synthetic node-classification benchmarks that papers use to compare graph networks \citep{dwivedi2023benchmarking}. Each of its graphs is built in two steps: a host graph is drawn from a stochastic block model, and a smaller random graph, the pattern, is planted in it and joined to the host's nodes. A model must say which nodes belong to the pattern. Model papers read their scores on such tasks as evidence of what an architecture can capture; GraphGPS, for example, presents its results as ``showcasing the ability to perform very well on a variety of synthetic tasks designed to test the model expressivity'' \citep{rampasek2022gps}. At least \appGPapers{} papers from 2024 to 2026 report results on PATTERN (Appendix~\ref{app:papers}). Its generator, however, draws \recDistinctPatterns{} patterns once and plants the same ones in the training, validation and test splits. What does a PATTERN score measure under this reuse: the ability to find a planted structure (detection), or recall of the structures seen in training?

\begin{figure}[t]
\centering
\includegraphics[width=\linewidth]{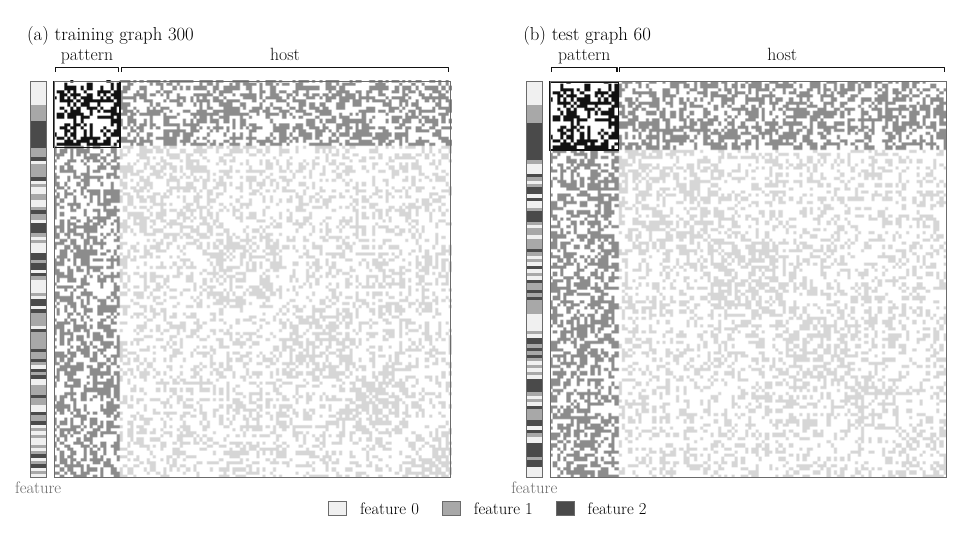}
\caption{The same pattern in a training graph and a test graph of PATTERN's released files (training graph \reuseTrainGraph{} and test graph \reuseTestGraph{}; the pattern has \reusePatternSize{} nodes, the size the benchmark's paper gives). Each adjacency matrix lists the pattern's nodes first, in the same canonical order, followed by the host's \reuseTrainHost{} and \reuseTestHost{} nodes grouped by spectral clustering. Edges within the pattern are black, edges between pattern and host dark gray, and edges within the host light gray; the strip shows node features. The pattern's edges and features are identical in the two graphs, while the host and the edges joining it to the pattern are redrawn. This pattern occurs in \recGraphsPerPattern{} training graphs.}
\label{fig:reuse}
\end{figure}

The task comes from the subgraph-matching experiment of \citet{scarselli2009gnn}, in which one randomly generated subgraph was inserted into training and test graphs alike, and the benchmark paper describes it as finding ``a fixed graph pattern P embedded in larger graphs G of variable sizes'' \citep{dwivedi2023benchmarking}. Each pattern is planted in \recGraphsPerPattern{} training graphs and in \recTestGraphsPerPattern{} graphs of each of the validation and test splits, and only the host and the edges joining it to the pattern are drawn anew (Figure~\ref{fig:reuse} and Table~\ref{tab:recurrence}). We call a structure that a generator draws once and plants in many graphs a template; PATTERN's templates are its patterns. The reuse fits the stated task. It also means that a high score does not show that a model can find a pattern it has not seen, and that is the reading this paper examines.

We read model scores against four reference scores, each computed from what the benchmark itself provides (Table~\ref{tab:inputs}). Two use only the benchmark's own data, as a trained model does: logistic regression on node statistics, which ignores the patterns, and a lookup of the training patterns. Two also take values from the generator's code: a pattern-blind predictor that infers the pattern's nodes without knowing which patterns were drawn, and a matcher that knows them. They are yardsticks for reading a model's score, like an estimate of the Bayes rate in a simulation study. A reference score is computed once for a benchmark (Appendix~\ref{app:cost} gives the costs). Our findings concern PATTERN; CLUSTER and Minesweeper, whose generators plant no templates, serve as contrast cases.

\begin{table}[t]
\centering
\caption{The four reference scores on PATTERN's official test split, by what they use. Each is the class-weighted accuracy of a stated procedure, so the best score attainable from the same information is at least as high, up to the sampling error of the test split. Benchmark data means the training split, plus the validation split for the regression's threshold, as for a trained model's checkpoint. The pattern-blind predictor uses the generator's edge probabilities and block sizes, and the matcher every value the generator's code sets (Table~\ref{tab:generators}). No method reads test labels.}
\label{tab:inputs}
\small
\begin{tabular}{>{\raggedright\arraybackslash}p{0.25\linewidth}>{\raggedright\arraybackslash}p{0.33\linewidth}>{\raggedright\arraybackslash}p{0.33\linewidth}}
\toprule
 & Without the patterns & With the patterns \\
\midrule
Benchmark data only & Two-hop logistic regression, \ladderLogisticTwoHopRealPooled{} (Section~\ref{sec:blind}) & Lookup of the training patterns, \lookupRealPooled{} (Section~\ref{sec:recall}) \\
\addlinespace
Benchmark data and the generator's code & Pattern-blind predictor, \realDensityPooled{} (Section~\ref{sec:blind}) & Likelihood matcher, \realMatcherPooled{} (Section~\ref{sec:recall}) \\
\bottomrule
\end{tabular}
\end{table}

The test split can be solved from memory (Section~\ref{sec:recall}). The lookup fits nothing and uses nothing from the generator, yet it reaches \lookupRealPooled{} class-weighted accuracy, above every published score and about \lookupAbovePubRound{} points above the best model in Table~\ref{tab:ladder}.

Without the patterns, the reference scores lie within a point of most published ones (Section~\ref{sec:blind}). Logistic regression on two-hop statistics reaches \ladderLogisticTwoHopRealPooled{} and the pattern-blind predictor \realDensityPooled{}, and the published scores up to 2023 in Table~\ref{tab:ladder} lie at most \realGRITAboveDensity{} points above the predictor. A node's degree and its graph's size alone give \ladderLogisticDegreeRealPooled{}.

On new patterns the official order of the five trained models we tested turns over, and \fiveAboveCountWord{} of them depend on the training patterns (Section~\ref{sec:fresh}). We drew new graphs from a re-implementation of the generator, carrying either the training patterns (the familiar set) or newly drawn ones (the fresh set, and a size-matched version of it), and trained five published architectures with their released configurations: GatedGCN-PE \citep{dwivedi2023benchmarking}, GRIT \citep{ma2023grit}, CKGConv \citep{ma2024ckgconv}, GRASS \citep{liao2025grass} and PPGT \citep{ma2026ppgt}. Every model scores about as well on the familiar set as on the official test split. GRASS and PPGT, the two highest-scoring published models we compare with, hold the top two places on the official test split and the bottom two on fresh patterns, and each falls \sizematchGrassMean{} points from familiar patterns to fresh ones of the same sizes (Figure~\ref{fig:reversal}). The \fiveAboveCountWord{} models that score above the pattern-blind predictor on the official test split fall below it on fresh patterns in every seed, and the falls track each model's excess over the predictor on the official test split (rank correlation \fiveSpearmanDeficit{}, Figure~\ref{fig:excessdrop}). GRIT depends on its training patterns least. GatedGCN-PE, which scores at the predictor on the official test split, shows no detectable dependence and scores highest of the five on fresh patterns. There the two-hop regression scores above every seed of all five networks, although it is fitted to the same training split and uses nothing from the generator.

In CKGConv the advantage follows the patterns it was trained on (Sections~\ref{sec:heldout} to~\ref{sec:reuse}). Trained without \heldPatterns{} of the \recDistinctPatterns{} released patterns, it falls below the pattern-blind predictor on the released test graphs that carry them. Trained with a new pattern in every training and validation graph, it scores at the predictor's level on fresh and reused patterns alike. Trained with its released recipe on our own graphs that reuse \reuseLibraryPatterns{} new patterns, it scores above the predictor on test graphs that carry those patterns and below it on fresh ones. Reuse across the training and validation graphs is therefore enough to produce the gap in CKGConv trained with its released recipe.

On CLUSTER and Minesweeper, reference scores computed from the generator without training match or beat the published results (Section~\ref{sec:ceilings}): CLUSTER's best lies within the interval of its reference score, and the Minesweeper reference score beats every published result. PATTERN's best published scores, in contrast, lie above the reference scores that leave out the shared patterns. Where no template is reused, no published score exceeds a reference score computed without training.

We propose a reporting protocol (Section~\ref{sec:practices}): scores on familiar and fresh patterns from the same checkpoint, reference scores beside them, and the metric's weighting and pooling. Appendix~\ref{app:registrations} sets what we expected of each experiment beside what happened.

\section{PATTERN and its generator}
\label{sec:generators}

What a model can carry from the training graphs to the test graphs depends on what the generator draws once and what it draws anew for every graph. PATTERN builds a graph from a host and a pattern \citep{dwivedi2023benchmarking}, with the parameters given in Appendix~\ref{app:generators}. The host is a stochastic block model with five blocks of random sizes, denser within blocks than across them, and its node features are drawn uniformly. The pattern is a random graph with node features of its own. Each pattern node joins each host node independently, with the host's within-block probability. The nodes are then shuffled, and the label marks membership in the pattern. The benchmark paper states ``We randomly generate 100 patterns P'' in its Appendix~C.6. Its code draws their edges and features once and plants each pattern in \recGraphsPerPattern{} training graphs, \recValGraphsPerPattern{} validation graphs, and \recTestGraphsPerPattern{} test graphs. Exphormer also describes ``100 possible predetermined subgraph patterns'' \citep{shirzad2023exphormer} and does not discuss what the reuse means for transfer. The released patterns vary in size, although the benchmark paper gives a single size; a maintainer disclosed this change in 2020 \citep{issue28}.

We use class-weighted accuracy, the mean over the two classes of the share of each class labeled correctly, pooled over all nodes of a test split. The benchmark's original code averages this metric over batches; GraphGPS, GRIT, CKGConv, and GRASS pool the split \citep{rampasek2022gps,ma2023grit,ma2024ckgconv,liao2025grass}. Appendix~\ref{app:metric} compares the two conventions with plain node accuracy.

\section{PATTERN's test split can be solved from memory}
\label{sec:recall}

Every test pattern already occurs in the training split. For every graph we take the subgraph induced by the pattern labels, with its node features, and compute its canonical form, which does not depend on the order of the nodes (Appendix~\ref{app:generators}). The released training split contains exactly \recDistinctPatterns{} forms, each repeated \recGraphsPerPattern{} times. Every validation and test pattern is one of them, and each form appears in \recTestGraphsPerPattern{} test graphs. A pattern's features are drawn once and planted with its edges, so they are part of its identity, although a taxonomy of graph benchmarks lists PATTERN's features as irrelevant to its task \citep{liu2022taxonomy}.

A lookup reads this set of patterns off the training labels and searches each test graph for induced copies with matching features, trying the patterns of \matcherBig{} nodes or more from the largest down and otherwise taking the largest copy of a smaller pattern (Appendix~\ref{app:withoutgen}). It fits nothing and uses no value from the generator's code. On the official test split it scores \lookupRealPooled{} class-weighted accuracy, above every published model (Table~\ref{tab:ladder}), and it recovers the planted pattern exactly in \lookupRealExactPctBig\% of the graphs whose pattern has \matcherBig{} nodes or more. Patterns that large almost never occur by chance: treating node pairs as independent, a typical host holds about \chanceCopiesBig{} chance copies of a given \matcherBig{}-node pattern, against about \chanceCopiesSmall{} of a \chanceKSmall{}-node pattern (Appendix~\ref{app:generators}). On graphs with new patterns the same lookup scores \lookupFreshPooled{}, about what labeling every node as host scores (Section~\ref{sec:fresh}), so its official score is recall of the training patterns.

Small patterns are where the lookup errs, and the generator can resolve them. Our likelihood matcher scores each copy by an approximate log-likelihood ratio of the hypothesis that the copy is the planted pattern against the hypothesis that every node belongs to the host, accounting for features, edges, the host's size, and placement (Appendix~\ref{app:likelihood}). It searches for large patterns as the lookup does and looks for smaller ones among a shortlist of nodes ranked by a version of the pattern-blind predictor (Section~\ref{sec:blind}). The matcher reaches \realMatcherPooled{}. We fixed its settings before scoring it on the official test split (Appendix~\ref{app:likelihood}); on the validation split, which played no part in fixing them, it scores \valMatcherPooled{}. A version of the lookup that seeks the smaller patterns among each graph's nodes of highest degree, also without the generator, reaches \lookupDegreeRealPooled{} (Appendix~\ref{app:withoutgen}).

\begin{table}[t]
\centering
\caption{PATTERN's official test split: selected published models, ordered by the year of the cited publication, and our reference scores. Appendix~\ref{app:papers} lists the papers from 2024 to 2026 that report PATTERN results. Published values are the authors' means over seeds; our intervals resample graphs. The benchmark's own baselines average the metric over batches, while the code of GraphGPS, GRIT, CKGConv and GRASS pools it over the split (Appendix~\ref{app:metric}). The regressions are fitted to the official training split, with the decision threshold chosen on the validation split; the lookup and the matcher have no trained parameters. QPE's \pubPatternGRITQiRWQPE{} \citep{thabet2024qpe} is left out (Appendix~\ref{app:metric}).}
\label{tab:ladder}
\small
\begin{tabular}{llr}
\toprule
Predictor & Reference & Class-weighted accuracy \\
\midrule
SAN & \citet{kreuzer2021san} & \pubPatternSAN{} \\
DGN & \citet{beaini2021dgn} & \pubPatternDGN{} \\
GraphGPS & \citet{rampasek2022gps} & \pubPatternGraphGPS{} \\
EGT & \citet{hussain2022egt} & \pubPatternEGT{} \\
k-subtree SAT & \citet{chen2022sat} & \pubPatternKSubtreeSAT{} \\
GCN-AK+ & \citet{zhao2022gnnak} & \pubPatternGCNAK{} \\
GRPE & \citet{park2022grpe} & \pubPatternGRPE{} \\
MLP & \citet{dwivedi2023benchmarking} & \pubPatternMLP{} \\
GatedGCN-PE & \citet{dwivedi2023benchmarking} & \pubPatternGatedGCNPE{} \\
GRIT & \citet{ma2023grit} & \pubPatternGRIT{} \\
Gradformer & \citet{liu2024gradformer} & \pubPatternGradformer{} \\
CKGConv & \citet{ma2024ckgconv} & \pubPatternCKGConv{} \\
GRASS & \citet{liao2025grass} & \pubPatternGRASS{} \\
PPGT & \citet{ma2026ppgt} & \pubPatternPPGT{} \\
\midrule
Logistic regression, degree and graph size & this work & \ladderLogisticDegreeRealPooled{} (\ladderLogisticDegreeRealLo{} to \ladderLogisticDegreeRealHi{}) \\
Logistic regression, two-hop statistics & this work & \ladderLogisticTwoHopRealPooled{} (\ladderLogisticTwoHopRealLo{} to \ladderLogisticTwoHopRealHi{}) \\
Pattern-blind predictor & this work & \realDensityPooled{} (\realDensityLo{} to \realDensityHi{}) \\
Lookup of the training patterns & this work & \lookupRealPooled{} (\lookupRealLo{} to \lookupRealHi{}) \\
Likelihood matcher & this work & \realMatcherPooled{} (\realMatcherLo{} to \realMatcherHi{}) \\
\bottomrule
\end{tabular}

\end{table}

\section{Transfer to new patterns}
\label{sec:measured}

A model that relies on the patterns it has seen should lose its advantage on new ones. We compare trained models with predictors that do not know the patterns, on the official test split and on three sets of \freshGraphs{} new graphs each, drawn with our re-implementation of the generator (Section~\ref{sec:fresh}). The familiar set carries the training patterns. In the fresh set, each graph carries a newly drawn pattern. The size-matched set also has newly drawn patterns, each as large as the pattern of the corresponding graph in the familiar set.

\subsection{Predictors that do not know the patterns}
\label{sec:blind}

Consider a predictor that knows every parameter of the generator but not which patterns were drawn; we call the best expected score of such a predictor the pattern-blind level. A fresh pattern's features follow the same law as the host's, so the features alone do not single out its nodes. In block terms the pattern is a sixth block: its nodes join each other and every host block with the within-block probability, while distinct host blocks are joined with a lower probability. We approximate the posterior over these six blocks with a Gibbs sampler started from a mean-field fit (Appendix~\ref{app:likelihood}); with the threshold below, this sampler is the pattern-blind predictor.

The metric sets the decision threshold. Write $p_v$ for a node's posterior probability of belonging to the pattern, $\hat y_v$ for its predicted label, and $n_1,n_0$ for the numbers of pattern and host nodes. Treating these counts as fixed, the expected class-weighted accuracy is
\begin{equation}
\frac12 \sum_v \left( \frac{p_v}{n_1}\,\hat y_v + \frac{1-p_v}{n_0}\,(1-\hat y_v) \right).
\label{eq:bayes}
\end{equation}
Each term of Equation~\eqref{eq:bayes} is largest when the node is labeled pattern exactly when $p_v>n_1/(n_0+n_1)$, the class-prior threshold for balanced accuracy \citep{menon2013statistical}. We use the training split's share of pattern nodes. A threshold of one half instead lowers the predictor's official score to \realDensityHalfPooled{} (Appendix~\ref{app:likelihood}).

The pattern-blind predictor scores \realDensityPooled{} on the official test split and \freshDensityPooled{} on the fresh set. The sampler reaches the same score from other starting points, and its posterior is close to calibrated (Appendix~\ref{app:likelihood}). Its posterior is sampled and its threshold treats the class counts as fixed, so its score is a lower estimate of the pattern-blind level. A model whose expected score lies below the predictor's therefore lies below the level as well, and each such comparison in this paper is conservative with respect to the approximation.

Simple node statistics give most of this score. A node's degree already separates the classes: a pattern node joins every host node with the host's within-block probability, while a host node joins most of the host at the lower rate across blocks. We fitted logistic regression and boosted trees on the official training split to node statistics computed from the edges alone, from degree up to two-hop neighborhoods (Appendix~\ref{app:withoutgen}). Logistic regression on a node's degree and the graph's size scores \ladderLogisticDegreeRealPooled{} on the official test split, and two-hop statistics bring it to \ladderLogisticTwoHopRealPooled{}. These predictors show no detectable use of pattern identity: each scores about as well on the fresh set as on the official test split.

On the fresh set the pattern-blind predictor, which knows the generator's values and fits nothing, and logistic regression on two-hop statistics, fitted to the official training split, lie about a tenth of a point apart, and no node-statistics predictor scores above the predictor there. The agreement indicates that the predictor lies near the pattern-blind level. With every value estimated from the training split instead of read from the generator's code, the predictor's score changes by about a tenth of a point (Appendix~\ref{app:withoutgen}), so a predictor that must learn the generator's values from data, as a graph network must, can score about where this one does, given the generator's form.

The published scores up to 2023 in Table~\ref{tab:ladder} lie at most \realGRITAboveDensity{} points above the pattern-blind predictor's score on the official test split, and the three highest lie \realCKGConvAboveDensity{} to \realPPGTAboveDensity{} points above it (Table~\ref{tab:ladder}). Whether a model uses the patterns shows on patterns it has not seen.

\subsection{Five networks on new patterns}
\label{sec:fresh}

For patterns that no method has seen, we re-implement the generator from the benchmark's released code and draw the fresh set with it, with pattern sizes uniform over the range of the released patterns. The released patterns' sizes, features, internal density and rate of edges to the host are consistent with the generator's laws (Appendix~\ref{app:fresh}). The familiar set comes from the same code, with a training pattern in each graph, so the two sets differ only in their planted patterns.

We train five published architectures, GatedGCN-PE, GRIT, CKGConv, GRASS and PPGT, with their released PATTERN configurations and three or four seeds each, and choose each checkpoint by the model's released validation rule. Appendix~\ref{app:fresh} gives the training details of every model.

Two controls tie the regenerated graphs to the benchmark. Every method scores about as well on the familiar set as on the official test split (Table~\ref{tab:fivemodels}), so the regenerated hosts behave like the released ones. Graphs of the official test split passed through our conversion and CKGConv's encoding also receive exactly the predictions they receive through the original loader. All methods are scored on the same fresh graphs, so comparisons between methods are paired; comparisons between the familiar and fresh sets are not.

\begin{table}[t]
\centering
\caption{Five models under the fresh-pattern test, each trained on PATTERN's training split with its released configuration: class-weighted accuracy and differences, as means over the seeds in the second column (Table~\ref{tab:fivemodels_seeds} gives every seed and the spread over seeds). Familiar minus fresh compares different graphs and is not paired. Familiar minus size-matched is paired by position, graph $i$ of each set carrying a pattern of the same size. The last column is the difference from the pattern-blind predictor on the same fresh graphs, paired by graph. Rows are in order of official score, and differences are computed before rounding. The last three rows are reference scores: the two-hop regression is fitted to the same training split and uses nothing from the generator, and the lookup reads the training patterns (Section~\ref{sec:recall}).}
\label{tab:fivemodels}
\small
\begin{tabular}{lccccccc}
\toprule
 & & Official & Familiar & Fresh & Familiar & Familiar minus & Fresh minus \\
Model & Seeds & test split & set & set & minus fresh & size-matched & predictor \\
\midrule
GatedGCN-PE & \fiveGatedSeeds{} & \fiveGatedMeanOfficial{} & \fiveGatedMeanFamiliar{} & \fiveGatedMeanFresh{} & \fiveGatedMeanDrop{} & \sizematchGatedMean{} & \fiveGatedMeanFreshMinusDensity{} \\
GRIT & \fiveGritSeeds{} & \fiveGritMeanOfficial{} & \fiveGritMeanFamiliar{} & \fiveGritMeanFresh{} & \fiveGritMeanDrop{} & \sizematchGritMean{} & \fiveGritMeanFreshMinusDensity{} \\
CKGConv & \fiveCkgSeeds{} & \fiveCkgMeanOfficial{} & \fiveCkgMeanFamiliar{} & \fiveCkgMeanFresh{} & \fiveCkgMeanDrop{} & \sizematchCkgMean{} & \fiveCkgMeanFreshMinusDensity{} \\
GRASS & \fiveGrassSeeds{} & \fiveGrassMeanOfficial{} & \fiveGrassMeanFamiliar{} & \fiveGrassMeanFresh{} & \fiveGrassMeanDrop{} & \sizematchGrassMean{} & \fiveGrassMeanFreshMinusDensity{} \\
PPGT & \fivePpgtSeeds{} & \fivePpgtMeanOfficial{} & \fivePpgtMeanFamiliar{} & \fivePpgtMeanFresh{} & \fivePpgtMeanDrop{} & \sizematchPpgtMean{} & \fivePpgtMeanFreshMinusDensity{} \\
\midrule
Pattern-blind predictor & & \realDensityPooled{} & \familiarDensityPooled{} & \freshDensityPooled{} & \fiveDensityDrop{} & \sizematchDensityFamiliarMinusMatched{} & \\
Two-hop logistic regression & & \ladderLogisticTwoHopRealPooled{} & \ladderLogisticTwoHopFamiliarPooled{} & \ladderLogisticTwoHopFreshPooled{} & \ladderLogisticTwoHopFamiliarMinusFresh{} & \sizematchLadderFamiliarMinusMatched{} & \\
Lookup of the training patterns & & \lookupRealPooled{} & \lookupFamiliarPooled{} & \lookupFreshPooled{} & \lookupFamiliarMinusFresh{} & & \lookupFreshMinusDensity{} \\
\bottomrule
\end{tabular}

\end{table}

\begin{figure}[t]
\centering
\includegraphics[width=0.78\linewidth]{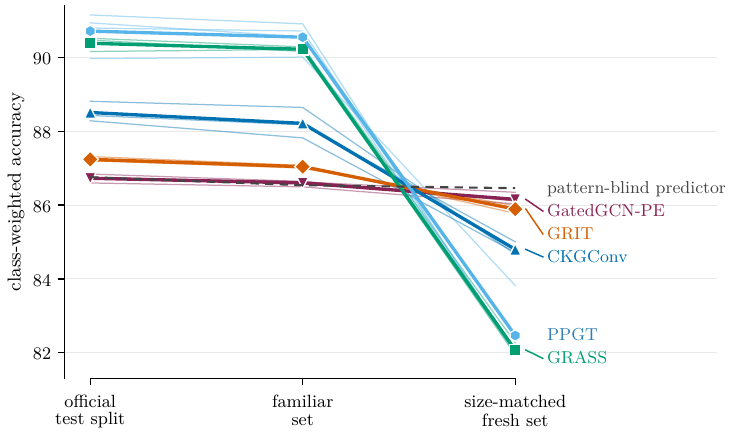}
\caption{GRASS and PPGT lead the five networks on the official test split and trail the other three on size-matched fresh patterns, each falling \sizematchGrassMean{} points from the familiar set. The size-matched set's new patterns have the familiar set's pattern sizes. Thick lines join each network's means over its seeds, and thin lines its single seeds; the dashed line is the pattern-blind predictor, which GatedGCN-PE's line covers on the first two sets. The two-hop regression lies close to the predictor on every set and is not drawn. Table~\ref{tab:fivemodels_seeds} gives the official scores, and Table~\ref{tab:sizematch} the familiar and size-matched ones.}
\label{fig:reversal}
\end{figure}

On fresh patterns the official order of the five networks turns over (Table~\ref{tab:fivemodels} and Figure~\ref{fig:reversal}). GRASS \citep{liao2025grass} and PPGT \citep{ma2026ppgt}, which hold the two highest scores in Table~\ref{tab:ladder}, take the top two places on the official test split and the bottom two on the fresh set. GatedGCN-PE, the benchmark's own message-passing model, goes from last to first, and CKGConv stays in the middle. GRASS and PPGT keep their scores on the familiar set and fall the furthest of the five on fresh patterns, \sizematchGrassMean{} points each even when the fresh patterns match the familiar ones in size. Every seed of both lies below the pattern-blind predictor there, with paired intervals below zero. Our runs of the two score about a point above their published means, and the PPGT seed closest to its published score also falls below the predictor on fresh patterns (Appendix~\ref{app:fresh}). The lookup marks the other end of the table: it keeps its score on the familiar set and falls to about what labeling every node as host scores on the fresh set, where it can label only chance copies of known patterns.

The fresh and familiar sets also differ in the distribution of pattern sizes. A size-matched set of \sizematchGraphs{} graphs with new patterns removes that difference, and every predictor, trained or not, scores on it within about half a point of its fresh score (Table~\ref{tab:sizematch}). Matching pattern sizes therefore changes any model's fall by at most about half a point.

A model that never saw the familiar set's patterns loses little between the familiar and fresh sets. CKGConv trained with its released recipe on our own graphs that reuse other patterns (the reuse models of Section~\ref{sec:reuse}) loses \librefMinFamiliarMinusFresh{} to \librefMaxFamiliarMinusFresh{} points, and its paired falls to the size-matched set have intervals that contain zero (Table~\ref{tab:libref}). We read a model as depending on its training patterns when it falls from the familiar to the fresh set by more than the reuse models do, about as much on the size-matched set, and lands below the pattern-blind predictor on fresh patterns. GRASS and PPGT meet this criterion by a wide margin.

\paragraph{CKGConv.} Our three seeds of CKGConv score close to its published result on the official test split (Table~\ref{tab:ckgseeds}), and all three fall below the pattern-blind predictor on fresh patterns, by \fiveCkgBelowDensityMin{} to \fiveCkgBelowDensityMax{} points, with paired intervals below zero. Neither the decision threshold nor the size of the patterns explains this: its AUC on fresh patterns, which no threshold affects, lies below the predictor's in every seed (Table~\ref{tab:auc}), and it lies below the predictor among small and large patterns alike (Table~\ref{tab:ckgstrata}). From the familiar to the fresh set it loses \fiveCkgMeanDrop{} points on average, against at most \librefMaxFamiliarMinusFresh{} for a reuse model, so CKGConv depends on its training patterns as well.

\paragraph{GRIT.} Trained with its released configuration for \fiveGritSeedsWord{} seeds, GRIT reproduces its published official score, which lies \realGRITAboveDensity{} points above the pattern-blind predictor's (Appendix~\ref{app:fresh}). It loses \fiveGritMeanDrop{} points from the familiar to the fresh set and about as much on the size-matched set, more in every seed than any reuse model loses; every seed falls below the predictor on fresh patterns with paired intervals below zero, and its fresh AUC lies below the predictor's in every seed (Tables~\ref{tab:fivemodels_seeds} and~\ref{tab:auc}). GRIT therefore depends on its training patterns too, though least of the four.

\paragraph{GatedGCN-PE.} Trained with its released configuration for \fiveGatedSeedsWord{} seeds, GatedGCN-PE scores on the official test split as the pattern-blind predictor does. It lies \fiveGatedMeanBelowDensity{} points below the predictor on fresh patterns, with paired intervals below zero in \fiveGatedSeedsBelowDensityWord{} of its \fiveGatedSeedsWord{} seeds (Table~\ref{tab:fivemodels_seeds}), so all five networks fall short of the predictor there. Its fall from the familiar to the fresh set, \fiveGatedMeanDrop{} points, lies within that of the reuse models, and its paired falls to the size-matched set have intervals that contain zero in every seed. With no official excess to lose, GatedGCN-PE shows no detectable dependence on the training patterns.

\begin{figure}[t]
\centering
\includegraphics[width=\linewidth]{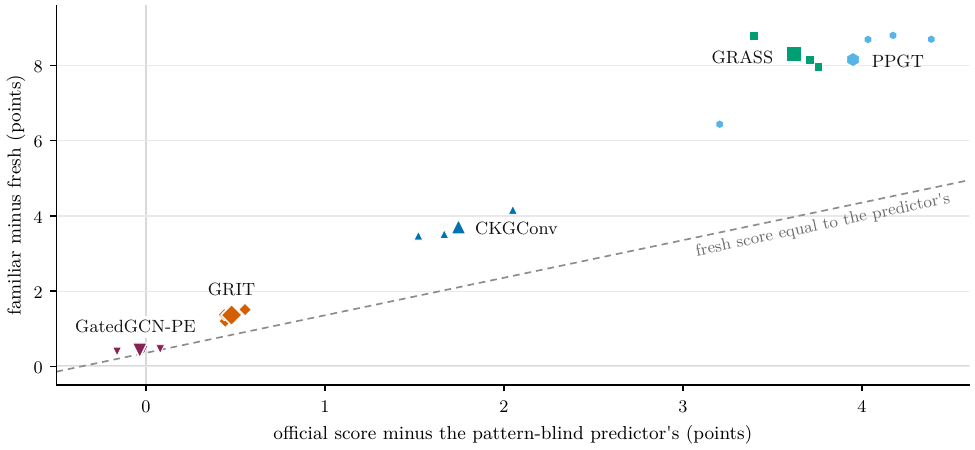}
\caption{Each model's official score minus the pattern-blind predictor's (horizontal) against its fall from the familiar to the fresh set (vertical), in points of class-weighted accuracy. A model whose familiar score equals its official score and whose fresh score equals the predictor's lies on the dashed line; a model's height above the line is about how far it falls below the predictor on fresh patterns. Large markers are means over seeds and small markers single seeds (Tables~\ref{tab:fivemodels} and~\ref{tab:fivemodels_seeds}).}
\label{fig:excessdrop}
\end{figure}

A model's fall should be read against its official score. Since every model scores about as well on the familiar set as on the official test split, its height above the dashed line of Figure~\ref{fig:excessdrop} is about how far it falls below the predictor on fresh patterns. GatedGCN-PE lies near the line and the other four above it, and a model's fall below the predictor tracks its excess over the predictor on the official test split (Table~\ref{tab:fivemodels}; rank correlation \fiveSpearmanDeficit{} over the five models, descriptive only).

The same reading holds without the generator. Logistic regression on two-hop statistics, fitted to the same training split as the networks, scores \ladderLogisticTwoHopFreshPooled{} on the fresh set, above every seed of the five networks: \ladderMarginDependentMin{} to \ladderMarginDependentMax{} points above the means of the four networks that depend on the training patterns, and \ladderAboveGatedMean{} above GatedGCN-PE's (Tables~\ref{tab:fivemodels} and~\ref{tab:fivemodels_seeds}). On new patterns, a linear model of node statistics fitted to the same training split outscores the five networks we tested.

\subsection{Patterns held out of training}
\label{sec:heldout}

Withholding patterns from training tests the same dependence on the released graphs alone, with no regenerated graph. We hold out \heldPatterns{} of the \recDistinctPatterns{} patterns, chosen within size groups, train CKGConv on the remaining graphs, and select its checkpoint on the remaining validation graphs. On the official test graphs that carry a held-out pattern it scores \heldBelowFullMin{} to \heldBelowFullMax{} points below CKGConv trained on all the patterns with the same seed, and \heldBelowDensityMin{} to \heldBelowDensityMax{} points below the pattern-blind predictor, in both seeds and both draws of held-out patterns, with paired intervals below zero (Table~\ref{tab:heldout}; Figure~\ref{fig:fresh} in Appendix~\ref{app:fresh} plots the first draw beside the predictor). In the first draw the loss lies mostly on patterns of \matcherBig{} nodes or more, and accuracy on the kept patterns rises slightly. Predictors on node statistics, retrained without the same patterns, score as well on the graphs that carry them as on the others (Appendix~\ref{app:withoutgen}).

\begin{figure}[t]
\centering
\includegraphics[width=\linewidth]{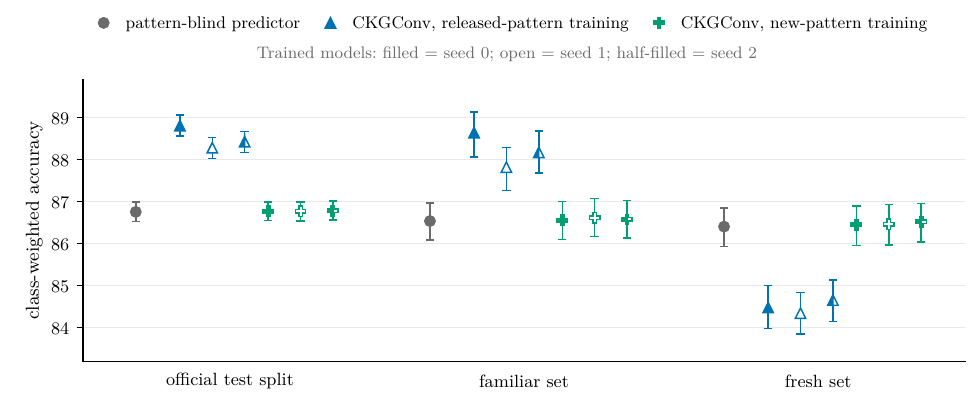}
\caption{CKGConv trained on the released patterns (triangles) or with a new pattern per training and validation graph (plus signs), and the pattern-blind predictor (circles), on the same graph sets. Bars are 95\% graph-bootstrap intervals for each method and seed, not intervals across seeds. Table~\ref{tab:ckgfreshcompleted} gives the values.}
\label{fig:freshtraining}
\end{figure}

\subsection{Training with a new pattern per graph}
\label{sec:freshtraining}

Training on many distinct patterns might remove the dependence. We keep CKGConv's architecture and optimization recipe but draw a new pattern for every training and validation graph, distinct from one another, from the released patterns and from the fresh test patterns, and select the checkpoint on the new validation set (Appendix~\ref{app:fresh}). We call the result the new-pattern model.

On the fresh set, accuracy rises by about \freshTrainGainRound{} points in each of three seeds and reaches the pattern-blind predictor's score, \newPatternAboveDensityMin{} to \newPatternAboveDensityMax{} points above it with paired intervals that contain zero (Table~\ref{tab:ckgfreshcompleted} and Figure~\ref{fig:freshtraining}). On the official test split and the familiar set it scores at the predictor's level too: the new-pattern model gives up the advantage on reused patterns and scores as the predictor does on every set. Selection on the new validation patterns chose earlier checkpoints than the original recipe, so part of the gain may come from shorter training (Section~\ref{sec:reuse}).

Three routes that rest on different information thus reach about the same score on fresh patterns: the pattern-blind predictor, the two-hop regression and the new-pattern model.

The new protocol changes the distinctness of the patterns, their size and feature makeup, and the validation set together; Section~\ref{sec:reuse} isolates reuse.

\subsection{Reusing a library of new patterns}
\label{sec:reuse}

Repeating PATTERN's design with our own generator tests whether reuse is enough to produce the dependence, without anything particular to the released data. We draw three libraries of \reuseLibraryPatterns{} patterns under the fresh set's laws, and for each library the code that drew the new-pattern training graphs (Appendix~\ref{app:fresh}) draws training, validation and test graphs that take their patterns from the library as PATTERN's generator does, so each pattern occurs in \recGraphsPerPattern{} training graphs. No library pattern occurs in the released data, the fresh or size-matched sets, the new-pattern training, or another library. CKGConv is trained on each library with its released recipe, one seed per library, and its checkpoint is chosen on the library's validation graphs (Appendix~\ref{app:fresh}). The data differ from the new-pattern training with the same seed only in that the training and validation graphs reuse the library's patterns, and because the released rule selects on these graphs, it also picks a later checkpoint; we call these models the reuse models. The official test split and the familiar set carry patterns these models have never seen.

\begin{table}[htb]
\centering
\caption{CKGConv trained with its released recipe on graphs that reuse a library of \reuseLibraryPatterns{} new patterns, one library per seed. The library set holds \reuseTestGraphs{} test graphs that carry the library's patterns; the official test split and the familiar set carry the released patterns, which these models never saw. The last three columns give the difference from the pattern-blind predictor on the same graphs, paired by graph, with 95\% graph-bootstrap intervals. Table~\ref{tab:reusedetail} compares these models with the other CKGConv models, and Table~\ref{tab:libref} gives the predictor's scores on the library sets.}
\label{tab:reuse}
\small
\setlength{\tabcolsep}{3.5pt}
\resizebox{\linewidth}{!}{\begin{tabular}{lccccccc}
\toprule
 & Library & Fresh & Official & Familiar & \multicolumn{3}{c}{Minus the pattern-blind predictor, paired} \\
\cmidrule(lr){6-8}
Seed & set & set & test split & set & Library set & Fresh set & Official test split \\
\midrule
\reuseRunOneSeed{} & \reuseRunOneLibrary{} & \reuseRunOneFresh{} & \reuseRunOneOfficial{} & \reuseRunOneFamiliar{} & \librefOneReuseMinusBlind{} (\librefOneReuseMinusBlindLo{} to \librefOneReuseMinusBlindHi{}) & \reuseRunOneFreshMinusDensity{} (\reuseRunOneFreshMinusDensityLo{} to \reuseRunOneFreshMinusDensityHi{}) & \reuseRunOneOfficialMinusDensity{} (\reuseRunOneOfficialMinusDensityLo{} to \reuseRunOneOfficialMinusDensityHi{}) \\
\reuseRunTwoSeed{} & \reuseRunTwoLibrary{} & \reuseRunTwoFresh{} & \reuseRunTwoOfficial{} & \reuseRunTwoFamiliar{} & \librefTwoReuseMinusBlind{} (\librefTwoReuseMinusBlindLo{} to \librefTwoReuseMinusBlindHi{}) & \reuseRunTwoFreshMinusDensity{} (\reuseRunTwoFreshMinusDensityLo{} to \reuseRunTwoFreshMinusDensityHi{}) & \reuseRunTwoOfficialMinusDensity{} (\reuseRunTwoOfficialMinusDensityLo{} to \reuseRunTwoOfficialMinusDensityHi{}) \\
\reuseRunThreeSeed{} & \reuseRunThreeLibrary{} & \reuseRunThreeFresh{} & \reuseRunThreeOfficial{} & \reuseRunThreeFamiliar{} & \librefThreeReuseMinusBlind{} (\librefThreeReuseMinusBlindLo{} to \librefThreeReuseMinusBlindHi{}) & \reuseRunThreeFreshMinusDensity{} (\reuseRunThreeFreshMinusDensityLo{} to \reuseRunThreeFreshMinusDensityHi{}) & \reuseRunThreeOfficialMinusDensity{} (\reuseRunThreeOfficialMinusDensityLo{} to \reuseRunThreeOfficialMinusDensityHi{}) \\
\bottomrule
\end{tabular}
}
\end{table}

Trained this way, CKGConv behaves as it does on the released benchmark (Table~\ref{tab:reuse}). On test graphs that carry their library's patterns the three models score \librefMinReuseMinusBlind{} to \librefMaxReuseMinusBlind{} points above the pattern-blind predictor, as CKGConv trained on the released patterns does on the official test split, while the new-pattern model scores at the predictor there (Table~\ref{tab:libref}). These differences are paired with the predictor on the same graphs, and the predictor itself scores only slightly higher on the library sets than on the fresh set (Table~\ref{tab:libref}). On fresh patterns, and on the official test split, whose patterns they have not seen, the reuse models fall below the predictor, with paired intervals below zero, and on the fresh set they score about as CKGConv trained on the released patterns does.

Selection on reused validation patterns chose much later checkpoints than selection on new ones, and late in training the new-pattern model also falls below the predictor on the official test split (Appendix~\ref{app:traininglength}). At the new-pattern models' own epochs, however, the reuse models already score \reuseAtNewEpochBelowMin{} to \reuseAtNewEpochBelowMax{} points below them on the official test split (Table~\ref{tab:reusedetail}), so the gap does not come from training length alone.

Reusing \reuseLibraryPatterns{} patterns across the training and validation graphs is enough to produce the gap between seen and new patterns in CKGConv trained with its released recipe, and the gap appears with our generator as it does with the released files. CKGConv's advantage on the official test split comes from having trained on its \recDistinctPatterns{} patterns: trained on other patterns, the same recipe falls below the pattern-blind predictor there.

\section{CLUSTER and Minesweeper}
\label{sec:ceilings}

CLUSTER and Minesweeper serve as contrast cases, since their generators plant no templates; Minesweeper's splits divide the cells of one board, but no structure is planted in more than one graph. Table~\ref{tab:ceilings} gives their reference scores.

\begin{table}[t]
\centering
\caption{Reference scores built from the generators of CLUSTER and Minesweeper. CLUSTER uses pooled class-weighted accuracy with a graph-bootstrap interval; chain agreement is, for each chain after the first, the share of labels that agree with the earlier chains pooled, and the lowest share is shown. Minesweeper uses mean ROC AUC and standard deviation across the ten masks of its one board. Inference from the features alone and inference conditioned on training labels use different information. Belief propagation stopped at its sweep limit, and neither it nor the samplers are known to have converged.}
\label{tab:ceilings}
\small
\setlength{\tabcolsep}{3pt}
\begin{tabular}{l>{\raggedright\arraybackslash}p{0.12\linewidth}>{\raggedright\arraybackslash}p{0.13\linewidth}r>{\raggedright\arraybackslash}p{0.1\linewidth}>{\raggedright\arraybackslash}p{0.065\linewidth}>{\raggedright\arraybackslash}p{0.3\linewidth}}
\toprule
Benchmark & Information used & Method & Estimate & 95\% interval & SD over masks & Diagnostics \\
\midrule
CLUSTER & edges, labeled nodes & Gibbs sampler, block swaps, \clusterChains{} chains & \clusterCeilingPooled{} & \clusterCeilingLo{} to \clusterCeilingHi{} & & chain agreement \clusterChainAgreementMedian{} (median graph), \clusterChainAgreementMin{} (worst graph) \\
 & & best published & \pubClusterBest{} & & & GRPE-Small \citep{park2022grpe}, SD \pubClusterBestSd{} over its seeds; GRIT \pubClusterGRIT{}, GRASS \pubClusterGRASS{}, PPGT \pubClusterPPGT{} \\
\midrule
Minesweeper & features & belief propagation & \minesFeaturesAuc{} & & \minesFeaturesSd{} & stopped at its limit of \minesBpIterations{} sweeps, last change \minesBpLastChange{} \\
 & features & strip sampler & \minesExactAuc{} & & \minesExactSd{} & not converged; \minesExactRan{} of \minesExactChains{} chains ran, and all \minesExactValidChains{} reached a valid board; each alone \minesExactChainLo{} to \minesExactChainHi{} \\
 & features, training labels & belief propagation & \minesLabelsAuc{} & & \minesLabelsSd{} & stopped at its limit of \minesBpIterations{} sweeps on every split \\
 & & best published & \pubMinesBest{} & & \pubMinesBestSd{} & SIES \citep{chen2026sies} \\
 & & best in a peer-reviewed venue & \pubMinesPeerBest{} & & \pubMinesPeerBestSd{} & CSNN \citep{ribeiro2026csnn} \\
\bottomrule
\end{tabular}

\end{table}

CLUSTER draws each graph independently from a six-block stochastic block model \citep{dwivedi2023benchmarking}. In each block one randomly chosen node carries the block's index as its feature, and the other nodes carry an uninformative feature. The label is the block, scored with PATTERN's metric. Training and test graphs share a law but no templates, so a prediction rests on the test graph's edges and its six revealed labels. We sample the posterior over assignments of nodes to the six blocks given the edges and the revealed labels, under the generator's edge probabilities, size prior and rule for revealing labels (Appendix~\ref{app:cluster}), and label each node by its largest sampled marginal, a rule not tuned to the metric, so that the score is a lower estimate of what the same information supports. This scores \clusterCeilingPooled{} (95\% interval \clusterCeilingLo{} to \clusterCeilingHi{}), and the best published score, \pubClusterBest{}, lies within that interval. The chains agree poorly on some graphs, so whether CLUSTER has headroom left remains open.

Minesweeper is a single graph, a square grid in which each cell is joined to its eight neighbors \citep{platonov2023critical}. A fixed number of randomly chosen cells hold mines. Randomly selected cells show the number of neighboring mines, capped, as a one-hot feature; the other cells show that their count is unknown. Ten official masks split the cells of this one board into training, validation, and test sets, and the metric is ROC AUC averaged over the masks. A supervised model can use the board's training labels through its learned weights, even though the labels are not among its inputs.

Inference on Minesweeper keeps only the boards that reproduce every revealed count (Appendix~\ref{app:mines} discusses its cost). We replace the fixed total number of mines with independent mines at the generator's rate (Appendix~\ref{app:mines}) and sample the posterior by redrawing narrow strips of rows or columns from their exact conditionals. Averaging the conditional probabilities over updates and chains gives \minesExactAuc{} ROC AUC from the revealed counts alone, without training, against \pubMinesBest{} for the best published result we found \citep{chen2026sies} and \pubMinesPeerBest{} for the best in a peer-reviewed venue \citep{ribeiro2026csnn}. Each chain alone scores above the peer-reviewed result (Table~\ref{tab:ceilings}). The chains had not converged when the time budget fixed in advance ran out, and only \minesExactRanWord{} of the \minesExactChainsWord{} chains we set up had run, since the code ran one chain per processor core on a machine with \minesExactRanWord{} (Appendix~\ref{app:mines}). Their averaged probabilities are computed from the revealed counts and the generator alone, so the score they reach is a lower estimate of what that information supports on the released board, however far they lie from the exact posterior. Belief propagation conditioned on each mask's training labels also scores above every published result; from the features alone it does not.

On these two benchmarks no published score exceeds what inference under the generator reaches without training. On PATTERN they exceed the reference scores that leave out the patterns, and on fresh patterns that excess turns into a deficit, because it rests on the reused patterns (Section~\ref{sec:measured}).

\section{A reporting protocol}
\label{sec:practices}

\paragraph{Reporting a model on PATTERN.} We propose that a paper reporting PATTERN results give the following, all from the same trained checkpoint.
\begin{enumerate}
\item The official test score, with the metric named: class-weighted accuracy, pooled over the nodes of the split, or whatever weighting and pooling were used.
\item Scores on a familiar set, new graphs that carry the training patterns, and on a size-matched fresh set, new graphs with newly drawn patterns whose sizes match the familiar set's, both from the generator's code. The familiar score checks the regenerated graphs against the official ones, and the difference between the two measures dependence on the training patterns. Our familiar, fresh and size-matched sets are in the supplement.
\item Reference scores beside them: the lookup (\lookupRealPooled{}) as a floor for what recall of the training patterns reaches; two-hop logistic regression (\ladderLogisticTwoHopRealPooled{} on the official test split, \sizematchLadderMatched{} on our size-matched set), a baseline fitted to the same split without the generator; and the pattern-blind predictor (\realDensityPooled{} and \sizematchDensityMatched{}) for detection without the patterns.
\item The spread over training seeds, reported apart from the intervals from sampling graphs.
\end{enumerate}
A model whose fall to fresh patterns exceeds what the change of sets alone produces, on the fresh and the size-matched set alike, and that then lies below the pattern-blind predictor, rests on the training patterns (Section~\ref{sec:fresh}); a fresh score clearly above the pattern-blind predictor would show that the predictor falls short of the pattern-blind level. On the fresh set the predictor, the regression and the new-pattern model score within a few tenths of a point of one another, and no method we tested scores more than about a tenth of a point above the predictor. With PATTERN's random patterns, a fresh score shows whether a model relies on its training patterns. A paper that claims recognition of patterns absent from training should also train with a validation set of fresh patterns, since validation picks the checkpoint (Section~\ref{sec:freshtraining}). Where new graphs cannot be generated, holding patterns out of training and scoring the released graphs that carry them gives a weaker version of the same test (Section~\ref{sec:heldout}).

\paragraph{Building or revising a synthetic benchmark.} A benchmark whose generator is public can support the same reading from the start.
\begin{enumerate}
\item List what the generator draws once and shares across the splits: on PATTERN the patterns, with their edges and features; on CLUSTER only the generator's law; on Minesweeper one board, whose cells the splits divide.
\item Keep a test split of fresh templates beside any split that reuses them, with a validation set drawn the same way. A split that reuses templates can legitimately test matching on new hosts, the task PATTERN's original description sets; a claim about templates absent from training needs the second split.
\item Remove cues that make the task easier than intended. With fresh random patterns, PATTERN mainly tests detection of a planted block, and degree gives most of what detection reaches (Section~\ref{sec:blind}). Drawing the pattern's edges to the host at the host's mean pair probability ($\bar p$ in Appendix~\ref{app:generators}) could weaken that cue. Templates drawn from a structured family of subgraphs would make recognition of that family the task, and a fresh split from it could rank models as well as check them.
\item Publish reference scores computed from the generator, with what each one uses (the training templates, the generator's parameters, or the training labels as evidence) and with its errors of sampling, approximation and convergence stated separately. Achieved scores are lower bounds on the attainable score, never ceilings. They can show how much information a leaderboard leaves unused, as on Minesweeper, or leave open whether headroom remains, as on CLUSTER.
\item Name the metric's weighting and pooling. On fixed predictions, the conventions move our predictors' scores by up to \convMaxPlainMinusPooled{} points (Appendix~\ref{app:metric}). PATTERN papers use plain and class-weighted accuracy, pooled or averaged over batches, and most of the papers we surveyed do not name the weighting (Appendices~\ref{app:metric} and~\ref{app:papers}).
\end{enumerate}

\section{Related work}
\label{sec:related}

\paragraph{Synthetic benchmarks and their critics.} PATTERN and CLUSTER came with a framework that fixes budgets and protocols \citep{dwivedi2023benchmarking}; Minesweeper comes from a collection on which standard graph networks challenged architectures designed for heterophily \citep{platonov2023critical}. GraphWorld compares models on many generated datasets in place of a few fixed ones \citep{palowitch2022graphworld}. Tuned classic message-passing networks match the graph transformers they compare with on PATTERN and CLUSTER \citep{luo2025classic}, and on several datasets of the Long Range Graph Benchmark the reported advantage of graph transformers vanishes after basic tuning \citep{tonshoff2024gap}. A position paper calls synthetic graph tasks saturated \citep{bechler2025position}, and a graph-alignment benchmark criticizes PATTERN for resting on block-model topologies \citep{lagesse2026alignment}. A rewiring study finds that no method it tests improves Minesweeper, calls its grid a computational graph whose connectivity carries no information about the labels, and questions it as a node-classification benchmark \citep{linkerhagner2025jdr}. None of these critiques measures transfer to unseen templates. On PATTERN, a vision transformer with frame averaging has been reported to memorize the frames of training graphs \citep{kim2023probabilistic}, and models to overfit in ways attributed to node positions \citep{soga2023bridging}; neither study examines reuse across splits. A question in the benchmark's repository asks why the \recDistinctPatterns{} patterns are treated as a single label \citep{issue66}.

\paragraph{Inference in planted models.} Recovery in block models has sharp asymptotic thresholds, studied through belief propagation \citep{decelle2011asymptotic,abbe2018community}, and revealed labels shift them \citep{zhang2014phase}. \citet{duranthon2024optimal} conjecture an optimal belief-propagation algorithm for the contextual block model and find gaps to published networks. For sparse contextual block models, the locally Bayes-optimal node classifier can be implemented as a message-passing network \citep{baranwal2023optimality}, and the maximum a posteriori rule contains the kind of non-linear neighbor aggregation that graph networks use \citep{wei2022nonlinearity}. Recovering a planted subgraph in Erd\H{o}s--R\'enyi noise also has sharp thresholds \citep{mossel2023sharp}. These results do not give exact optima for small finite graphs with revealed labels, or for PATTERN's reused patterns; our inference on finite graphs gives approximate reference scores.

\paragraph{Expressivity bounds.} Graphtester uses Weisfeiler--Leman indistinguishability to bound the performance of architecture classes on graph datasets, PATTERN and CLUSTER among them \citep{akbiyik2023graphtester}. Such bounds ask what an architecture can distinguish. A model that can distinguish a familiar template may use it at test time without generalizing to an unseen one. We ask what the data protocol makes available and how models transfer when its templates change.

\paragraph{Shortcuts and reused data.} Regularities in a benchmark can support shortcuts in place of the skill it is meant to test \citep{geirhos2020shortcut}. Isomorphic copies across training and test graphs can inflate graph-classification scores \citep{ivanov2019isomorphism}, and duplicate images \citep{barz2020cifar} and familiar question-answer pairs \citep{lewis2021question} have similar effects. New test sets generated with an existing benchmark's recipe are one way to test transfer \citep{recht2019imagenet}. Our fresh set applies this to subgraphs that are reused by design.

\paragraph{Training-free baselines.} Fixed aggregation followed by a tabular predictor can match graph networks on many node benchmarks, with Minesweeper among the exceptions \citep{rubio2026fixed}. On PATTERN's fresh patterns a logistic regression on node statistics outscores the networks we tested, and on Minesweeper the public generator is a different resource: sampling its posterior from the features alone, without training, scores above every published result on the released board.

\section{Limitations}
\label{sec:limitations}

We test five architectures with \fiveGrassSeedsWord{} or \fiveGatedSeedsWord{} seeds each. Every seed of the four affected models shows the effect, but so few seeds measure its spread across training runs only roughly (Table~\ref{tab:fivemodels_seeds}). The five do not cover the field evenly: GRIT, CKGConv and PPGT come from one line of work, and PPGT shares CKGConv's code; GraphGPS and a tuned classic message-passing network \citep{luo2025classic} are not among them. Our GRASS and PPGT runs score about a point above the published means, and PPGT's seeds vary more than its paper reports, so the published checkpoints might fall by a different amount; the PPGT seed whose official score is closest to the published one still falls \fivePpgtSeedADrop{} points (Appendix~\ref{app:fresh}). The graph-bootstrap intervals hold the trained models fixed; they do not measure variation across seeds or across sets of training patterns, and they are not corrected for multiple comparisons. The three libraries of Section~\ref{sec:reuse} give a view of the variation across sets of training patterns for CKGConv.

The familiar and fresh sets differ in more than familiarity: the fresh set draws a separate pattern for each graph, while the familiar set reuses a fixed set. For the reuse models, trained on other patterns, we detect no effect of that difference, though one of up to about a point is not excluded (Table~\ref{tab:libref}). Training length matters too: late in training, a new-pattern model falls about \matchedNewPatternAtReuseEpochBelowDensity{} points below the pattern-blind predictor on the official test split (Appendix~\ref{app:traininglength}, one seed). GRIT's falls exceed those of every reuse model in every seed, though we did not test that difference formally. Each design changes more than one thing, as Sections~\ref{sec:fresh} to~\ref{sec:reuse} state, but they depart from the benchmark in different ways: the held-out test uses no regenerated graphs, and the reuse models keep the code and recipe of the new-pattern training. Together they show that CKGConv depends on the patterns it was trained on, and that a stated change of training protocol brings it to the predictor's score on fresh patterns; the fresh-pattern test alone shows the same dependence for GRASS, PPGT and GRIT. They do not split a model's accuracy into recall and detection, and they do not identify a mechanism inside the networks.

The reference scores are approximate (Sections~\ref{sec:blind} and~\ref{sec:ceilings}). Some CLUSTER chains mix poorly. Only \minesExactRanWord{} of the \minesExactChainsWord{} Minesweeper chains we set up ran (Appendix~\ref{app:mines}); they did not converge, and they use a simplified prior. Their pooled score lies above every single chain, likely because the chains sit in different parts of the posterior, and more chains could move it either way. Neither the intervals nor the spread across masks covers these errors of approximation, and Minesweeper's one board does not establish performance on new boards.

\section{Conclusion}
\label{sec:conclusion}

We asked what a PATTERN score measures. The test split reuses the \recDistinctPatterns{} training patterns, and a lookup that uses only the training split finds them well enough to score \lookupRealPooled{} class-weighted accuracy, above every published model. Without the patterns, logistic regression on node statistics and inference under the generator score within a point of most published models, and node degree and graph size alone give \ladderLogisticDegreeRealPooled{}. Of the five models we tested, \fiveAboveCountWord{} score above the pattern-blind predictor on the official test split, and that excess does not carry over to fresh patterns, where each falls below the predictor. The order of the five turns over: GRASS and PPGT hold the top two places on the official test split and the bottom two on fresh patterns, where the regression scores above every seed of all five. Trained on our own graphs that reuse \reuseLibraryPatterns{} new patterns across the training and validation splits, CKGConv shows the same gap, so such reuse, with the released rule for choosing the checkpoint, is enough to produce it; trained with a new pattern in every graph, it reaches the predictor's score on fresh patterns, so the detection task that a fresh split poses is learnable. On CLUSTER and Minesweeper, which reuse no planted structure, approximate inference under the generator without training matches the best published CLUSTER result and beats every published Minesweeper result. On Minesweeper this score comes from a partial run: a \minesExactRanWord{}-core machine ran \minesExactRanWord{} of the \minesExactChainsWord{} planned chains within the time budget fixed in advance, and they stopped before converging. On PATTERN the official score measures how well a model finds the training patterns in new hosts, and a score on fresh patterns from the same checkpoint, reported beside it, shows whether that performance carries over to patterns the model has not seen.

\section*{Code and data}

The anonymous supplement contains the code of every experiment, the raw outputs its analysis reads, the identities of the held-out patterns, and our familiar, fresh and size-matched sets, identical to those the experiments used, with the code that regenerates them on a CPU. Its analysis code computes every number and table in the paper, and the figures that plot scores, from those outputs. Its reproduction script rescores the saved per-graph predictions of every seed of GatedGCN-PE, GRASS and PPGT and of two seeds each of GRIT and of CKGConv trained with a new pattern per graph, reads the recorded summaries of the other evaluations of trained models, and checks every recomputed value against the printed one. Retraining needs a GPU and the upstream repositories at the revisions the code names. The trained weights will be released with the paper.

\bibliography{refs}
\bibliographystyle{tmlr}

\appendix

\section{The generators and the count of patterns}
\label{app:generators}

Our samplers and our regenerated graphs follow each generator's steps with the parameters its code sets \citep{dwivedi2023benchmarking,platonov2023critical}, which Table~\ref{tab:generators} lists. We loaded PATTERN and CLUSTER through PyTorch Geometric's conversion of the benchmark's released files \citep{fey2019pyg}, and Minesweeper from the file its authors released. For PATTERN we also loaded the benchmark's original release with its own loader, and every graph of the three splits matches PyTorch Geometric's conversion in nodes, features, edges, and labels.

To count patterns, we took the subgraph induced by the nodes labeled as pattern in every graph of every split, colored each of its nodes by its feature, and computed a canonical form of the colored graph with BLISS \citep{junttila2007bliss} through igraph \citep{csardi2006igraph}. Two colored graphs have the same canonical form exactly when they are isomorphic with their colors, so counting forms counts patterns whatever the order in which their nodes are stored (Table~\ref{tab:recurrence}). The count uses only node features, edges, and labels, and the labels of the validation and test splits serve only to extract their patterns.

\paragraph{Chance copies.} We estimate the expected number of copies of a given pattern of $m$ nodes, with its features, among a test graph's host nodes, treating node pairs as independent. A host node has a given feature with probability $1/c$, where $c$ is the number of feature values. Under the generator's law of block sizes, a share $s = \chanceSameBlock{}$ of host node pairs lie in one block, so two host nodes are joined with probability $\bar p = s\,p_{\mathrm{in}} + (1-s)\,p_{\mathrm{out}} = \chanceEdgeProb{}$, where $p_{\mathrm{in}}$ and $p_{\mathrm{out}}$ are the host's edge probabilities within and across blocks (Table~\ref{tab:generators}). For a pattern with half of its node pairs joined, as its generator's edge probability gives on average, and a host of the generator's mean size, $N = \chanceHostMean{}$ nodes, the expected number of copies is
\begin{equation*}
N(N-1)\cdots(N-m+1)\; c^{-m}\; \bigl(\bar p\,(1-\bar p)\bigr)^{m(m-1)/4},
\end{equation*}
about \chanceCopiesSmall{} for $m = \chanceKSmall{}$ and \chanceCopiesBig{} for $m = \matcherBig{}$. The estimate is rough, since the pairs of a block model are not independent and a pattern's edges vary around half of its pairs, but it shows why the matcher can stop at the first copy it finds of a pattern of \matcherBig{} nodes or more and must weigh every copy of a smaller pattern (Appendix~\ref{app:likelihood}). Its two main inputs can also be read off the training split without the generator's code: there, host nodes are joined at a rate of \estHostPairDensity{} (Appendix~\ref{app:withoutgen}), and the features take three values.

\begin{table}[h]
\centering
\caption{What each generator draws, with the parameters as its code sets them.}
\label{tab:generators}
\small
\begin{tabular}{l>{\raggedright\arraybackslash}p{0.78\linewidth}}
\toprule
Benchmark & What the generator draws \\
\midrule
PATTERN & Host: a stochastic block model with \genPatternHostBlocks{} blocks of sizes drawn uniformly from \genPatternBlockMin{} to \genPatternBlockMax{}, edge probability \genPatternPIn{} within a block and \genPatternPOut{} across blocks; features uniform on three values. Pattern: a random graph of a size drawn uniformly from \genPatternSizeMin{} to \genPatternSizeMax{} nodes, with edge probability \genPatternPInternal{} and uniform features, drawn once for each of the \recDistinctPatterns{} patterns; each pattern node joined to each host node with probability \genPatternPHost{}. Released pattern sizes: \recSizeMin{} to \recSizeMax{} nodes. \\
CLUSTER & A stochastic block model with \genClusterBlocks{} blocks of sizes drawn uniformly from \genClusterBlockMin{} to \genClusterBlockMax{}, edge probability \genClusterPIn{} within a block and \genClusterPOut{} across; one node per block, chosen uniformly, carries the block's index as its feature, every other node a zero. \\
Minesweeper & One square grid of side \minesBoardSide{} with each cell joined to its eight neighbors; mines on a fixed share \minesPrior{} of the cells, placed uniformly at random; a random \minesRevealedPct{}\% of the cells show the number of mined neighbors, capped at \minesCap{}, as a one-hot feature, and the rest show a flag for an unknown count. \\
\bottomrule
\end{tabular}

\end{table}

\begin{table}[h]
\centering
\caption{PATTERN's patterns in the released files, counted by canonical form with node features.}
\label{tab:recurrence}
\small
\begin{tabular}{lr}
\toprule
Quantity & Value \\
\midrule
Training graphs & \recTrainGraphs{} \\
Distinct planted patterns in training & \recDistinctPatterns{} \\
Training graphs per pattern & \recGraphsPerPattern{} \\
Validation graphs whose pattern occurs in training (\%) & \recValRecur{} \\
Test graphs whose pattern occurs in training (\%) & \recTestRecur{} \\
Distinct planted patterns in the test split & \recTestDistinctPatterns{} \\
Test graphs per pattern & \recTestGraphsPerPattern{} \\
Pattern size (nodes): min, median, max & \recSizeMin{}, \recSizeMedian{}, \recSizeMax{} \\
\bottomrule
\end{tabular}

\end{table}

\Needspace{10\baselineskip}
\section{The likelihood matcher and the pattern-blind predictor}
\label{app:likelihood}

\paragraph{The score.} The score of a copy $S$ of pattern $k$, with $m$ nodes, in a test graph of $N$ nodes approximates the log-likelihood ratio between two hypotheses, that $S$ is the planted pattern and that every node belongs to the host:
\begin{equation}
\begin{split}
\ell(S,k) \;=\; {} & m\log c \;-\sum_{\{u,v\}\subseteq S}\log p_H(a_{uv}) \;+\sum_{u\in S,\,w\notin S}\log\frac{q(a_{uw})}{p_H(a_{uw})} \\
& +\;\log\frac{\pi(N-m)}{\pi(N)} \;+\;\log\frac{|\mathrm{Aut}_k|\,(N-m)!}{N!} \;+\; m\log b .
\end{split}
\label{eq:score}
\end{equation}
Here $a_{uv}$ is one if $u$ and $v$ are joined and zero otherwise; $p_H$ gives its probability under the host's block model as fitted to the graph; $q(a) = q^{a}(1-q)^{1-a}$ gives it under the law that joins each pattern node to each host node with probability $q$; $\pi$ is the law of the host's size; $c$ and $b$ are the numbers of feature values and host blocks; and $\mathrm{Aut}_k$ is the group of feature-preserving automorphisms of pattern $k$. The terms follow the generator step by step. The pattern fixes the copy's features and internal edges, which are random under the host; the copy's edges to the rest of the graph follow $q$ instead of the block model; the host loses $m$ nodes; the uniform shuffle must put the pattern's nodes on $S$; and the host hypothesis must also place the copy's nodes in blocks.

\paragraph{Derivation.} The generator draws the host's block sizes, edges, and features, adds a pattern with its fixed edges and features, joins each pattern node to each host node with probability $q$, and shuffles the nodes uniformly. Under the hypothesis that the copy $S$ of pattern $k$ is the planted pattern, the probability of the observed graph is the product of six factors:
the probability $|\mathrm{Aut}_k|\,(N-m)!/N!$ that the shuffle puts the pattern's nodes on $S$, in any of the $|\mathrm{Aut}_k|$ arrangements that give this copy;
one for the copy's features and internal edges, which the pattern fixes;
$q$ or $1-q$ for each pair of a copy node and another node;
$c^{-(N-m)}$ for the host's features;
the host's block-model likelihood on the other $N-m$ nodes;
and $\pi(N-m)$ for the host's size, where $\pi$ is the law of the sum of $b$ independent block sizes.
Under the reference hypothesis that all $N$ nodes belong to the host, the probability is $c^{-N}$ times the block-model likelihood on all nodes times $\pi(N)$. We approximate the block-model likelihoods with a mean-field fit of the host's blocks. Each node outside the copy keeps its fitted probabilities of block membership, and each node of the copy is placed in the host block that best explains its edges. $p_H(a_{uv})$ is the resulting probability of the observed value of the pair, and placing the copy's $m$ nodes in blocks contributes the factor $b^{-m}$ to the reference hypothesis. The logarithm of this approximate ratio is Equation~\eqref{eq:score}. The reference hypothesis is the same for every copy in a graph, so it shifts all scores in a graph by the same amount.

\paragraph{The candidates.} The pattern set holds the distinct patterns of the training split (Appendix~\ref{app:generators}), each with its features and its number of feature-preserving automorphisms. A candidate is an induced copy of one of these patterns in the test graph with matching features. We find candidates with the LAD algorithm for subgraph isomorphism \citep{solnon2010alldifferent}, letting each pattern node map only to graph nodes with its feature and at least its degree. Patterns of \matcherBig{} nodes or more are tried from the largest down, and the first copy found ends the search. Otherwise every copy of every smaller pattern is listed among the \matcherTop{} nodes with the highest posterior of the pattern block under the sampler without the size prior (below), up to \matcherCap{} copies per pattern. The matcher labels the highest-scoring candidate its search evaluates; both the score and the search are approximate. The largest-match rule labels the largest candidate the same search finds, the earliest found among candidates of equal size, without the score (Table~\ref{tab:recall}).

The law of the host's size is zero above \hostSizeMax{} nodes, where the size term is undefined. The \realAboveHostGraphs{} official test graphs larger than that each carry a pattern of \matcherBig{} nodes or more, whose search stops at its first copy, so no score is needed, and each is recovered exactly. On one fresh-pattern graph of that kind all \freshAboveHostCands{} candidates had undefined scores, and the matcher fell back to the largest copy.

We fixed the matcher's settings without the validation split and before scoring it on the official test split; only the largest-match rule had been tried beforehand, on a sample of \fastGraphs{} test graphs. On the \valGraphs{} graphs of the validation split, scored afterward with the same settings, the matcher reaches \valMatcherPooled{} (95\% interval \valMatcherLo{} to \valMatcherHi{}), close to its official score.

\paragraph{The pattern-blind predictor.} The predictor's model has six blocks, five for the host and one for the pattern, with the block matrix of Table~\ref{tab:generators}: the host's probabilities within and across its blocks, and the pattern-to-host probability for every pair that involves the pattern block. As in the generator's code, each of the six blocks has a size drawn uniformly from \blindSizeMin{} to \blindSizeMax{} nodes, and the nodes are shuffled. The fresh patterns, drawn from \freshSizeMin{} to \freshSizeMax{} like the released ones, differ from this law only at the smallest size. Given the number of nodes, the prior of an assignment is therefore proportional to the product of the factorials of the six block sizes, and zero if any size leaves that range. A Gibbs move of one node into block $k$ is weighted by $s_k + 1$, the size block $k$ would reach (with $s_k$ counted without the node), times the node's edge likelihood under the block matrix. A mean-field fit with \blindRestarts{} random restarts of up to \blindMeanfieldIters{} iterations each keeps the restart with the highest evidence lower bound. The Gibbs sampler starts from that fit's most probable assignment and runs \blindSweeps{} sweeps, discarding the first \blindBurn{}. A node's posterior probability of the pattern block is the share of the remaining sweeps in which it sits there.

On \blindOutOfBoundsPct\% of the graphs, the fit's assignment gives some block a size outside the prior's range. While any block is out of range, a node in a block above the maximum must leave it, and a node in a block below the minimum must stay, though other nodes may join that block. Once every block is in range, the moves keep it there. The chain entered the range within \blindEnterMax{} sweeps on every graph, during the discarded sweeps, so every retained sweep samples the stated posterior. The threshold is the share of pattern nodes in the training split, \realTrainPosRate\%; a threshold of one half costs several points (Table~\ref{tab:blindchecks}).

\paragraph{The sampler without the size prior.} A simpler variant of the sampler has no prior on block sizes and runs \firstBlindSweeps{} sweeps, discarding the first \firstBlindBurn{}. The matcher's shortlist uses it, and several expectations in Appendix~\ref{app:registrations} refer to it. The prior and the longer chains together raise the score by \blindMoveOfficial{} points on the official test split (95\% interval \blindMoveOfficialLo{} to \blindMoveOfficialHi{}), and by similar amounts on the familiar and fresh sets (Table~\ref{tab:blindchecks}). This comparison does not separate the prior's share of the gain from that of the longer chains.

\paragraph{Other starts and chains.} As a check on mixing, we started both samplers at the planted pattern's nodes, with every other node in its most probable host block under the fit. This start uses the test labels, so it is not a predictor. Started there, each sampler scores as it does from its usual start (Table~\ref{tab:blindchecks}). On the first \blindChainGraphs{} official test graphs we also ran \blindChainsWord{} chains of the sampler with the size prior per graph, the first from its usual start and the others from random assignments. The other chains agree with the first on a median \blindChainAgreePct\% of each graph's labels, and the average of the \blindChainsWord{} chains' posteriors scores as the first chain alone does.

\paragraph{Calibration.} The predictor's posterior is close to calibrated. Equation~\eqref{eq:bayes}, pooled over a set's nodes and evaluated at each node's posterior, with the class counts that the posterior itself expects, gives the class-weighted accuracy the posterior expects of its predictions. That expectation exceeds the realized score by \calibRealExpectedMinusRealizedTwo{} points on the official test split, and by similar amounts on the familiar and fresh sets, so the posterior is slightly more confident than its realized accuracy. Its expected calibration error over ten bins of equal width is \calibRealEce{} on the official test split (Table~\ref{tab:blindchecks} gives both measures on every set).

\begin{table}[h]
\centering
\caption{Checks on the pattern-blind predictor and its sampler. Scores are class-weighted accuracy pooled over each set. Differences are paired by graph, with 95\% graph-bootstrap intervals, and computed before rounding. The comparison of starts covers the official test split, and that of chains its first \blindChainGraphs{} graphs. Expected minus realized is the class-weighted accuracy the posterior expects of its own predictions minus the score it reaches; the calibration error uses ten bins of equal width. Blank cells were not computed.}
\label{tab:blindchecks}
\small
\setlength{\tabcolsep}{4pt}
\resizebox{\linewidth}{!}{%
\begin{tabular}{lccc}
\toprule
 & Official test split & Familiar set & Fresh set \\
\midrule
Pattern-blind predictor (with the size prior) & \realDensityPooled{} & \familiarDensityPooled{} & \freshDensityPooled{} \\
Sampler without the size prior & \firstBlindRealPooled{} & \firstBlindFamiliarPooled{} & \firstBlindFreshPooled{} \\
With minus without the size prior & \blindMoveOfficial{} (\blindMoveOfficialLo{} to \blindMoveOfficialHi{}) & \blindMoveFamiliar{} (\blindMoveFamiliarLo{} to \blindMoveFamiliarHi{}) & \blindMoveFresh{} (\blindMoveFreshLo{} to \blindMoveFreshHi{}) \\
Pattern-blind predictor, threshold one half & \realDensityHalfPooled{} & & \freshDensityHalfPooled{} \\
\addlinespace
Start at the planted pattern minus usual start & & & \\
\quad with the size prior & \blindTrueStartBOneDiff{} (\blindTrueStartBOneDiffLo{} to \blindTrueStartBOneDiffHi{}) & & \\
\quad without the size prior & \blindTrueStartMinusFirst{} (\blindTrueStartMinusFirstLo{} to \blindTrueStartMinusFirstHi{}) & & \\
Average of \blindChainsWord{} chains minus the first chain & \blindFourChainDiff{} (\blindFourChainDiffLo{} to \blindFourChainDiffHi{}) & & \\
\addlinespace
Expected minus realized accuracy & \calibRealExpectedMinusRealizedTwo{} (\calibRealExpectedMinusRealizedLoTwo{} to \calibRealExpectedMinusRealizedHiTwo{}) & \calibFamiliarExpectedMinusRealizedTwo{} (\calibFamiliarExpectedMinusRealizedLoTwo{} to \calibFamiliarExpectedMinusRealizedHiTwo{}) & \calibFreshExpectedMinusRealizedTwo{} (\calibFreshExpectedMinusRealizedLoTwo{} to \calibFreshExpectedMinusRealizedHiTwo{}) \\
Expected calibration error & \calibRealEce{} & \calibFamiliarEce{} & \calibFreshEce{} \\
\bottomrule
\end{tabular}}
\end{table}

\paragraph{Where the matcher fails.} Exact recovery depends on pattern size, and Table~\ref{tab:exact} splits it at \matcherBig{} nodes. Each graph that is not recovered exactly falls into one of the kinds of error that the table defines, by the size of its labeled set and that set's overlap with the planted set. Patterns of \realSmallestSize{} nodes are the hardest: \realSmallestSizeExact{} of the \realSmallestSizeGraphs{} test graphs that carry one are recovered exactly. Among patterns of \matcherBig{} nodes or more, the \realMatcherFailWrongPlaceBig{} inexact graphs are all of one kind: a single candidate of the right size that shares all but one node with the planted set. A host node takes the place of a pattern node when it has the same feature and the same edges to the other pattern nodes. For a pattern of $m$ nodes this happens with probability about $c^{-1}q^{\,m-1}$ for each pair of a host node and a pattern node, since $q = \genPatternPHost{}$ makes an edge and a non-edge equally likely. If such a tie is lost half the time, the test graphs with large patterns should hold about \realMimicExpected{} errors of this kind. These near-ties come from the generator, and in each of them the matcher labels all but one pattern node correctly. On the fresh set no graph is recovered exactly, and \freshMatcherFailOtherPattern{} of the \freshGraphs{} receive a copy of some training pattern of another size.

\begin{table}[h]
\centering
\caption{Exact recovery of the planted set by pattern size, for the likelihood matcher and the largest-match rule, and the likelihood matcher's kinds of error, on the official test split and on the fresh set. The kinds depend only on the labeled set's size and its overlap with the planted set: a near miss has the planted set's size and shares all but one or two of its nodes; other size has a different size; low overlap has the planted set's size and shares fewer nodes; none found labels no node.}
\label{tab:exact}
\small
\setlength{\tabcolsep}{4pt}
\resizebox{\linewidth}{!}{\begin{tabular}{llrrrrrrr}
\toprule
 & & & \multicolumn{2}{c}{Exact recovery (\%)} & \multicolumn{4}{c}{Likelihood matcher, inexact graphs} \\
\cmidrule(lr){4-5}\cmidrule(lr){6-9}
Graphs & Pattern size & Count & Likelihood & Largest & Near miss & Other size & Low overlap & None found \\
\midrule
Official test split & 6 to 13 & \realGraphsSmall{} & \realMatcherExactPctSmall{} & \realLargestExactPctSmall{} & \realMatcherFailWrongPlaceSmall{} & \realMatcherFailOtherPatternSmall{} & \realMatcherFailLowOverlapSmall{} & \realMatcherFailMissedSmall{} \\
Official test split & 14 to 34 & \realGraphsBig{} & \realMatcherExactPctBig{} & \realLargestExactPctBig{} & \realMatcherFailWrongPlaceBig{} & \realMatcherFailOtherPatternBig{} & \realMatcherFailLowOverlapBig{} & \realMatcherFailMissedBig{} \\
Official test split & all & \realGraphs{} & \realMatcherExactPct{} & \realLargestExactPct{} & \realMatcherFailWrongPlace{} & \realMatcherFailOtherPattern{} & \realMatcherFailLowOverlap{} & \realMatcherFailMissed{} \\
\midrule
Fresh set & 6 to 13 & \freshGraphsSmall{} & \freshMatcherExactPctSmall{} & \freshLargestExactPctSmall{} & \freshMatcherFailWrongPlaceSmall{} & \freshMatcherFailOtherPatternSmall{} & \freshMatcherFailLowOverlapSmall{} & \freshMatcherFailMissedSmall{} \\
Fresh set & 14 to 34 & \freshGraphsBig{} & \freshMatcherExactPctBig{} & \freshLargestExactPctBig{} & \freshMatcherFailWrongPlaceBig{} & \freshMatcherFailOtherPatternBig{} & \freshMatcherFailLowOverlapBig{} & \freshMatcherFailMissedBig{} \\
Fresh set & all & \freshGraphs{} & \freshMatcherExactPct{} & \freshLargestExactPct{} & \freshMatcherFailWrongPlace{} & \freshMatcherFailOtherPattern{} & \freshMatcherFailLowOverlap{} & \freshMatcherFailMissed{} \\
\bottomrule
\end{tabular}
}

\end{table}

\section{The metric and its pooling}
\label{app:metric}

Class-weighted accuracy on a set of nodes is the mean, over the two classes, of the share of each class labeled correctly, and published work computes a PATTERN score in three ways. Pooled class-weighted accuracy is computed once over all nodes of the test split, as the code of GraphGPS and GRIT computes it \citep{rampasek2022gps,ma2023grit}; CKGConv's own evaluation gives the same values as ours (Appendix~\ref{app:fresh}). The batch mean computes it on each batch of \batchGraphs{} test graphs in the loader's order and averages over the batches, as the benchmark's own training code does with a batch size that depends on the model \citep{dwivedi2023benchmarking}. Plain node accuracy is the share of all test nodes labeled correctly, so it weights each class by its size, and pattern nodes are \convPositiveRate\% of the test nodes. Every interval in this paper is a 95\% percentile interval over bootstrap resamples of the graphs of a set: \hvProtocolBootstrapReplicatesGrouped{} resamples, or \matcherBootstrapReplicates{} for the likelihood matcher and the largest-match rule. The intervals hold the trained models and the training patterns fixed.

Table~\ref{tab:conventions} scores one set of predictions from each of our predictors under all three conventions. Plain node accuracy is at least pooled class-weighted accuracy for each predictor and exceeds it by up to \convMaxPlainMinusPooled{} points, and pooling instead of averaging over batches moves a score by up to \convMaxPooledMinusBatch{}. A threshold of one half costs the pattern-blind predictor several points of class-weighted accuracy against the class-prior rule of Section~\ref{sec:measured} (Appendix~\ref{app:likelihood}).

Published numbers mix these conventions. Gradformer reports its method at \pubPatternGradformer{} in the standard metric and at \pubPatternGradformerPlainAccuracy{} in the metric DeepGraph uses \citep{liu2024gradformer}. DeepGraph's \pubPatternDeepGraphPlainAccuracy{} is plain node accuracy, and its code also trains with an unweighted loss \citep{zhao2023deepgraph}, so the gap between Gradformer's two numbers need not come from the metric alone. Neither plain-accuracy number compares with the class-weighted leaderboard. Under plain accuracy, whose Bayes rule thresholds at one half, the pattern-blind predictor with that threshold scores \blindPlainHalfReal{} on the official test split (95\% interval \blindPlainHalfRealLo{} to \blindPlainHalfRealHi{}). DeepGraph's number lies below this score and Gradformer's within its interval, so neither shows a gain over the pattern-blind predictor. GRASS's table calls its metric accuracy, but its code computes class-weighted accuracy pooled over the test split \citep{liao2025grass}. GraphGPS reported \pubPatternGraphGPSFirstVersions{} on PATTERN in its first two arXiv versions and \pubPatternGraphGPS{} in its NeurIPS version, under the same metric name \citep{rampasek2022gps}; Table~\ref{tab:ladder} uses the latter. QPE reports \pubPatternGRITQiRWQPE{} for its best model, GRIT 2-QiRW, and \pubPatternGRITQPESRerun{} for its own run of GRIT, above GRIT's published \pubPatternGRIT{}, and states no rule for choosing the reported epoch \citep{thabet2024qpe}. We could not access the code it links to check that it computes the pooled class-weighted accuracy used here, so Table~\ref{tab:ladder} leaves its numbers out.

\begin{table}[h]
\centering
\caption{One set of predictions per predictor on PATTERN's official test split, scored under the three conventions in use. Differences are computed before rounding.}
\label{tab:conventions}
\small
\begin{tabular}{lrrrrr}
\toprule
 & \multicolumn{2}{c}{Class-weighted accuracy} & Pooled minus & Plain node & Plain minus \\
\cmidrule(lr){2-3}
Predictor & pooled & batch mean & batch mean & accuracy & pooled \\
\midrule
Lookup & \convLookupPooled{} & \convLookupBatch{} & \convLookupPooledMinusBatch{} & \convLookupPlain{} & \convLookupPlainMinusPooled{} \\
Likelihood matcher & \convMatcherPooled{} & \convMatcherBatch{} & \convMatcherPooledMinusBatch{} & \convMatcherPlain{} & \convMatcherPlainMinusPooled{} \\
Largest-match rule & \convLargestPooled{} & \convLargestBatch{} & \convLargestPooledMinusBatch{} & \convLargestPlain{} & \convLargestPlainMinusPooled{} \\
Pattern-blind predictor & \convDensityPooled{} & \convDensityBatch{} & \convDensityPooledMinusBatch{} & \convDensityPlain{} & \convDensityPlainMinusPooled{} \\
\bottomrule
\end{tabular}

\end{table}

\section{The fresh-pattern protocol}
\label{app:fresh}

\paragraph{The fresh set.} We drew \freshGraphs{} test graphs with our re-implementation of the generator's steps and parameters (Table~\ref{tab:generators}), each with its own pattern. The generator's code draws a pattern's size uniformly from \genPatternSizeMin{} to \genPatternSizeMax{} nodes, and the released patterns span \recSizeMin{} to \recSizeMax{}. We drew each fresh pattern's size uniformly from \freshSizeMin{} to \freshSizeMax{} nodes, its edges with probability \genPatternPInternal{}, and its features uniformly from three values. The \recDistinctPatterns{} released patterns are consistent with this law: chi-square tests against uniform sizes and uniform features give $p = \releasedSizesChiP{}$ and $p = \releasedFeatureChiP{}$, and both their internal edge density, \releasedDensity{}, and their rate of edges to host nodes, \releasedHostRate{}, are close to the generator's probability of \genPatternPInternal{}. Every method was scored on the same graphs.

\paragraph{The familiar set.} We drew \gpuFamiliarGraphs{} more graphs by the same code with another random seed, each carrying a pattern drawn uniformly from the training split's patterns with its node features; \familiarPatternsUsed{} of the \recDistinctPatterns{} patterns occur among them. The matcher recovers the planted set exactly in \familiarMatcherExactPct\% of these graphs and in \realMatcherExactPct\% of the official test graphs.

\paragraph{The size-matched set.} The familiar set carries the released patterns, while each graph of the fresh set draws its pattern's size anew, so the two sets differ in how their pattern sizes are distributed. The size-matched set removes that difference. We drew \sizematchGraphs{} more graphs with the fresh set's code and a new random seed. Graph~$i$ carries a newly drawn pattern with exactly as many nodes as the pattern of graph~$i$ of the familiar set, and everything else is drawn as in the fresh set. Differences between the familiar and size-matched sets are therefore paired by pattern size: graph~$i$ of each set has a pattern of the same size, while their hosts and patterns differ. No new pattern equals a released pattern or another new one. One equals a training pattern of the new-pattern models, and their scores leave out its graph (Table~\ref{tab:sizematch}). Each network was scored on this set from the checkpoint behind its other scores, after reproducing its recorded predictions on the official test split. GRASS's checkpoints reproduced all but \sizematchGrassDifferingPercent{}\% of their recorded predictions, counted over nodes and GRASS's \fiveGrassPasses{} evaluation passes, and changed no pooled score by as much as \sizematchGrassPooledMoveBound{} points; their values are marked in the table. CKGConv's drop is larger on patterns of \matcherBig{} nodes or more, as in the held-out test.

\begin{table}[h]
\centering
\caption{Class-weighted accuracy pooled over the familiar set (\sizematchGraphs{} graphs), the fresh set (\freshGraphs{}) and the size-matched set (\sizematchGraphs{}), whose patterns match those of the familiar set in size, graph by graph. The last column is a difference paired by pattern size (graph~$i$ of the familiar and size-matched sets has a pattern of the same size), with its 95\% graph-bootstrap interval, computed before rounding. The rows for patterns of \matcherBig{} nodes or more cover \sizematchGraphsBig{} graphs of the familiar and size-matched sets each and \freshGraphsBig{} of the fresh set. For the new-pattern models (CKGConv trained with a new pattern per graph), the scores on the familiar and size-matched sets leave out one pair of graphs, the size-matched graph whose pattern is among their training patterns and its partner in the familiar set; with this pair, the differences are \sizematchCkgNewSeedAAllFamiliarMinusMatched{}, \sizematchCkgNewSeedBAllFamiliarMinusMatched{} and \sizematchCkgNewSeedCAllFamiliarMinusMatched{}. Two-hop logistic regression is logistic regression on two-hop statistics (Section~\ref{sec:blind} and Appendix~\ref{app:withoutgen}). $^{*}$GRASS's checkpoints reproduced their recorded predictions on all but a small fraction of them (see text).}
\label{tab:sizematch}
\small
\setlength{\tabcolsep}{3.5pt}
\resizebox{\linewidth}{!}{\begin{tabular}{lccccc}
\toprule
 & Seed & Familiar set & Fresh set & Size-matched set & Familiar minus \\
 & & & & & size-matched, paired \\
\midrule
GatedGCN-PE & \fiveGatedSeedANumber{} & \sizematchGatedSeedAFamiliar{} & \sizematchGatedSeedAFresh{} & \sizematchGatedSeedAMatched{} & \sizematchGatedSeedAFamiliarMinusMatched{} (\sizematchGatedSeedAFamiliarMinusMatchedLo{} to \sizematchGatedSeedAFamiliarMinusMatchedHi{}) \\
 & \fiveGatedSeedBNumber{} & \sizematchGatedSeedBFamiliar{} & \sizematchGatedSeedBFresh{} & \sizematchGatedSeedBMatched{} & \sizematchGatedSeedBFamiliarMinusMatched{} (\sizematchGatedSeedBFamiliarMinusMatchedLo{} to \sizematchGatedSeedBFamiliarMinusMatchedHi{}) \\
 & \fiveGatedSeedCNumber{} & \sizematchGatedSeedCFamiliar{} & \sizematchGatedSeedCFresh{} & \sizematchGatedSeedCMatched{} & \sizematchGatedSeedCFamiliarMinusMatched{} (\sizematchGatedSeedCFamiliarMinusMatchedLo{} to \sizematchGatedSeedCFamiliarMinusMatchedHi{}) \\
 & \fiveGatedSeedDNumber{} & \sizematchGatedSeedDFamiliar{} & \sizematchGatedSeedDFresh{} & \sizematchGatedSeedDMatched{} & \sizematchGatedSeedDFamiliarMinusMatched{} (\sizematchGatedSeedDFamiliarMinusMatchedLo{} to \sizematchGatedSeedDFamiliarMinusMatchedHi{}) \\
\addlinespace
GRIT & 0 & \sizematchGritSeedAFamiliar{} & \sizematchGritSeedAFresh{} & \sizematchGritSeedAMatched{} & \sizematchGritSeedAFamiliarMinusMatched{} (\sizematchGritSeedAFamiliarMinusMatchedLo{} to \sizematchGritSeedAFamiliarMinusMatchedHi{}) \\
 & 1 & \sizematchGritSeedBFamiliar{} & \sizematchGritSeedBFresh{} & \sizematchGritSeedBMatched{} & \sizematchGritSeedBFamiliarMinusMatched{} (\sizematchGritSeedBFamiliarMinusMatchedLo{} to \sizematchGritSeedBFamiliarMinusMatchedHi{}) \\
 & \fiveGritSeedCNumber{} & \sizematchGritSeedCFamiliar{} & \sizematchGritSeedCFresh{} & \sizematchGritSeedCMatched{} & \sizematchGritSeedCFamiliarMinusMatched{} (\sizematchGritSeedCFamiliarMinusMatchedLo{} to \sizematchGritSeedCFamiliarMinusMatchedHi{}) \\
\addlinespace
CKGConv, released recipe & 0 & \sizematchCkgSeedAFamiliar{} & \sizematchCkgSeedAFresh{} & \sizematchCkgSeedAMatched{} & \sizematchCkgSeedAFamiliarMinusMatched{} (\sizematchCkgSeedAFamiliarMinusMatchedLo{} to \sizematchCkgSeedAFamiliarMinusMatchedHi{}) \\
 & 1 & \sizematchCkgSeedBFamiliar{} & \sizematchCkgSeedBFresh{} & \sizematchCkgSeedBMatched{} & \sizematchCkgSeedBFamiliarMinusMatched{} (\sizematchCkgSeedBFamiliarMinusMatchedLo{} to \sizematchCkgSeedBFamiliarMinusMatchedHi{}) \\
 & \fiveCkgSeedCNumber{} & \sizematchCkgSeedCFamiliar{} & \sizematchCkgSeedCFresh{} & \sizematchCkgSeedCMatched{} & \sizematchCkgSeedCFamiliarMinusMatched{} (\sizematchCkgSeedCFamiliarMinusMatchedLo{} to \sizematchCkgSeedCFamiliarMinusMatchedHi{}) \\
\quad patterns of \matcherBig{} nodes or more & 0 & \sizematchCkgSeedABigFamiliar{} & \sizematchCkgSeedABigFresh{} & \sizematchCkgSeedABigMatched{} & \sizematchCkgSeedABigFamiliarMinusMatched{} (\sizematchCkgSeedABigFamiliarMinusMatchedLo{} to \sizematchCkgSeedABigFamiliarMinusMatchedHi{}) \\
 & 1 & \sizematchCkgSeedBBigFamiliar{} & \sizematchCkgSeedBBigFresh{} & \sizematchCkgSeedBBigMatched{} & \sizematchCkgSeedBBigFamiliarMinusMatched{} (\sizematchCkgSeedBBigFamiliarMinusMatchedLo{} to \sizematchCkgSeedBBigFamiliarMinusMatchedHi{}) \\
 & \fiveCkgSeedCNumber{} & \sizematchCkgSeedCBigFamiliar{} & \sizematchCkgSeedCBigFresh{} & \sizematchCkgSeedCBigMatched{} & \sizematchCkgSeedCBigFamiliarMinusMatched{} (\sizematchCkgSeedCBigFamiliarMinusMatchedLo{} to \sizematchCkgSeedCBigFamiliarMinusMatchedHi{}) \\
\addlinespace
GRASS$^{*}$ & \fiveGrassSeedANumber{} & \sizematchGrassSeedAFamiliar{} & \sizematchGrassSeedAFresh{} & \sizematchGrassSeedAMatched{} & \sizematchGrassSeedAFamiliarMinusMatched{} (\sizematchGrassSeedAFamiliarMinusMatchedLo{} to \sizematchGrassSeedAFamiliarMinusMatchedHi{}) \\
 & \fiveGrassSeedBNumber{} & \sizematchGrassSeedBFamiliar{} & \sizematchGrassSeedBFresh{} & \sizematchGrassSeedBMatched{} & \sizematchGrassSeedBFamiliarMinusMatched{} (\sizematchGrassSeedBFamiliarMinusMatchedLo{} to \sizematchGrassSeedBFamiliarMinusMatchedHi{}) \\
 & \fiveGrassSeedCNumber{} & \sizematchGrassSeedCFamiliar{} & \sizematchGrassSeedCFresh{} & \sizematchGrassSeedCMatched{} & \sizematchGrassSeedCFamiliarMinusMatched{} (\sizematchGrassSeedCFamiliarMinusMatchedLo{} to \sizematchGrassSeedCFamiliarMinusMatchedHi{}) \\
\addlinespace
PPGT & \fivePpgtSeedANumber{} & \sizematchPpgtSeedAFamiliar{} & \sizematchPpgtSeedAFresh{} & \sizematchPpgtSeedAMatched{} & \sizematchPpgtSeedAFamiliarMinusMatched{} (\sizematchPpgtSeedAFamiliarMinusMatchedLo{} to \sizematchPpgtSeedAFamiliarMinusMatchedHi{}) \\
 & \fivePpgtSeedBNumber{} & \sizematchPpgtSeedBFamiliar{} & \sizematchPpgtSeedBFresh{} & \sizematchPpgtSeedBMatched{} & \sizematchPpgtSeedBFamiliarMinusMatched{} (\sizematchPpgtSeedBFamiliarMinusMatchedLo{} to \sizematchPpgtSeedBFamiliarMinusMatchedHi{}) \\
 & \fivePpgtSeedCNumber{} & \sizematchPpgtSeedCFamiliar{} & \sizematchPpgtSeedCFresh{} & \sizematchPpgtSeedCMatched{} & \sizematchPpgtSeedCFamiliarMinusMatched{} (\sizematchPpgtSeedCFamiliarMinusMatchedLo{} to \sizematchPpgtSeedCFamiliarMinusMatchedHi{}) \\
 & \fivePpgtSeedDNumber{} & \sizematchPpgtSeedDFamiliar{} & \sizematchPpgtSeedDFresh{} & \sizematchPpgtSeedDMatched{} & \sizematchPpgtSeedDFamiliarMinusMatched{} (\sizematchPpgtSeedDFamiliarMinusMatchedLo{} to \sizematchPpgtSeedDFamiliarMinusMatchedHi{}) \\
\addlinespace
CKGConv, new pattern per graph & 0 & \sizematchCkgNewSeedAFamiliar{} & \sizematchCkgNewSeedAFresh{} & \sizematchCkgNewSeedAMatched{} & \sizematchCkgNewSeedAFamiliarMinusMatched{} (\sizematchCkgNewSeedAFamiliarMinusMatchedLo{} to \sizematchCkgNewSeedAFamiliarMinusMatchedHi{}) \\
 & 1 & \sizematchCkgNewSeedBFamiliar{} & \sizematchCkgNewSeedBFresh{} & \sizematchCkgNewSeedBMatched{} & \sizematchCkgNewSeedBFamiliarMinusMatched{} (\sizematchCkgNewSeedBFamiliarMinusMatchedLo{} to \sizematchCkgNewSeedBFamiliarMinusMatchedHi{}) \\
 & \fiveCkgSeedCNumber{} & \sizematchCkgNewSeedCFamiliar{} & \sizematchCkgNewSeedCFresh{} & \sizematchCkgNewSeedCMatched{} & \sizematchCkgNewSeedCFamiliarMinusMatched{} (\sizematchCkgNewSeedCFamiliarMinusMatchedLo{} to \sizematchCkgNewSeedCFamiliarMinusMatchedHi{}) \\
\midrule
Two-hop logistic regression &  & \sizematchLadderFamiliar{} & \sizematchLadderFresh{} & \sizematchLadderMatched{} & \sizematchLadderFamiliarMinusMatched{} (\sizematchLadderFamiliarMinusMatchedLo{} to \sizematchLadderFamiliarMinusMatchedHi{}) \\
Pattern-blind predictor &  & \sizematchDensityFamiliar{} & \sizematchDensityFresh{} & \sizematchDensityMatched{} & \sizematchDensityFamiliarMinusMatched{} (\sizematchDensityFamiliarMinusMatchedLo{} to \sizematchDensityFamiliarMinusMatchedHi{}) \\
\bottomrule
\end{tabular}
}
\end{table}

\begin{figure}[ht]
\centering
\includegraphics[width=\linewidth]{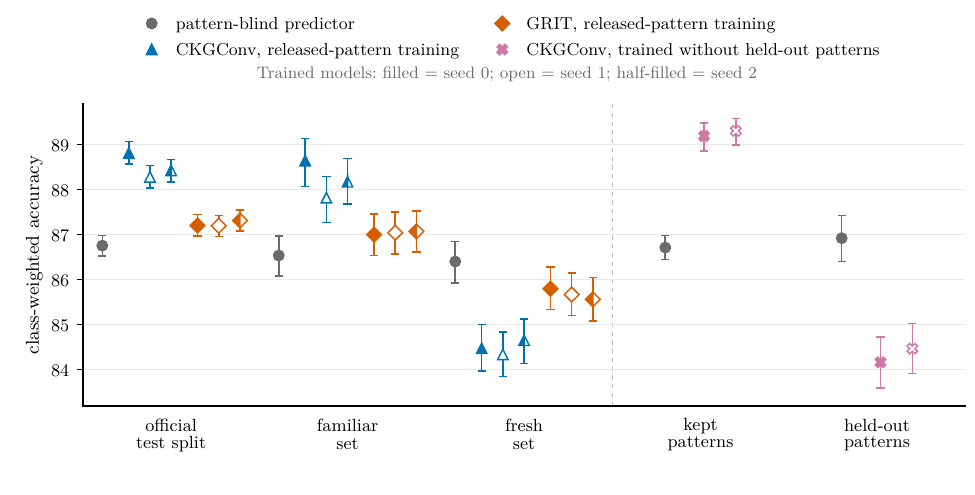}
\caption{Class-weighted accuracy on the official test split and the familiar and fresh sets. Right of the divider are graphs of the official test split whose patterns were kept in or held out of CKGConv's training, in the first draw of held-out patterns. Circles: the pattern-blind predictor; triangles: CKGConv trained on the released patterns; diamonds: GRIT; crosses: CKGConv trained without the held-out patterns. Bars are 95\% graph-bootstrap intervals for each method and seed, not intervals across seeds. The lookup and the matcher lie outside this range (Table~\ref{tab:fresh}).}
\label{fig:fresh}
\end{figure}

\paragraph{Two CKGConv seeds beside the methods without training.} Table~\ref{tab:fresh} gives CKGConv's first two seeds with the lookup, the matcher and the pattern-blind predictor, with intervals, on the official test split and the familiar and fresh sets, and Figure~\ref{fig:fresh} plots CKGConv's three seeds and the predictor beside GRIT and CKGConv trained without the held-out patterns (Section~\ref{sec:heldout}). On the fresh set the lookup and the matcher can label only chance copies of known patterns, and they fall below every trained model.

\begin{table}[ht]
\centering
\caption{CKGConv trained with its released recipe (seeds 0 and 1; Table~\ref{tab:ckgseeds} gives all three), and three methods without training, on the official test split and the familiar and fresh sets. Class-weighted accuracy is pooled over each set, and intervals resample whole graphs. The last column is the difference from the pattern-blind predictor on the fresh set, paired by graph and computed before rounding.}
\label{tab:fresh}
\small
\setlength{\tabcolsep}{3.5pt}
\resizebox{\linewidth}{!}{\begin{tabular}{lcccc}
\toprule
 & Official test split & Familiar set & Fresh set & Fresh set, minus the \\
 & & & & pattern-blind predictor \\
\midrule
Lookup of the training patterns & \lookupRealPooled{} (\lookupRealLo{} to \lookupRealHi{}) & \lookupFamiliarPooled{} (\lookupFamiliarLo{} to \lookupFamiliarHi{}) & \lookupFreshPooled{} (\lookupFreshLo{} to \lookupFreshHi{}) & \lookupFreshMinusDensity{} (\lookupFreshMinusDensityLo{} to \lookupFreshMinusDensityHi{}) \\
Likelihood matcher & \realMatcherPooled{} (\realMatcherLo{} to \realMatcherHi{}) & \familiarMatcherPooled{} (\familiarMatcherLo{} to \familiarMatcherHi{}) & \freshMatcherPooled{} (\freshMatcherLo{} to \freshMatcherHi{}) & \freshMatcherMinusDensity{} (\freshMatcherMinusDensityLo{} to \freshMatcherMinusDensityHi{}) \\
Pattern-blind predictor & \realDensityPooled{} (\realDensityLo{} to \realDensityHi{}) & \familiarDensityPooled{} (\familiarDensityLo{} to \familiarDensityHi{}) & \freshDensityPooled{} (\freshDensityLo{} to \freshDensityHi{}) & \\
CKGConv, seed 0 & \gpuOriginalPooled{} (\gpuOriginalLo{} to \gpuOriginalHi{}) & \gpuFamiliarPooled{} (\gpuFamiliarLo{} to \gpuFamiliarHi{}) & \gpuFreshPooled{} (\gpuFreshLo{} to \gpuFreshHi{}) & \gpuMinusDensity{} (\gpuMinusDensityLo{} to \gpuMinusDensityHi{}) \\
CKGConv, seed 1 & \gpuSeedBOriginalPooled{} (\gpuSeedBOriginalLo{} to \gpuSeedBOriginalHi{}) & \gpuSeedBFamiliarPooled{} (\gpuSeedBFamiliarLo{} to \gpuSeedBFamiliarHi{}) & \gpuSeedBFreshPooled{} (\gpuSeedBFreshLo{} to \gpuSeedBFreshHi{}) & \gpuSeedBMinusDensity{} (\gpuSeedBMinusDensityLo{} to \gpuSeedBMinusDensityHi{}) \\
\bottomrule
\end{tabular}
}
\end{table}

\paragraph{Scores and intervals.} Every accuracy in this appendix is class-weighted accuracy pooled over all nodes of a set (Appendix~\ref{app:metric}), computed from saved predictions, except GatedGCN-PE's official scores in the release's batch mean, which its paragraph names. For CKGConv these scores equal its own evaluation, on every set for its first two seeds and on the official test split for its third. Intervals resample whole graphs (Appendix~\ref{app:metric}). Two predictors scored on the same graphs, including two models with the same seed, are resampled together, and two sets of graphs independently; resampling the fresh set includes its independent draws of pattern and host. ROC AUC is computed from pattern-class probabilities pooled over nodes (Table~\ref{tab:auc}). A paired difference from the pattern-blind predictor near zero is not an equivalence test.

\paragraph{Conversion and round trip.} Each graph of our sets was converted to the data format of PyTorch Geometric's PATTERN graphs, with the same types and shapes, and CKGConv's own code computed its random-walk positional encodings. To check the conversion, we took the first \gpuRoundtripGraphs{} official test graphs apart into features, edges, and labels, passed them through the same conversion and encoding, and scored them. CKGConv's predictions on them agree with those through its own loader on every node.

\paragraph{Resumed training.} The runs of CKGConv, GRIT, GRASS and PPGT were resumed from saved checkpoints that restore the model, the optimizer, the learning-rate schedule and the epoch counter, so the learning rate follows the released schedule without a break; for CKGConv and PPGT, which share code, this required a change to how the code saves the schedule. GRIT, GRASS and PPGT also restore their random states, and a short GRIT run resumed this way matched an uninterrupted one exactly. CKGConv's random streams for data order and dropout restart at each resumption, so its runs follow the released recipe without matching an uninterrupted run. Every scored checkpoint is the one the released rule selects over the full training history.

\paragraph{Training CKGConv.} We trained CKGConv with the configuration its authors released for PATTERN for \gpuEpochs{} epochs, with seeds 0, 1 and 2, each on one T4 GPU for about \gpuTrainHoursRound{} hours. Its positional encodings were computed as each graph was loaded, which gives the same values in less memory than storing them. The released rule selects the last epoch after the warm-up at which validation accuracy reaches a new maximum over all epochs so far, the first occurrence winning a tie; Table~\ref{tab:ckgseeds} gives the selected epochs. The model has about \gpuParamsRound{} parameters.

\paragraph{Patterns held out of training.} The \heldPatterns{} patterns were drawn with a fixed seed, separately among the \heldSmallPool{} patterns with fewer than \matcherBig{} nodes and the \heldBigPool{} with \matcherBig{} or more, and found in every split by their canonical form (Appendix~\ref{app:generators}). Every graph of the three splits carries one of the \recDistinctPatterns{} patterns. Removing the graphs that carry a drawn pattern leaves \heldTrainGraphs{} training and \heldValGraphs{} validation graphs, and divides the official test split into \heldGraphs{} graphs with a held-out pattern and \keptGraphs{} with a kept one. CKGConv was trained on the remaining training graphs with the recipe above and seeds 0 and 1, and its checkpoint was chosen by the released rule on the remaining validation graphs. A second draw of \heldPatterns{} patterns, by the same rule among the patterns not held out the first time, leaves \heldSecondGraphs{} official test graphs with a held-out pattern and \heldSecondKeptGraphs{} with a kept one; CKGConv was trained without it and scored in the same way. Differences from the pattern-blind predictor, and from CKGConv trained on all \recDistinctPatterns{} patterns with the same seed, are paired on the same graphs; the difference between held-out and kept graphs is not (Table~\ref{tab:heldout}).

\begin{table}[h]
\centering
\caption{CKGConv trained with its released recipe, seeds 0, 1 and 2. Class-weighted accuracy pooled over each set, with 95\% bootstrap intervals over graphs. Epochs count from zero. No interval was computed for seed 2's familiar set minus official. Differences are computed before rounding.}
\label{tab:ckgseeds}
\small
\setlength{\tabcolsep}{3.5pt}
\begin{tabular}{llll}
\toprule
 & Seed 0 & Seed 1 & Seed 2 \\
\midrule
Best-validation epoch, of \gpuEpochs{} & \gpuCheckpointEpoch{} & \gpuSeedBCheckpointEpoch{} & \hvCompletedCkgSeedCFullRecipeBestEpoch{} \\
Training time (hours on one T4) & \gpuTrainHours{} & \gpuSeedBTrainHours{} & \gpuSeedCTrainHours{} \\
Official test split & \gpuOriginalPooled{} (\gpuOriginalLo{} to \gpuOriginalHi{}) & \gpuSeedBOriginalPooled{} (\gpuSeedBOriginalLo{} to \gpuSeedBOriginalHi{}) & \gpuSeedCOriginalPooled{} (\gpuSeedCOriginalLo{} to \gpuSeedCOriginalHi{}) \\
Familiar set & \gpuFamiliarPooled{} (\gpuFamiliarLo{} to \gpuFamiliarHi{}) & \gpuSeedBFamiliarPooled{} (\gpuSeedBFamiliarLo{} to \gpuSeedBFamiliarHi{}) & \gpuSeedCFamiliarPooled{} (\gpuSeedCFamiliarLo{} to \gpuSeedCFamiliarHi{}) \\
Fresh set & \gpuFreshPooled{} (\gpuFreshLo{} to \gpuFreshHi{}) & \gpuSeedBFreshPooled{} (\gpuSeedBFreshLo{} to \gpuSeedBFreshHi{}) & \gpuSeedCFreshPooled{} (\gpuSeedCFreshLo{} to \gpuSeedCFreshHi{}) \\
Familiar set minus official & \gpuFamiliarMinusOriginal{} (\gpuFamiliarMinusOriginalLo{} to \gpuFamiliarMinusOriginalHi{}) & \gpuSeedBFamiliarMinusOriginal{} (\gpuSeedBFamiliarMinusOriginalLo{} to \gpuSeedBFamiliarMinusOriginalHi{}) & \gpuSeedCFamiliarMinusOriginal{} \\
Fresh set minus familiar set & \gpuFreshMinusFamiliar{} (\gpuFreshMinusFamiliarLo{} to \gpuFreshMinusFamiliarHi{}) & \gpuSeedBFreshMinusFamiliar{} (\gpuSeedBFreshMinusFamiliarLo{} to \gpuSeedBFreshMinusFamiliarHi{}) & \gpuSeedCFreshMinusFamiliar{} (\gpuSeedCFreshMinusFamiliarLo{} to \gpuSeedCFreshMinusFamiliarHi{}) \\
Fresh minus the pattern-blind predictor, paired & \gpuMinusDensity{} (\gpuMinusDensityLo{} to \gpuMinusDensityHi{}) & \gpuSeedBMinusDensity{} (\gpuSeedBMinusDensityLo{} to \gpuSeedBMinusDensityHi{}) & \gpuSeedCMinusDensity{} (\gpuSeedCMinusDensityLo{} to \gpuSeedCMinusDensityHi{}) \\
\bottomrule
\end{tabular}

\end{table}

\begin{table}[h]
\centering
\caption{CKGConv's scores by pattern size, and the pattern-blind predictor's on the same graphs. Seed 2's scores on the official test split were not computed by pattern size. Pooled over a whole set, class-weighted accuracy can lie above its value on both halves: pattern nodes come mostly from large patterns, host nodes from every graph.}
\label{tab:ckgstrata}
\small
\begin{tabular}{llcccc}
\toprule
Test set & Pattern size & Seed 0 & Seed 1 & Seed 2 & Pattern-blind predictor \\
\midrule
Official test split & under \matcherBig{} nodes & \gpuOriginalSmall{} & \gpuSeedBOriginalSmall{} &  & \realDensitySmall{} \\
Official test split & \matcherBig{} nodes or more & \gpuOriginalBig{} & \gpuSeedBOriginalBig{} &  & \realDensityBig{} \\
Familiar set & under \matcherBig{} nodes & \gpuFamiliarSmall{} & \gpuSeedBFamiliarSmall{} & \gpuSeedCFamiliarSmall{} & \familiarDensitySmall{} \\
Familiar set & \matcherBig{} nodes or more & \gpuFamiliarBig{} & \gpuSeedBFamiliarBig{} & \gpuSeedCFamiliarBig{} & \familiarDensityBig{} \\
Fresh set & under \matcherBig{} nodes & \gpuFreshSmall{} & \gpuSeedBFreshSmall{} & \gpuSeedCFreshSmall{} & \gpuFreshDensitySmall{} \\
Fresh set & \matcherBig{} nodes or more & \gpuFreshBig{} & \gpuSeedBFreshBig{} & \gpuSeedCFreshBig{} & \gpuFreshDensityBig{} \\
\bottomrule
\end{tabular}

\end{table}

\begin{table}[h]
\centering
\caption{CKGConv trained without \heldPatterns{} of the \recDistinctPatterns{} patterns, in two draws, on the official test graphs split by whether their pattern was held out or kept (\heldGraphs{} and \keptGraphs{} graphs in the first draw, \heldSmallGraphs{} of the held-out ones with patterns under \matcherBig{} nodes; \heldSecondGraphs{} and \heldSecondKeptGraphs{} in the second). Blank cells were not computed for the second draw. Class-weighted accuracy pooled over each set, with 95\% bootstrap intervals over graphs where given. CKGConv trained on all \recDistinctPatterns{} patterns is the model of Table~\ref{tab:ckgseeds} with the same seed, and the pattern-blind predictor is the same for both seeds. Differences are computed before rounding.}
\label{tab:heldout}
\small
\setlength{\tabcolsep}{3pt}
\resizebox{\linewidth}{!}{\begin{tabular}{lllll}
\toprule
 & \multicolumn{2}{c}{First draw} & \multicolumn{2}{c}{Second draw} \\
\cmidrule(lr){2-3}\cmidrule(lr){4-5}
 & Seed 0 & Seed 1 & Seed 0 & Seed 1 \\
\midrule
Test graphs whose pattern was held out & \heldH{} (\heldHLo{} to \heldHHi{}) & \heldSeedBH{} (\heldSeedBHLo{} to \heldSeedBHHi{}) & \heldSecondH{} (\heldSecondHLo{} to \heldSecondHHi{}) & \heldSecondSeedBH{} (\heldSecondSeedBHLo{} to \heldSecondSeedBHHi{}) \\
Test graphs whose pattern was kept & \heldKept{} (\heldKeptLo{} to \heldKeptHi{}) & \heldSeedBKept{} (\heldSeedBKeptLo{} to \heldSeedBKeptHi{}) & \heldSecondKept{} (\heldSecondKeptLo{} to \heldSecondKeptHi{}) & \heldSecondSeedBKept{} (\heldSecondSeedBKeptLo{} to \heldSecondSeedBKeptHi{}) \\
Held out minus kept & \heldHMinusKept{} (\heldHMinusKeptLo{} to \heldHMinusKeptHi{}) & \heldSeedBHMinusKept{} (\heldSeedBHMinusKeptLo{} to \heldSeedBHMinusKeptHi{}) & \heldSecondHMinusKept{} (\heldSecondHMinusKeptLo{} to \heldSecondHMinusKeptHi{}) & \heldSecondSeedBHMinusKept{} (\heldSecondSeedBHMinusKeptLo{} to \heldSecondSeedBHMinusKeptHi{}) \\
The pattern-blind predictor on the held-out graphs & \multicolumn{2}{c}{\heldBlindH{} (\heldBlindHLo{} to \heldBlindHHi{})} & \multicolumn{2}{c}{\heldSecondDensityH{} (\heldSecondDensityHLo{} to \heldSecondDensityHHi{})} \\
The pattern-blind predictor on the kept graphs & \multicolumn{2}{c}{\heldBlindKept{} (\heldBlindKeptLo{} to \heldBlindKeptHi{})} & \multicolumn{2}{c}{} \\
Held out minus the pattern-blind predictor, paired & \heldHMinusBlind{} (\heldHMinusBlindLo{} to \heldHMinusBlindHi{}) & \heldSeedBHMinusBlind{} (\heldSeedBHMinusBlindLo{} to \heldSeedBHMinusBlindHi{}) & \heldSecondHMinusDensity{} (\heldSecondHMinusDensityLo{} to \heldSecondHMinusDensityHi{}) & \heldSecondSeedBHMinusDensity{} (\heldSecondSeedBHMinusDensityLo{} to \heldSecondSeedBHMinusDensityHi{}) \\
CKGConv trained on all \recDistinctPatterns{} patterns, on the held-out graphs & \heldFullH{} & \heldSeedBFullH{} & \heldSecondFullH{} & \heldSecondSeedBFullH{} \\
Held out minus CKGConv trained on all \recDistinctPatterns{} patterns, paired & \heldHMinusFull{} (\heldHMinusFullLo{} to \heldHMinusFullHi{}) & \heldSeedBHMinusFull{} (\heldSeedBHMinusFullLo{} to \heldSeedBHMinusFullHi{}) & \heldSecondHMinusFull{} (\heldSecondHMinusFullLo{} to \heldSecondHMinusFullHi{}) & \heldSecondSeedBHMinusFull{} (\heldSecondSeedBHMinusFullLo{} to \heldSecondSeedBHMinusFullHi{}) \\
Kept minus CKGConv trained on all \recDistinctPatterns{} patterns, paired & \heldKeptMinusFull{} (\heldKeptMinusFullLo{} to \heldKeptMinusFullHi{}) & \heldSeedBKeptMinusFull{} (\heldSeedBKeptMinusFullLo{} to \heldSeedBKeptMinusFullHi{}) & \heldSecondKeptMinusFull{} (\heldSecondKeptMinusFullLo{} to \heldSecondKeptMinusFullHi{}) & \heldSecondSeedBKeptMinusFull{} (\heldSecondSeedBKeptMinusFullLo{} to \heldSecondSeedBKeptMinusFullHi{}) \\
\midrule
Held out, patterns of \matcherBig{} nodes or more & \heldBigH{} & \heldSeedBBigH{} &  &  \\
Kept, patterns of \matcherBig{} nodes or more & \heldBigKept{} & \heldSeedBBigKept{} &  &  \\
Held out, patterns under \matcherBig{} nodes & \heldSmallH{} & \heldSeedBSmallH{} &  &  \\
Kept, patterns under \matcherBig{} nodes & \heldSmallKept{} & \heldSeedBSmallKept{} &  &  \\
\bottomrule
\end{tabular}
}

\end{table}

\paragraph{Training GRIT.} We trained GRIT with its released PATTERN configuration and unchanged model code \citep{ma2023grit} for \hvCompletedGritSeedAEpochs{} epochs, with seeds 0, 1 and 2 and the released training and validation splits. The released batch did not fit on a T4 GPU, so each optimizer step averaged the gradient over \hvProtocolGritAccumulation{} micro-batches of \hvProtocolGritMicroBatch{} graphs. Batch normalization and class weights were computed within each micro-batch, which differs from a single batch with the same total number of graphs. Random-walk encodings were computed each time a graph was loaded. GRIT's released rule for choosing the checkpoint is CKGConv's, with a warm-up of \hvProtocolGritWarmup{} epochs, and it selects epochs \hvCompletedGritSeedABestEpoch{}, \hvCompletedGritSeedBBestEpoch{} and \hvCompletedGritSeedCBestEpoch{} for the three seeds, counting from zero. Seeds 0 and 1 print the same official score to two decimals, but they are different models: their predicted labels differ on \reviewGritSeedsDifferNodes{} of the \reviewOfficialTestNodes{} official test nodes (\reviewGritSeedsDifferPct\%). The round trip on \gpuRoundtripGraphs{} official test graphs matched labels, transformed inputs, and predictions on the CPU, where GRIT's arithmetic is deterministic. Every seed reproduces the published official score of \pubPatternGRIT{} to about a tenth of a point and scores about as well on the familiar set (Table~\ref{tab:gritcompleted}).

\paragraph{Training GatedGCN-PE.} We trained GatedGCN-PE with the benchmark's released code and its PATTERN configuration with positional encodings, unchanged \citep{dwivedi2023benchmarking}, with the release's four seeds, \gatedSeedASeed{}, \gatedSeedBSeed{}, \fiveGatedSeedCNumber{} and \fiveGatedSeedDNumber{}, each on one T4 GPU. The published runs were cut by a time limit, so we capped training at \gatedEpochsCap{} epochs, their mean length, and kept every other rule of the release; its floor on the learning rate stopped the first two seeds after \gatedSeedAEpochs{} and \gatedSeedBEpochs{} epochs. As in the release, the last epoch's model is scored. PATTERN was read from PyTorch Geometric's copy and converted to the release's format, with every graph checked against the source arrays, and the familiar and fresh sets were converted the same way. In the release's batch mean, the first two seeds score \gatedSeedAOfficialBatch{} and \gatedSeedBOfficialBatch{} on the official test split, against the published \pubPatternGatedGCNPE{}; every other score is pooled.

\paragraph{Training GRASS.} We trained GRASS with its released PATTERN configuration and code \citep{liao2025grass} for its full \fiveGrassEpochs{} epochs, with seeds 0, 1 and 2 set in every random number generator, which the release does not do. The released batch of \fiveGrassReleasedBatch{} graphs did not fit on a T4 GPU, so each optimizer step accumulated the gradients of \fiveGrassAccumulation{} micro-batches of \fiveGrassMicroBatch{} graphs. This keeps the recipe's number of optimizer steps, learning-rate schedule and optimizer update, while batch normalization and the loss's class weights see \fiveGrassMicroBatch{} graphs at a time; GRASS's own option to recompute activations was turned on to save memory. As in the release, the model after the last epoch is scored, its batch-normalization statistics are re-estimated on the training split, and every set is scored in \fiveGrassPasses{} passes with new random edges, whose accuracies are averaged. GRASS's own evaluation draws other random edges, and its official scores differ from ours by less than \grassOwnEvaluationGapBound{} point in every seed.

\paragraph{Training PPGT.} We trained PPGT with its released PATTERN configuration and code \citep{ma2026ppgt} for \fivePpgtEpochs{} epochs with seeds 0 to 3, and scored the checkpoint its validation rule selects: epochs \fivePpgtSeedABestEpoch{}, \fivePpgtSeedBBestEpoch{}, \fivePpgtSeedCBestEpoch{} and \fivePpgtSeedDBestEpoch{}.

\paragraph{GRASS and PPGT by seed.} Table~\ref{tab:fivemodels_seeds} gives each seed's official, familiar and fresh scores and its fall from the familiar to the fresh set. Our runs of both models score about a point above the published means: on average by \fiveGrassOverPublished{} points for GRASS and \fivePpgtOverPublished{} for PPGT. PPGT's seeds span \fivePpgtOfficialSpread{} points on the official test split, while its paper reports a standard deviation of \pubPatternPPGTSd{} over seeds. These runs depart from the authors' in seeding, in GRASS's micro-batches and in resumed training. Within each model a seed's fall does not shrink consistently as its official score approaches the published value. PPGT's seed 0, whose official score is the closest to the published one, loses the least, and still more than any seed of GatedGCN-PE, GRIT or CKGConv; in GRASS the seed closest to the published score loses the most.

\paragraph{Training CKGConv on new patterns.} We generated a new training split of \hvProtocolFreshTrainGraphsGrouped{} graphs and a new validation split of \hvProtocolFreshValGraphsGrouped{} graphs. Every graph has its own newly drawn pattern, under the same laws of pattern size, internal edges, features, and host as the fresh test graphs. All these patterns are distinct under feature-preserving canonical forms, and none of them occurs among the released patterns or the patterns of the fresh set. We trained three seeds on these splits, the new-pattern models, with the architecture, optimizer, learning-rate schedule, positional encodings, and settings of the released recipe \citep{ma2024ckgconv}, for \hvCompletedCkgSeedAEpochs{} epochs, and their checkpoints were selected on the new validation split by the released rule (Table~\ref{tab:ckgfreshcompleted}). As with GRIT, seeds 0 and 1 print the same official score to two decimals from checkpoints of different epochs, and their predicted labels differ on \reviewCkgNewSeedsDifferNodes{} official test nodes (\reviewCkgNewSeedsDifferPct\%). The official test accuracy that CKGConv logs during training played no part in selecting a checkpoint, here or in the reuse control below. Since the validation split changed too, the comparison with the released recipe includes the effect of selecting checkpoints on new validation patterns as well as that of training on new patterns.

\paragraph{Training CKGConv on a reused library.} For the reuse control (Section~\ref{sec:reuse}) we drew three libraries of \reuseLibraryPatterns{} patterns each, under the fresh set's laws of pattern size, internal edges and features. Under feature-preserving canonical forms, every library pattern differs from the released patterns, the patterns of the fresh and size-matched sets, those of the new training and validation splits above, and the patterns of the other libraries. For each library, the code that generated those splits generated \reuseTrainGraphs{} training, \reuseValGraphs{} validation and \reuseTestGraphs{} test graphs with one change: the graphs of each split take the library's patterns in consecutive blocks, as the benchmark's generator assigns its own, so each pattern occurs in \recGraphsPerPattern{} training graphs, as a released pattern does, and in \reuseTestGraphsPerPattern{} of the library's test graphs. CKGConv was trained on each library with its released recipe for \gpuEpochs{} epochs, the first library with seed 0, the second with seed 1 and the third with seed 2, and its checkpoint was chosen by the released rule on the library's validation graphs; these are the reuse models. Each model was scored on its library's test graphs, the fresh and familiar sets and the official test split. Its comparisons with the new-pattern model and with CKGConv trained on the released patterns use the model with the same seed and are paired by graph.

CKGConv's training logs the class-weighted accuracy on the official test split at every epoch, and at each scored checkpoint the logged value equals our pooled score. Table~\ref{tab:reusedetail} reads the reuse models' logs at the epochs selected for the new-pattern models of the same seeds. For seed 2, the new-pattern model's log covers every epoch, and the comparison over all epochs in the next paragraph uses it.

\paragraph{Training length in the reuse control.}\label{app:traininglength} Selection on reused validation patterns chose late checkpoints, and selection on new validation patterns chose early ones (Table~\ref{tab:reusedetail}). A reuse model's official accuracy peaks soon after warm-up, where it equals the pattern-blind predictor's to one decimal, and then falls as training on the library continues. At the new-pattern models' checkpoint epochs the reuse models score higher on the official test split than at their own checkpoints, and still lie below the new-pattern models (the last three columns of Table~\ref{tab:reusedetail}). For seed 2, where both models' logs cover every epoch, the reuse model lies below the new-pattern model at \matchedEpochsBelow{} of the \matchedEpochsAfterWarmup{} epochs after warm-up. Longer training lowers the new-pattern model's official accuracy too: at the reuse model's checkpoint epoch it lies about \matchedNewPatternAtReuseEpochBelowDensity{} points below the pattern-blind predictor, and the reuse model a further \matchedGapAtReuseEpoch{}. Part of the reuse models' loss against the new-pattern models therefore comes from the longer training that selection on reused patterns allows.

\begin{table}[h]
\centering
\caption{The reuse control by seed (Section~\ref{sec:reuse}). The selected epoch is the checkpoint the released rule chose on the library's validation graphs. The next three columns are differences paired by graph, with 95\% graph-bootstrap intervals: the reuse model's score minus that of the new-pattern model with the same seed (Section~\ref{sec:freshtraining}), on the library set and on the fresh set, and its official score minus that of CKGConv trained on the released patterns with the same seed (Section~\ref{sec:fresh}). The last three columns give the epoch selected for the new-pattern model of the same seed, the reuse model's official accuracy logged at that epoch, and the new-pattern model's official score there. Differences are computed before rounding.}
\label{tab:reusedetail}
\small
\setlength{\tabcolsep}{3.5pt}
\resizebox{\linewidth}{!}{\begin{tabular}{cccccccc}
\toprule
 & Selected & \multicolumn{2}{c}{Minus new-pattern training, paired} & Official minus released- & \multicolumn{3}{c}{At the new-pattern model's epoch} \\
\cmidrule(lr){3-4}\cmidrule(lr){6-8}
Seed & epoch & Library set & Fresh set & pattern training, paired & Epoch & Reuse model & New-pattern model \\
\midrule
\reuseRunOneSeed{} & \reuseRunOneEpoch{} & \librefOneReuseMinusNew{} (\librefOneReuseMinusNewLo{} to \librefOneReuseMinusNewHi{}) & \reuseRunOneMinusNewPatternTraining{} (\reuseRunOneMinusNewPatternTrainingLo{} to \reuseRunOneMinusNewPatternTrainingHi{}) & \reuseRunOneMinusReleasedPatternTraining{} (\reuseRunOneMinusReleasedPatternTrainingLo{} to \reuseRunOneMinusReleasedPatternTrainingHi{}) & \hvCompletedCkgSeedABestEpoch{} & \reuseRunOneAtNewPatternEpoch{} & \hvCompletedCkgSeedAOfficialAccuracyRound{} \\
\reuseRunTwoSeed{} & \reuseRunTwoEpoch{} & \librefTwoReuseMinusNew{} (\librefTwoReuseMinusNewLo{} to \librefTwoReuseMinusNewHi{}) & \reuseRunTwoMinusNewPatternTraining{} (\reuseRunTwoMinusNewPatternTrainingLo{} to \reuseRunTwoMinusNewPatternTrainingHi{}) & \reuseRunTwoMinusReleasedPatternTraining{} (\reuseRunTwoMinusReleasedPatternTrainingLo{} to \reuseRunTwoMinusReleasedPatternTrainingHi{}) & \hvCompletedCkgSeedBBestEpoch{} & \reuseRunTwoAtNewPatternEpoch{} & \hvCompletedCkgSeedBOfficialAccuracyRound{} \\
\reuseRunThreeSeed{} & \reuseRunThreeEpoch{} & \librefThreeReuseMinusNew{} (\librefThreeReuseMinusNewLo{} to \librefThreeReuseMinusNewHi{}) & \reuseRunThreeMinusNewPatternTraining{} (\reuseRunThreeMinusNewPatternTrainingLo{} to \reuseRunThreeMinusNewPatternTrainingHi{}) & \reuseRunThreeMinusReleasedPatternTraining{} (\reuseRunThreeMinusReleasedPatternTrainingLo{} to \reuseRunThreeMinusReleasedPatternTrainingHi{}) & \hvCompletedCkgSeedCBestEpoch{} & \reuseRunThreeAtNewPatternEpoch{} & \hvCompletedCkgSeedCOfficialAccuracyRound{} \\
\bottomrule
\end{tabular}
}
\end{table}

\paragraph{A reference on the library graphs.} To read the gap between seen and new patterns on the same graphs (Section~\ref{sec:reuse}), we scored three more things: the pattern-blind predictor on each library's \reuseTestGraphs{} test graphs, with the code, settings and threshold it uses on the fresh and size-matched sets; each new-pattern model on the test graphs of the library whose reuse model has its seed; and each reuse model on the size-matched set. Each model first reproduced its recorded predictions on the official test split and the fresh and familiar sets. The libraries' patterns follow the fresh set's laws but are not matched to its sizes (Table~\ref{tab:libref}).

\begin{table}[h]
\centering
\caption{The reference on the reuse control's library sets, by seed (Section~\ref{sec:reuse}). Each library set holds the \reuseTestGraphs{} test graphs that carry that library's patterns, and the new-pattern model is CKGConv trained with a new pattern per graph with the seed of the library's reuse model. Familiar minus size-matched is paired by position, graph $i$ of each set having a pattern of the same size (Table~\ref{tab:sizematch}); the other differences marked paired are paired by graph, and the rest compare two sets of graphs. Intervals are 95\% graph-bootstrap intervals, and differences are computed before rounding. The fresh set's patterns have mean size \librefFreshPatternSize{} nodes, and \librefFreshSmallPatterns{} of its graphs carry a pattern under \matcherBig{} nodes.}
\label{tab:libref}
\small
\setlength{\tabcolsep}{4pt}
\begin{tabular}{lccc}
\toprule
 & Seed \reuseRunOneSeed{}, library 1 & Seed \reuseRunTwoSeed{}, library 2 & Seed \reuseRunThreeSeed{}, library 3 \\
\midrule
\multicolumn{4}{l}{\itshape Pattern-blind predictor} \\
\quad Library set & \librefOneBlindLibrary{} & \librefTwoBlindLibrary{} & \librefThreeBlindLibrary{} \\
\quad Library minus fresh & \librefOneBlindLibraryMinusFresh{} (\librefOneBlindLibraryMinusFreshLo{} to \librefOneBlindLibraryMinusFreshHi{}) & \librefTwoBlindLibraryMinusFresh{} (\librefTwoBlindLibraryMinusFreshLo{} to \librefTwoBlindLibraryMinusFreshHi{}) & \librefThreeBlindLibraryMinusFresh{} (\librefThreeBlindLibraryMinusFreshLo{} to \librefThreeBlindLibraryMinusFreshHi{}) \\
\addlinespace
\multicolumn{4}{l}{\itshape CKGConv trained with a new pattern per graph, same seed} \\
\quad Library set & \librefOneNewLibrary{} & \librefTwoNewLibrary{} & \librefThreeNewLibrary{} \\
\quad Library minus predictor, paired & \librefOneNewMinusBlind{} (\librefOneNewMinusBlindLo{} to \librefOneNewMinusBlindHi{}) & \librefTwoNewMinusBlind{} (\librefTwoNewMinusBlindLo{} to \librefTwoNewMinusBlindHi{}) & \librefThreeNewMinusBlind{} (\librefThreeNewMinusBlindLo{} to \librefThreeNewMinusBlindHi{}) \\
\quad Library minus fresh & \librefOneNewLibraryMinusFresh{} (\librefOneNewLibraryMinusFreshLo{} to \librefOneNewLibraryMinusFreshHi{}) & \librefTwoNewLibraryMinusFresh{} (\librefTwoNewLibraryMinusFreshLo{} to \librefTwoNewLibraryMinusFreshHi{}) & \librefThreeNewLibraryMinusFresh{} (\librefThreeNewLibraryMinusFreshLo{} to \librefThreeNewLibraryMinusFreshHi{}) \\
\addlinespace
\multicolumn{4}{l}{\itshape CKGConv trained on the reused library} \\
\quad Library minus fresh & \reuseRunOneLibraryMinusFresh{} (\reuseRunOneLibraryMinusFreshLo{} to \reuseRunOneLibraryMinusFreshHi{}) & \reuseRunTwoLibraryMinusFresh{} (\reuseRunTwoLibraryMinusFreshLo{} to \reuseRunTwoLibraryMinusFreshHi{}) & \reuseRunThreeLibraryMinusFresh{} (\reuseRunThreeLibraryMinusFreshLo{} to \reuseRunThreeLibraryMinusFreshHi{}) \\
\quad Size-matched set & \librefOneMatched{} & \librefTwoMatched{} & \librefThreeMatched{} \\
\quad Familiar minus size-matched, paired & \librefOneFamiliarMinusMatched{} (\librefOneFamiliarMinusMatchedLo{} to \librefOneFamiliarMinusMatchedHi{}) & \librefTwoFamiliarMinusMatched{} (\librefTwoFamiliarMinusMatchedLo{} to \librefTwoFamiliarMinusMatchedHi{}) & \librefThreeFamiliarMinusMatched{} (\librefThreeFamiliarMinusMatchedLo{} to \librefThreeFamiliarMinusMatchedHi{}) \\
\quad Size-matched minus fresh & \librefOneMatchedMinusFresh{} (\librefOneMatchedMinusFreshLo{} to \librefOneMatchedMinusFreshHi{}) & \librefTwoMatchedMinusFresh{} (\librefTwoMatchedMinusFreshLo{} to \librefTwoMatchedMinusFreshHi{}) & \librefThreeMatchedMinusFresh{} (\librefThreeMatchedMinusFreshLo{} to \librefThreeMatchedMinusFreshHi{}) \\
\addlinespace
\multicolumn{4}{l}{\itshape The library's patterns} \\
\quad Mean size, in nodes & \librefOnePatternSize{} & \librefTwoPatternSize{} & \librefThreePatternSize{} \\
\quad Test graphs with a pattern under \matcherBig{} nodes & \librefOneSmallPatterns{} & \librefTwoSmallPatterns{} & \librefThreeSmallPatterns{} \\
\bottomrule
\end{tabular}

\end{table}

\begin{table}[t]
\centering
\caption{GRIT trained with its released configuration and the micro-batches described in Appendix~\ref{app:fresh}, for \hvCompletedGritSeedAEpochs{} epochs in every seed. Accuracy is class-weighted and pooled over each set. Familiar minus fresh compares two sets of graphs and is not paired. Differences from the pattern-blind predictor are paired by graph on the fresh set, with graph-bootstrap intervals. Table~\ref{tab:auc} gives every seed's AUC. Checkpoint epochs count from zero, and differences are computed before rounding.}
\label{tab:gritcompleted}
\small
\begin{tabular}{llll}
\toprule
Quantity & Seed 0 & Seed 1 & Seed 2 \\
\midrule
Official accuracy (\%) & \hvCompletedGritSeedAOfficialAccuracy{} & \hvCompletedGritSeedBOfficialAccuracy{} & \fiveGritSeedCOfficialTwo{} \\
Fresh accuracy (\%) & \hvCompletedGritSeedAFreshAccuracy{} & \hvCompletedGritSeedBFreshAccuracy{} & \fiveGritSeedCFreshTwo{} \\
Familiar accuracy (\%) & \hvCompletedGritSeedAFamiliarAccuracy{} & \hvCompletedGritSeedBFamiliarAccuracy{} & \fiveGritSeedCFamiliarTwo{} \\
Familiar minus fresh (points) & \hvCompletedGritSeedAFamiliarMinusFresh{} & \hvCompletedGritSeedBFamiliarMinusFresh{} & \fiveGritSeedCDropTwo{} \\
Fresh minus the pattern-blind predictor (points) & \hvCompletedGritSeedAFreshMinusBlind{} & \hvCompletedGritSeedBFreshMinusBlind{} & \fiveGritSeedCFreshMinusDensityTwo{} \\
Graph-paired 95\% interval & \hvCompletedGritSeedAFreshMinusBlindInterval{} & \hvCompletedGritSeedBFreshMinusBlindInterval{} & (\fiveGritSeedCFreshMinusDensityLoTwo{} to \fiveGritSeedCFreshMinusDensityHiTwo{}) \\
Familiar minus fresh AUC (points) & \hvCompletedGritSeedAFamiliarMinusFreshAuc{} & \hvCompletedGritSeedBFamiliarMinusFreshAuc{} & \gritSeedCAucGap{} \\
Selected checkpoint epoch & \hvCompletedGritSeedABestEpoch{} & \hvCompletedGritSeedBBestEpoch{} & \hvCompletedGritSeedCBestEpoch{} \\
\bottomrule
\end{tabular}

\end{table}

\begin{table}[t]
\centering
\caption{CKGConv trained and validated with a new pattern per graph, compared with the same-seed models trained on the released patterns. Accuracy is pooled class-weighted accuracy. Gains and differences from the pattern-blind predictor have paired graph-bootstrap intervals. Both the training and the validation distributions change; checkpoint epochs count from zero.}
\label{tab:ckgfreshcompleted}
\small
\begin{tabular}{llll}
\toprule
Quantity & Seed 0 & Seed 1 & Seed 2 \\
\midrule
New-pattern training, official accuracy (\%) & \hvCompletedCkgSeedAOfficialAccuracy{} & \hvCompletedCkgSeedBOfficialAccuracy{} & \hvCompletedCkgSeedCOfficialAccuracy{} \\
New-pattern training, fresh accuracy (\%) & \hvCompletedCkgSeedAFreshAccuracy{} & \hvCompletedCkgSeedBFreshAccuracy{} & \hvCompletedCkgSeedCFreshAccuracy{} \\
New-pattern training, familiar accuracy (\%) & \hvCompletedCkgSeedAFamiliarAccuracy{} & \hvCompletedCkgSeedBFamiliarAccuracy{} & \hvCompletedCkgSeedCFamiliarAccuracy{} \\
Released-pattern training, official accuracy (\%) & \hvCompletedCkgSeedAFullRecipeOfficial{} & \hvCompletedCkgSeedBFullRecipeOfficial{} & \hvCompletedCkgSeedCFullRecipeOfficial{} \\
Released-pattern training, fresh accuracy (\%) & \hvCompletedCkgSeedAFullRecipeFresh{} & \hvCompletedCkgSeedBFullRecipeFresh{} & \hvCompletedCkgSeedCFullRecipeFresh{} \\
Released-pattern training, familiar accuracy (\%) & \hvCompletedCkgSeedAFullRecipeFamiliar{} & \hvCompletedCkgSeedBFullRecipeFamiliar{} & \hvCompletedCkgSeedCFullRecipeFamiliar{} \\
Fresh accuracy gain (points) & \hvCompletedCkgSeedAFreshImprovement{} & \hvCompletedCkgSeedBFreshImprovement{} & \hvCompletedCkgSeedCFreshImprovement{} \\
Gain, graph-paired 95\% interval & \hvCompletedCkgSeedAFreshImprovementInterval{} & \hvCompletedCkgSeedBFreshImprovementInterval{} & \hvCompletedCkgSeedCFreshImprovementInterval{} \\
Fresh minus the pattern-blind predictor (points) & \hvCompletedCkgSeedAFreshMinusBlind{} & \hvCompletedCkgSeedBFreshMinusBlind{} & \hvCompletedCkgSeedCFreshMinusBlind{} \\
Difference, graph-paired 95\% interval & \hvCompletedCkgSeedAFreshMinusBlindInterval{} & \hvCompletedCkgSeedBFreshMinusBlindInterval{} & \hvCompletedCkgSeedCFreshMinusBlindInterval{} \\
Training epochs & \hvCompletedCkgSeedAEpochs{} & \hvCompletedCkgSeedBEpochs{} & \hvCompletedCkgSeedCEpochs{} \\
Selected checkpoint epoch & \hvCompletedCkgSeedABestEpoch{} & \hvCompletedCkgSeedBBestEpoch{} & \hvCompletedCkgSeedCBestEpoch{} \\
\bottomrule
\end{tabular}

\end{table}

\begin{table}[t]
\centering
\caption{Every seed behind Table~\ref{tab:fivemodels}, with the same columns to two decimals. Familiar minus fresh compares two sets of graphs and is not paired. The last column is a difference paired by graph on the fresh set, with its 95\% graph-bootstrap interval. Differences are computed before rounding.}
\label{tab:fivemodels_seeds}
\small
\begin{tabular}{lcccccc}
\toprule
 & & Official & Familiar & Fresh & Familiar & Fresh set, minus the \\
Model & Seed & test split & set & set & minus fresh & pattern-blind predictor \\
\midrule
GatedGCN-PE & \fiveGatedSeedANumber{} & \fiveGatedSeedAOfficialTwo{} & \fiveGatedSeedAFamiliarTwo{} & \fiveGatedSeedAFreshTwo{} & \fiveGatedSeedADropTwo{} & \fiveGatedSeedAFreshMinusDensityTwo{} (\fiveGatedSeedAFreshMinusDensityLoTwo{} to \fiveGatedSeedAFreshMinusDensityHiTwo{}) \\
 & \fiveGatedSeedBNumber{} & \fiveGatedSeedBOfficialTwo{} & \fiveGatedSeedBFamiliarTwo{} & \fiveGatedSeedBFreshTwo{} & \fiveGatedSeedBDropTwo{} & \fiveGatedSeedBFreshMinusDensityTwo{} (\fiveGatedSeedBFreshMinusDensityLoTwo{} to \fiveGatedSeedBFreshMinusDensityHiTwo{}) \\
 & \fiveGatedSeedCNumber{} & \fiveGatedSeedCOfficialTwo{} & \fiveGatedSeedCFamiliarTwo{} & \fiveGatedSeedCFreshTwo{} & \fiveGatedSeedCDropTwo{} & \fiveGatedSeedCFreshMinusDensityTwo{} (\fiveGatedSeedCFreshMinusDensityLoTwo{} to \fiveGatedSeedCFreshMinusDensityHiTwo{}) \\
 & \fiveGatedSeedDNumber{} & \fiveGatedSeedDOfficialTwo{} & \fiveGatedSeedDFamiliarTwo{} & \fiveGatedSeedDFreshTwo{} & \fiveGatedSeedDDropTwo{} & \fiveGatedSeedDFreshMinusDensityTwo{} (\fiveGatedSeedDFreshMinusDensityLoTwo{} to \fiveGatedSeedDFreshMinusDensityHiTwo{}) \\
\addlinespace
GRIT & \fiveGritSeedANumber{} & \fiveGritSeedAOfficialTwo{} & \fiveGritSeedAFamiliarTwo{} & \fiveGritSeedAFreshTwo{} & \fiveGritSeedADropTwo{} & \fiveGritSeedAFreshMinusDensityTwo{} (\fiveGritSeedAFreshMinusDensityLoTwo{} to \fiveGritSeedAFreshMinusDensityHiTwo{}) \\
 & \fiveGritSeedBNumber{} & \fiveGritSeedBOfficialTwo{} & \fiveGritSeedBFamiliarTwo{} & \fiveGritSeedBFreshTwo{} & \fiveGritSeedBDropTwo{} & \fiveGritSeedBFreshMinusDensityTwo{} (\fiveGritSeedBFreshMinusDensityLoTwo{} to \fiveGritSeedBFreshMinusDensityHiTwo{}) \\
 & \fiveGritSeedCNumber{} & \fiveGritSeedCOfficialTwo{} & \fiveGritSeedCFamiliarTwo{} & \fiveGritSeedCFreshTwo{} & \fiveGritSeedCDropTwo{} & \fiveGritSeedCFreshMinusDensityTwo{} (\fiveGritSeedCFreshMinusDensityLoTwo{} to \fiveGritSeedCFreshMinusDensityHiTwo{}) \\
\addlinespace
CKGConv & \fiveCkgSeedANumber{} & \fiveCkgSeedAOfficialTwo{} & \fiveCkgSeedAFamiliarTwo{} & \fiveCkgSeedAFreshTwo{} & \fiveCkgSeedADropTwo{} & \fiveCkgSeedAFreshMinusDensityTwo{} (\fiveCkgSeedAFreshMinusDensityLoTwo{} to \fiveCkgSeedAFreshMinusDensityHiTwo{}) \\
 & \fiveCkgSeedBNumber{} & \fiveCkgSeedBOfficialTwo{} & \fiveCkgSeedBFamiliarTwo{} & \fiveCkgSeedBFreshTwo{} & \fiveCkgSeedBDropTwo{} & \fiveCkgSeedBFreshMinusDensityTwo{} (\fiveCkgSeedBFreshMinusDensityLoTwo{} to \fiveCkgSeedBFreshMinusDensityHiTwo{}) \\
 & \fiveCkgSeedCNumber{} & \fiveCkgSeedCOfficialTwo{} & \fiveCkgSeedCFamiliarTwo{} & \fiveCkgSeedCFreshTwo{} & \fiveCkgSeedCDropTwo{} & \fiveCkgSeedCFreshMinusDensityTwo{} (\fiveCkgSeedCFreshMinusDensityLoTwo{} to \fiveCkgSeedCFreshMinusDensityHiTwo{}) \\
\addlinespace
GRASS & \fiveGrassSeedANumber{} & \fiveGrassSeedAOfficialTwo{} & \fiveGrassSeedAFamiliarTwo{} & \fiveGrassSeedAFreshTwo{} & \fiveGrassSeedADropTwo{} & \fiveGrassSeedAFreshMinusDensityTwo{} (\fiveGrassSeedAFreshMinusDensityLoTwo{} to \fiveGrassSeedAFreshMinusDensityHiTwo{}) \\
 & \fiveGrassSeedBNumber{} & \fiveGrassSeedBOfficialTwo{} & \fiveGrassSeedBFamiliarTwo{} & \fiveGrassSeedBFreshTwo{} & \fiveGrassSeedBDropTwo{} & \fiveGrassSeedBFreshMinusDensityTwo{} (\fiveGrassSeedBFreshMinusDensityLoTwo{} to \fiveGrassSeedBFreshMinusDensityHiTwo{}) \\
 & \fiveGrassSeedCNumber{} & \fiveGrassSeedCOfficialTwo{} & \fiveGrassSeedCFamiliarTwo{} & \fiveGrassSeedCFreshTwo{} & \fiveGrassSeedCDropTwo{} & \fiveGrassSeedCFreshMinusDensityTwo{} (\fiveGrassSeedCFreshMinusDensityLoTwo{} to \fiveGrassSeedCFreshMinusDensityHiTwo{}) \\
\addlinespace
PPGT & \fivePpgtSeedANumber{} & \fivePpgtSeedAOfficialTwo{} & \fivePpgtSeedAFamiliarTwo{} & \fivePpgtSeedAFreshTwo{} & \fivePpgtSeedADropTwo{} & \fivePpgtSeedAFreshMinusDensityTwo{} (\fivePpgtSeedAFreshMinusDensityLoTwo{} to \fivePpgtSeedAFreshMinusDensityHiTwo{}) \\
 & \fivePpgtSeedBNumber{} & \fivePpgtSeedBOfficialTwo{} & \fivePpgtSeedBFamiliarTwo{} & \fivePpgtSeedBFreshTwo{} & \fivePpgtSeedBDropTwo{} & \fivePpgtSeedBFreshMinusDensityTwo{} (\fivePpgtSeedBFreshMinusDensityLoTwo{} to \fivePpgtSeedBFreshMinusDensityHiTwo{}) \\
 & \fivePpgtSeedCNumber{} & \fivePpgtSeedCOfficialTwo{} & \fivePpgtSeedCFamiliarTwo{} & \fivePpgtSeedCFreshTwo{} & \fivePpgtSeedCDropTwo{} & \fivePpgtSeedCFreshMinusDensityTwo{} (\fivePpgtSeedCFreshMinusDensityLoTwo{} to \fivePpgtSeedCFreshMinusDensityHiTwo{}) \\
 & \fivePpgtSeedDNumber{} & \fivePpgtSeedDOfficialTwo{} & \fivePpgtSeedDFamiliarTwo{} & \fivePpgtSeedDFreshTwo{} & \fivePpgtSeedDDropTwo{} & \fivePpgtSeedDFreshMinusDensityTwo{} (\fivePpgtSeedDFreshMinusDensityLoTwo{} to \fivePpgtSeedDFreshMinusDensityHiTwo{}) \\
\bottomrule
\end{tabular}

\end{table}

\begin{table}[t]
\centering
\caption{Pooled ROC AUC, in percent, of every model and seed of Table~\ref{tab:fivemodels_seeds} and of the new-pattern models (CKGConv trained with a new pattern per graph), from the pattern-class probabilities pooled over the nodes of each set. Familiar minus fresh compares two sets of graphs and is not paired; it is computed before rounding and given to two decimals. The last two rows are the pattern-blind predictor and the sampler without the size prior, whose AUC was computed on the fresh set only (Appendix~\ref{app:likelihood}). No AUC was computed by pattern size.}
\label{tab:auc}
\small
\begin{tabular}{llcccc}
\toprule
 & & Official & Familiar & Fresh & Familiar \\
Model & Seed & test split & set & set & minus fresh \\
\midrule
GatedGCN-PE & \fiveGatedSeedANumber{} & \aucTableGatedSeedAOfficial{} & \aucTableGatedSeedAFamiliar{} & \aucTableGatedSeedAFresh{} & \aucTableGatedSeedADrop{} \\
 & \fiveGatedSeedBNumber{} & \aucTableGatedSeedBOfficial{} & \aucTableGatedSeedBFamiliar{} & \aucTableGatedSeedBFresh{} & \aucTableGatedSeedBDrop{} \\
 & \fiveGatedSeedCNumber{} & \aucTableGatedSeedCOfficial{} & \aucTableGatedSeedCFamiliar{} & \aucTableGatedSeedCFresh{} & \aucTableGatedSeedCDrop{} \\
 & \fiveGatedSeedDNumber{} & \aucTableGatedSeedDOfficial{} & \aucTableGatedSeedDFamiliar{} & \aucTableGatedSeedDFresh{} & \aucTableGatedSeedDDrop{} \\
\addlinespace
GRIT & \fiveGritSeedANumber{} & \aucTableGritSeedAOfficial{} & \aucTableGritSeedAFamiliar{} & \aucTableGritSeedAFresh{} & \aucTableGritSeedADrop{} \\
 & \fiveGritSeedBNumber{} & \aucTableGritSeedBOfficial{} & \aucTableGritSeedBFamiliar{} & \aucTableGritSeedBFresh{} & \aucTableGritSeedBDrop{} \\
 & \fiveGritSeedCNumber{} & \aucTableGritSeedCOfficial{} & \aucTableGritSeedCFamiliar{} & \aucTableGritSeedCFresh{} & \aucTableGritSeedCDrop{} \\
\addlinespace
CKGConv & \fiveCkgSeedANumber{} & \aucTableCkgSeedAOfficial{} & \aucTableCkgSeedAFamiliar{} & \aucTableCkgSeedAFresh{} & \aucTableCkgSeedADrop{} \\
 & \fiveCkgSeedBNumber{} & \aucTableCkgSeedBOfficial{} & \aucTableCkgSeedBFamiliar{} & \aucTableCkgSeedBFresh{} & \aucTableCkgSeedBDrop{} \\
 & \fiveCkgSeedCNumber{} & \aucTableCkgSeedCOfficial{} & \aucTableCkgSeedCFamiliar{} & \aucTableCkgSeedCFresh{} & \aucTableCkgSeedCDrop{} \\
\addlinespace
GRASS & \fiveGrassSeedANumber{} & \aucTableGrassSeedAOfficial{} & \aucTableGrassSeedAFamiliar{} & \aucTableGrassSeedAFresh{} & \aucTableGrassSeedADrop{} \\
 & \fiveGrassSeedBNumber{} & \aucTableGrassSeedBOfficial{} & \aucTableGrassSeedBFamiliar{} & \aucTableGrassSeedBFresh{} & \aucTableGrassSeedBDrop{} \\
 & \fiveGrassSeedCNumber{} & \aucTableGrassSeedCOfficial{} & \aucTableGrassSeedCFamiliar{} & \aucTableGrassSeedCFresh{} & \aucTableGrassSeedCDrop{} \\
\addlinespace
PPGT & \fivePpgtSeedANumber{} & \aucTablePpgtSeedAOfficial{} & \aucTablePpgtSeedAFamiliar{} & \aucTablePpgtSeedAFresh{} & \aucTablePpgtSeedADrop{} \\
 & \fivePpgtSeedBNumber{} & \aucTablePpgtSeedBOfficial{} & \aucTablePpgtSeedBFamiliar{} & \aucTablePpgtSeedBFresh{} & \aucTablePpgtSeedBDrop{} \\
 & \fivePpgtSeedCNumber{} & \aucTablePpgtSeedCOfficial{} & \aucTablePpgtSeedCFamiliar{} & \aucTablePpgtSeedCFresh{} & \aucTablePpgtSeedCDrop{} \\
 & \fivePpgtSeedDNumber{} & \aucTablePpgtSeedDOfficial{} & \aucTablePpgtSeedDFamiliar{} & \aucTablePpgtSeedDFresh{} & \aucTablePpgtSeedDDrop{} \\
\addlinespace
CKGConv, new pattern per graph & \aucTableCkgNewSeedANumber{} & \aucTableCkgNewSeedAOfficial{} & \aucTableCkgNewSeedAFamiliar{} & \aucTableCkgNewSeedAFresh{} & \aucTableCkgNewSeedADrop{} \\
 & \aucTableCkgNewSeedBNumber{} & \aucTableCkgNewSeedBOfficial{} & \aucTableCkgNewSeedBFamiliar{} & \aucTableCkgNewSeedBFresh{} & \aucTableCkgNewSeedBDrop{} \\
 & \aucTableCkgNewSeedCNumber{} & \aucTableCkgNewSeedCOfficial{} & \aucTableCkgNewSeedCFamiliar{} & \aucTableCkgNewSeedCFresh{} & \aucTableCkgNewSeedCDrop{} \\
\midrule
Pattern-blind predictor & & \aucTableBlindOfficial{} & \aucTableBlindFamiliar{} & \aucTableBlindFresh{} & \aucTableBlindDrop{} \\
Without the size prior & & & & \aucTableFirstBlindFresh{} & \\
\bottomrule
\end{tabular}

\end{table}

\paragraph{Class recalls.} For two seeds, we split the fresh-set scores into pattern recall (sensitivity) and host recall (specificity). GRIT scores below the pattern-blind predictor on both classes, with paired intervals below zero, and the new-pattern models have higher pattern recall and lower host recall than that predictor (Table~\ref{tab:classrecalls}). Their accuracy close to the predictor's therefore does not come from the same decisions.

\begin{table}[ht]
\centering
\caption{Exploratory differences in class recall from the pattern-blind predictor on the fresh set, in percentage points, with paired graph-bootstrap intervals. The CKGConv rows are the new-pattern models. The analysis covers seeds 0 and 1.}
\label{tab:classrecalls}
\footnotesize
\setlength{\tabcolsep}{4pt}
\begin{tabular}{llll}
\toprule
Model & Seed & Pattern recall difference & Host recall difference \\
\midrule
GRIT & 0 & \hvClassGritSeedASensitivity{} \hvClassGritSeedASensitivityInterval{} & \hvClassGritSeedASpecificity{} \hvClassGritSeedASpecificityInterval{} \\
GRIT & 1 & \hvClassGritSeedBSensitivity{} \hvClassGritSeedBSensitivityInterval{} & \hvClassGritSeedBSpecificity{} \hvClassGritSeedBSpecificityInterval{} \\
CKGConv & 0 & \hvClassCkgSeedASensitivity{} \hvClassCkgSeedASensitivityInterval{} & \hvClassCkgSeedASpecificity{} \hvClassCkgSeedASpecificityInterval{} \\
CKGConv & 1 & \hvClassCkgSeedBSensitivity{} \hvClassCkgSeedBSensitivityInterval{} & \hvClassCkgSeedBSpecificity{} \hvClassCkgSeedBSpecificityInterval{} \\
\bottomrule
\end{tabular}

\end{table}
\FloatBarrier

\section{The CLUSTER sampler}
\label{app:cluster}

\paragraph{The posterior.} Given a test graph's edges and its six labeled nodes, the generator gives a posterior over the assignments of the other nodes to the blocks. Edges are independent given the assignment, with probability \genClusterPIn{} within a block and \genClusterPOut{} across. The generator draws each block's size uniformly from \genClusterBlockMin{} to \genClusterBlockMax{}, assigns the nodes to the blocks at random, and reveals one node of each block, chosen uniformly within it. An assignment with block sizes $s_1, \dots, s_6$ therefore has prior probability proportional to $\prod_j s_j!$, the revealed nodes multiply it by $\prod_j 1/s_j$, and the posterior is proportional to $\prod_j (s_j-1)!$ times the likelihood of the edges, for sizes in the generator's range. The sampler keeps each labeled node in its block.

\paragraph{The sampler.} Each chain starts from the most probable block of each node under a mean-field fit and runs \clusterSweeps{} sweeps, discarding the first \clusterBurn{}. A sweep resamples each unlabeled node from its conditional given the others, including the factor that the block sizes contribute to the posterior. It then proposes, for every pair of blocks, to exchange all of their unlabeled nodes. The sizes are exchanged with the nodes, so that factor does not change, and the proposal is accepted by the Metropolis rule on the edges of the two blocks' labeled nodes, the only edges whose likelihood changes. Without these swaps, a sampler that moves one node at a time cannot move a whole block.

\paragraph{Prediction and diagnostics.} A node's block probabilities are estimated by their frequencies over the kept sweeps of \clusterChains{} chains, and its predicted block is the one with the largest estimate. The six classes have the same expected size, but their realized sizes vary, and the rule is not tuned to class-weighted accuracy, so the sampler's score is a lower estimate of the score the generator supports. To measure agreement, for each chain after the first we take the share of a graph's nodes whose most probable block under that chain matches their most probable block under the earlier chains pooled together. A graph's chain agreement is the lowest of these shares: \clusterChainAgreementMedian{} on the median test graph and \clusterChainAgreementMin{} on the worst. For two chains it is the share of nodes on which they agree. Under the three conventions of Appendix~\ref{app:metric}, the sampler scores \clusterCeilingPooledTwo{} pooled, \clusterCeilingBatchTwo{} as a batch mean, and \clusterPlainTwo{} in plain accuracy.

\paragraph{The sampler without the revealed nodes' factor.} Adding the revealed nodes' factor to the prior $\prod_j s_j!$ changes the sampler's prediction at \clusterFactorChangedPct\% of the nodes but not its pooled score, whose paired difference has an interval around zero. Running \clusterChains{} chains for twice as many sweeps, as the sampler above does, adds \clusterLongerDiff{} points (Table~\ref{tab:clusterfactor}).

\begin{table}[ht]
\centering
\caption{The CLUSTER sampler with and without the revealed nodes' factor, on the test split. Each difference is paired with the row above it, with a 95\% interval.}
\label{tab:clusterfactor}
\small
\begin{tabular}{lll}
\toprule
Sampler & Chains and sweeps & Score or difference \\
\midrule
Prior $\prod_j s_j!$ alone & \clusterFirstChains{} chains of \clusterFirstSweeps{} sweeps, \clusterFirstBurn{} discarded & \clusterFirstPooled{} (\clusterFirstLo{} to \clusterFirstHi{}) \\
Adding the revealed nodes' factor & the same & \clusterFactorDiff{} (\clusterFactorDiffLo{} to \clusterFactorDiffHi{}) \\
Longer sampling & \clusterChains{} chains, twice the sweeps & \clusterLongerDiff{} (\clusterLongerDiffLo{} to \clusterLongerDiffHi{}) \\
\bottomrule
\end{tabular}
\end{table}

\section{Minesweeper inference}
\label{app:mines}

Deciding whether any placement of mines reproduces a given set of revealed counts is NP-complete \citep{kaye2000minesweeper}, and \citet{bonet2016factored} compare exact and approximate inference methods for playing the game. We use two approximate methods on the released board, belief propagation and a strip sampler.

\paragraph{The model.} The board is a grid of \minesBoardSide{} by \minesBoardSide{} cells. The generator places mines on a fixed share \minesPrior{} of the cells, chosen uniformly at random, and \minesRevealed{} cells show the number of mines among their neighbors, capped at \minesCap{}. Both methods approximate this law by independent mines with probability \minesPrior{}; the posterior over boards given the revealed counts is then that prior restricted to the boards that reproduce every count. The output is a ranking of cells computed from the features under this approximate model, and we score that ranking directly on the released board.

\paragraph{Belief propagation.} On the factor graph with one binary variable per cell and one factor per revealed count, damped belief propagation, with damping \minesBpDamping{}, ran from the features alone for \minesBpIterations{} sweeps and stopped with a last change of \minesBpLastChange{}. With each mask's training labels fixed as evidence, it ran to the same limit on all \minesSplits{} masks. The two runs use different information. A graph network's fitted weights can depend on the training labels even when its inputs are features alone, so the format of its inputs does not make our features-only score a bound on it.

\paragraph{The strip sampler.} The state is a full board. An update picks a strip of \minesExactWidth{} adjacent columns and redraws it, over the whole height of the board, from its exact conditional under the independent-mine model given every cell outside the strip. Along the strip that conditional is a chain in the row index whose state is a pair of consecutive rows of the strip, so forward filtering and backward sampling draw it exactly \citep{carter1994gibbs}. The same forward and backward passes give each strip cell's exact conditional probability, which is what we average. A sweep covers the board with strips at a random offset, first by columns and then by rows. Every run checks the code against brute-force enumeration on small boards: a strip as wide as the board, which is exact enumeration, matches it to machine precision, and so do single strips' conditionals (Table~\ref{tab:minescriteria}).

\paragraph{The chains.} Each chain starts from a board drawn from the prior and first anneals a soft form of the constraints until no count is violated, penalizing each revealed count by its distance from the board's count. The penalty's weight starts at \minesAnnealStart{} and is multiplied by \minesAnnealGrowth{} after each sweep, up to a cap of \minesAnnealCap{}; after \minesAnnealStuck{} sweeps at the cap it is reset to \minesAnnealReset{} and rises again. The chain then uses the strip updates under the hard count constraints and discards its first \minesExactBurn{} sweeps. We set up \minesExactChainsWord{} chains to run for \minesExactHours{} hours, a budget fixed in advance. The code ran one chain per processor core and the machine had \minesExactRanWord{}, so the first \minesExactRanWord{} chains ran for the whole budget and the other \minesExactQueuedWord{} waited for a free core; they started only after the budget had run out, stopped after one annealing sweep, and drew no samples. Which chains ran depended on their order, fixed before the run, not on their output. Each of the \minesExactRanWord{} that ran reached a valid board within about \minesExactAnnealMaxRound{} annealing sweeps and then sampled about \minesExactSweepsMinRound{} to \minesExactSweepsMaxRound{} sweeps. The reported score is the ROC AUC of their averaged probabilities on each of the ten test masks: \minesExactAuc{} on average, and \minesExactSplitLo{} to \minesExactSplitHi{} over the masks. The ten masks overlap on one board, so their spread measures variation across the masks of this board.

\paragraph{Convergence.} The chains did not converge: \minesExactCritFailedWord{} of the six criteria we set before the run are not met (Table~\ref{tab:minescriteria}), with split $\hat R$ as \citet{vehtari2021rhat} define it. The chains' AUCs range from \minesExactChainLo{} to \minesExactChainHi{}, and their probabilities differ from cell to cell, so the chains sit in different parts of the posterior. Strip moves are not ergodic in general: on small boards with narrow strips, chains froze in different configurations, and a cell's probability was off by up to \minesSmallBoardErr{}. The averaged probabilities remain a function of the features and the generator alone. The test labels enter only the per-sweep AUC trace that the diagnostics use, never the probabilities or any choice about the chains. A public issue reports that the released labels align exactly with a permutation drawn from a fixed seed, so they could be recovered from that seed \citep{issue5mines}. Our inference reads only the revealed counts and the grid, and none of our code draws a permutation of the board's cells.

\paragraph{Ranking by marginals.} For any two test cells, the posterior probability that the first holds a mine and the second does not exceeds the reverse exactly when the first cell's marginal is higher, since the two probabilities differ by the difference of the marginals. Ranking by exact posterior marginals therefore maximizes the expected number of correctly ordered pairs. It also maximizes expected ROC AUC when the AUC's denominator, the product of the numbers of positive and negative test cells, is fixed given the conditioning information. That denominator can vary across posterior boards for a test mask, so for normalized AUC the argument is approximate, as are our marginals. The sampler's score is what inference from the features alone achieves on the released board.

\begin{table}[h]
\centering
\caption{Convergence of the strip sampler: the six criteria we set before the run, one check per row, with the required and observed values. A criterion is met when all of its checks are. Differences are computed before rounding.}
\label{tab:minescriteria}
\small
\begin{tabular}{l>{\raggedright\arraybackslash}p{0.3\linewidth}lll}
\toprule
Criterion & Check & Required & Observed & Met \\
\midrule
The code & full-width strip against enumeration & within \regCritCode{} & \minesExactCheckFull{} & \minesExactMetCheckFull{} \\
 & single strips against their conditionals & within \regCritCode{} & \minesExactCheckStrip{} & \minesExactMetCheckStrip{} \\
\addlinespace
Valid chains & chains that reach a valid board & at least \regCritValid{} of \minesExactChains{} & \minesExactRan{} ran; all \minesExactValidChains{} reached one & \minesExactCritValid{} \\
\addlinespace
Split $\hat{R}$ & mine count & at most \regCritRhat{} & \minesExactRhatMines{} & \minesExactMetRhatMines{} \\
 & AUC trace & at most \regCritRhat{} & \minesExactRhat{} & \minesExactMetRhatAuc{} \\
\addlinespace
Effective sample size & mine count & at least \regCritEss{} & \minesExactEssMines{} & \minesExactMetEssMines{} \\
 & AUC trace & at least \regCritEss{} & \minesExactEssAuc{} & \minesExactMetEssAuc{} \\
\addlinespace
Chain agreement & spread of the chains' AUCs & at most \regCritChains{} & \minesExactChainRange{} & \minesExactMetChainRange{} \\
 & AUC trace, first half against second & within \regCritHalves{} & \minesExactHalfDiff{} & \minesExactMetHalfDiff{} \\
\addlinespace
Node agreement & between-chain SD of marginals, 99th percentile & at most \regCritNodes{} & \minesExactNodeSd{} & \minesExactCritNodes{} \\
\bottomrule
\end{tabular}

\end{table}

\section{Papers from 2024 to 2026 that report PATTERN results}
\label{app:papers}

The inventory covers the papers from 2024 on that Semantic Scholar lists as citing the benchmark's paper, GRIT, or GraphGPS \citep{dwivedi2023benchmarking,ma2023grit,rampasek2022gps}. We searched the full text of each one we could obtain for PATTERN, read the results tables of those that mention it, and kept the papers that report PATTERN results in a table of their own. Table~\ref{tab:papers} lists the \appGPapers{} papers, with the score we take from each and the name it gives the metric. For a method paper we take the best result of a proposed method from its main-text PATTERN classification comparison, or from an appendix comparison when the main text has none, and mark that source. We exclude isolated ablation results. For a benchmark paper we apply the same rule to the models its authors evaluated. Of the \appGMethodPapers{} method papers, \appGAboveGrit{} report selected results above GRIT's \pubPatternGRITTwo{}. Across the inventory, \appGMetricNamed{} papers name the metric as weighted, class-weighted, or balanced accuracy, and none says how it is pooled. The active-learning paper reports the area under a learning curve and is counted neither among the method papers nor among those that name the metric, although the accuracy behind its curve is balanced accuracy. Since the papers name the metric differently, we cannot tell whether their numbers come from one implementation of it. The search misses papers that cite none of the three and papers without an open full text.

\begin{figure}[ht]
\centering
\includegraphics[width=\linewidth]{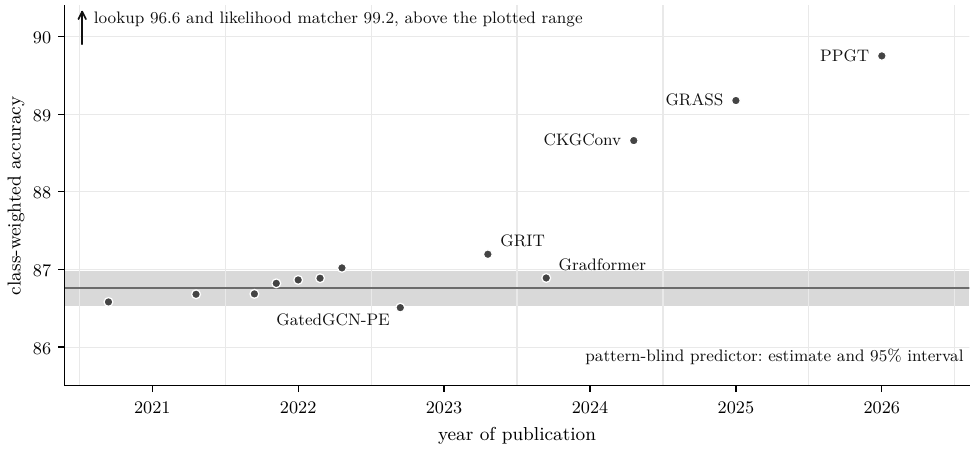}
\caption{Published class-weighted accuracy on PATTERN by year of publication; Table~\ref{tab:ladder} names every point, and points of one year are spread across its column. The gray band is the pattern-blind predictor's 95\% graph-bootstrap interval around its score. The lookup (\lookupRealPooled{}), the matcher (\realMatcherPooled{}) and the multilayer perceptron (\pubPatternMLP{}) lie outside the plotted range (Table~\ref{tab:ladder}).}
\label{fig:leaderboard}
\end{figure}

Figure~\ref{fig:leaderboard} plots the published scores selected in Table~\ref{tab:ladder} by year of publication. The scores published up to 2023 lie close to the pattern-blind predictor's, and the three highest come from later papers.

{\setlength{\LTcapwidth}{\linewidth}
\setlength{\tabcolsep}{4pt}
\newcommand{\papersCaption}{Papers from 2024 to 2026 that report PATTERN results, in decreasing order of the selected number, with an arXiv identifier where available. The source is the main-text PATTERN classification comparison, or the appendix comparison, marked (appendix), when the main text has none. For method papers we select the best result of a proposed method; for papers marked (benchmark), which compare existing models or encodings, the best result of a model the authors evaluated. Isolated ablation results are excluded. Numbers are in percent, rounded half up to two decimals from the printed value. The metric is the name the paper gives PATTERN's metric in its results or dataset table, the one that states a weighting where the two differ; where no table names it, or only the text states a weighting, it is the text's name, marked (text). The paper marked (other) reports active learning under a labeling budget: its number is the area under its learning curve (AULC), and it is not counted among the method papers. $^\dagger$QPE's best model; QPE's numbers are left out of Table~\ref{tab:ladder} (Appendix~\ref{app:metric}).}
\begin{longtable}{>{\footnotesize\raggedright\arraybackslash}p{0.3\linewidth}>{\footnotesize}l>{\footnotesize\raggedright\arraybackslash}p{0.14\linewidth}>{\footnotesize\raggedright\arraybackslash}p{0.2\linewidth}>{\footnotesize}r}
\caption{\papersCaption}\label{tab:papers}\\
\toprule
Model & arXiv & Venue & Metric as named & PATTERN \\
\midrule
\endfirsthead
\multicolumn{5}{l}{\footnotesize Table~\ref{tab:papers}, continued} \\
\toprule
Model & arXiv & Venue & Metric as named & PATTERN \\
\midrule
\endhead
GRIT 2-QiRW (QPE) & 2406.06547 & ICML & Weighted Accuracy & \appGRowA{}$^\dagger$ \\
PPGT & 2504.12588 & TMLR & Weighted Accuracy & \appGRowB{} \\
GRASS & 2407.05649 & ICLR & Accuracy & \appGRowC{} \\
CKGConv & 2404.13604 & ICML & Weighted Accuracy & \appGRowD{} \\
GSM++ (MoT) & 2411.15671 & ICML & Accuracy & \appGRowE{} \\
GraphMinNet & 2502.00282 & Neural Networks & Accuracy & \appGRowF{} \\
GSSC & 2406.05815 & IEEE TPAMI & Accuracy & \appGRowG{} \\
EHDGT \citep{mu2025graph} & none & Scientific Reports & Accuracy & \appGRowH{} \\
GRIT-SPSE & 2502.09365 & ICML & Accuracy & \appGRowI{} \\
GRIT with RRWP (their rerun) (benchmark) & 2411.12732 & KDD & Accuracy & \appGRowJ{} \\
TANGO-GPS & 2508.05070 & ICML & Accuracy & \appGRowK{} \\
GMN & 2402.08678 & KDD & Accuracy & \appGRowL{} \\
GatedGCN+ & 2502.09263 & ICML & Accuracy & \appGRowM{} \\
GEAET & 2405.21061 & ICML & Accuracy(\%) & \appGRowN{} \\
NeuralWalker & 2406.03386 & ICLR & Accuracy & \appGRowO{} \\
Gradformer & 2404.15729 & IJCAI & accuracy & \appGRowP{} \\
GIG-GatedGCN & 2407.00696 & IJCV & Acc(\%) & \appGRowQ{} \\
OGE-Aug & 2410.09737 & KDD & Acc & \appGRowR{} \\
GPS + RandAlign & 2404.09774 & arXiv & weighted accuracy (text) & \appGRowS{} \\
SANv2 (benchmark) & 2608.02558 & arXiv & Accuracy & \appGRowT{} \\
SFi-Former+ & 2504.20666 & ICMR & Accuracy & \appGRowU{} \\
GPS-GCCM & 2605.05689 & arXiv & Accuracy & \appGRowV{} \\
GPS+CoRe-GD* (appendix) & 2402.06706 & ICLR & not named & \appGRowW{} \\
GPS with QUEST attention & 2604.00199 & ICLR & class-weighted classification accuracy (text) & \appGRowX{} \\
EGTAS & 2405.19779 & IEEE TNNLS & \#Acc & \appGRowY{} \\
Eigenformer (feat.) & 2401.17791 & arXiv & Wt. Acc. & \appGRowZ{} \\
Graph-Mamba (appendix) & 2402.00789 & arXiv & Accuracy & \appGRowBa{} \\
Grafourierformer & 2504.19740 & ICASSP & accuracy & \appGRowBb{} \\
DARTS-GT & 2510.14336 & arXiv & Acc & \appGRowBc{} \\
TIGT & 2402.02005 & GRaM workshop & Weighted Accuracy & \appGRowBd{} \\
``Ours + GatedGCN'' (16 layers) & 2404.09809 & Applied Soft Computing & weighted accuracy (text) & \appGRowBe{} \\
WIRE with Performer attention & 2509.22259 & ICML & Accuracy & \appGRowBf{} \\
GatedGCN + CLFE & 2405.04755 & Applied Soft Computing & weighted average accuracy (text) & \appGRowBg{} \\
SINC-GCN & 2511.11046 & TMLR & accuracy (text) & \appGRowBh{} \\
SIR-GCN & 2403.12529 & TMLR & accuracy (text) & \appGRowBi{} \\
GRF++ & 2510.07716 & ICML & not named & \appGRowBj{} \\
TypiClust with sum aggregation (appendix) (other) & 2606.04647 & ECML PKDD & AULC of macro-averaged accuracy & \appGRowBk{} \\
\bottomrule
\end{longtable}
}

\FloatBarrier
\section{Without the generator's values}
\label{app:withoutgen}

\paragraph{A lookup without the generator.} The lookup of Section~\ref{sec:recall} reads the set of patterns off the training labels and runs the matcher's search for induced copies with matching features (Appendix~\ref{app:likelihood}). Patterns of \matcherBig{} nodes or more are tried from the largest down, and the first copy found ends the search. If none is found, the lookup lists the copies of every smaller pattern over the whole graph, up to \matcherCap{} for each pattern, and labels the largest copy, the earliest found among copies of equal size. The lookup with a degree shortlist seeks the smaller patterns only among the \matcherTop{} nodes of highest degree, as many as the matcher's shortlist holds. Neither lookup reads a value from the generator's code or fits a model. The largest-match rule labels copies in the same way within the matcher's shortlist, which the sampler without the size prior ranks with the generator's values. Table~\ref{tab:recall} gives all four methods on the three sets.

Where a pattern of \matcherBig{} nodes or more is found, the shortlist plays no part, so both lookups should make the largest-match rule's predictions there. They do, on every such graph of the official test split and of the familiar set; on the fresh set no such pattern is found. The lookup's losses therefore come from the smaller patterns. On the official test split a pattern of \matcherBig{} nodes or more is found in \lookupRealFoundBigPct{}\% of the graphs, and on the others the lookup recovers fewer planted sets exactly than the largest-match rule (Table~\ref{tab:lookupchecks}). Paired by graph, the lookup scores below the matcher and slightly below the largest-match rule, with intervals below zero. The lookup with a degree shortlist does not differ detectably from the largest-match rule, so a shortlist ranked by degree does about as well as one ranked with the generator's values.

\begin{table}[ht]
\centering
\caption{The lookup beside the matcher and the largest-match rule on the official test split. Exact recovery is the share of graphs whose planted set is labeled exactly. Differences are paired by graph, with 95\% intervals.}
\label{tab:lookupchecks}
\small
\begin{tabular}{ll}
\toprule
Graphs in which a pattern of \matcherBig{} nodes or more is found & \lookupRealFoundBigPct{}\% \\
Exact recovery by the lookup on those graphs & \lookupRealExactPctBig{}\% \\
Exact recovery by the lookup on the other graphs & \lookupRealExactPctSmall{}\% \\
Exact recovery by the largest-match rule on the other graphs & \realLargestExactPctSmall{}\% \\
\midrule
Lookup minus matcher & \lookupRealMinusMatcher{} (\lookupRealMinusMatcherLo{} to \lookupRealMinusMatcherHi{}) \\
Lookup minus largest-match rule & \lookupRealMinusLargest{} (\lookupRealMinusLargestLo{} to \lookupRealMinusLargestHi{}) \\
Lookup with a degree shortlist minus largest-match rule & \lookupDegreeRealMinusLargestTwo{} (\lookupDegreeRealMinusLargestLoTwo{} to \lookupDegreeRealMinusLargestHiTwo{}) \\
\bottomrule
\end{tabular}
\end{table}

\begin{table}[ht]
\centering
\caption{Four ways of looking up the training patterns, on PATTERN's official test split and on the familiar and fresh sets of Section~\ref{sec:fresh}: class-weighted accuracy pooled over each set, with 95\% graph-bootstrap intervals. All four seek patterns of \matcherBig{} nodes or more over the whole graph. The two lookups and the largest-match rule label the largest copy found, and the matcher labels the copy with the highest likelihood score (Appendix~\ref{app:likelihood}). Only the two lookups use nothing from the generator's code.}
\label{tab:recall}
\small
\resizebox{\linewidth}{!}{\begin{tabular}{>{\raggedright\arraybackslash}p{0.17\linewidth}>{\raggedright\arraybackslash}p{0.25\linewidth}ccc}
\toprule
Method & Seeks patterns under \matcherBig{} nodes among & Official test split & Familiar set & Fresh set \\
\midrule
Lookup & the whole graph & \lookupRealPooled{} (\lookupRealLo{} to \lookupRealHi{}) & \lookupFamiliarPooled{} (\lookupFamiliarLo{} to \lookupFamiliarHi{}) & \lookupFreshPooled{} (\lookupFreshLo{} to \lookupFreshHi{}) \\
Lookup with a degree shortlist & the \matcherTop{} nodes of highest degree & \lookupDegreeRealPooled{} (\lookupDegreeRealLo{} to \lookupDegreeRealHi{}) & \lookupDegreeFamiliarPooled{} (\lookupDegreeFamiliarLo{} to \lookupDegreeFamiliarHi{}) & \lookupDegreeFreshPooled{} (\lookupDegreeFreshLo{} to \lookupDegreeFreshHi{}) \\
Largest-match rule & the \matcherTop{} nodes ranked highest by the sampler without the size prior & \realLargestPooled{} (\realLargestLo{} to \realLargestHi{}) & \familiarLargestPooled{} (\familiarLargestLo{} to \familiarLargestHi{}) & \freshLargestPooled{} (\freshLargestLo{} to \freshLargestHi{}) \\
Likelihood matcher & the same nodes as the largest-match rule & \realMatcherPooled{} (\realMatcherLo{} to \realMatcherHi{}) & \familiarMatcherPooled{} (\familiarMatcherLo{} to \familiarMatcherHi{}) & \freshMatcherPooled{} (\freshMatcherLo{} to \freshMatcherHi{}) \\
\bottomrule
\end{tabular}
}
\end{table}

\paragraph{The structural ladder.} Each predictor sees statistics computed from a graph's edges alone, in rungs that each add to the one before (Table~\ref{tab:structural}). The first rung holds a node's degree, the graph's size, and the degree as a share of the other nodes. The second adds the degree's z-score and its rank within the graph. The third adds one-hop statistics: the mean, standard deviation, minimum, and maximum of the neighbors' degree shares, the local clustering coefficient, the number of triangles, and the share of the node's neighbors among the graph's nodes of highest degree. The fourth adds the neighbors' means of these statistics and the share of the graph within two hops of the node. A last rung adds the node's own feature and the number of its neighbors with each feature value; it is the only rung that could carry a pattern's identity. Logistic regression with an $L_2$ penalty on standardized statistics, and scikit-learn's histogram gradient-boosted trees \citep{pedregosa2011scikit} at their default settings except for more boosting rounds, are trained on the official training split. Each labels a node as pattern when its probability exceeds the threshold that maximizes pooled class-weighted accuracy on the official validation split. A rule without training labels each graph's nodes of highest degree, as many as the training split's share of pattern nodes gives; the same share sets the number of highest-degree nodes in the third rung. Every predictor is scored on the official test split and the familiar and fresh sets.

No predictor in the ladder scores above the pattern-blind predictor on the fresh set. The best of those without node features, two-hop logistic regression, differs from it by \ladderBestFreshMinusDensityTwo{} points (paired 95\% interval \ladderBestFreshMinusDensityLoTwo{} to \ladderBestFreshMinusDensityHiTwo{}). No predictor, with node features or without, loses more than \ladderLargestRealMinusFreshTwo{} points from the official test split to the fresh set.

\paragraph{Held-out patterns.} Predictors on node statistics show no detectable use of the training patterns' identity. For the held-out test of Section~\ref{sec:heldout}, we retrained logistic regression and boosted trees on the two-hop rungs, with and without node features, without the graphs that carry the \heldPatterns{} held-out patterns, and chose their thresholds on the remaining validation graphs. No retrained predictor's score differs by more than \ladderHeldLargestHMinusKeptTwo{} points between the test graphs that carry a held-out pattern and the others, while CKGConv, retrained without the same patterns, scores several points lower on those graphs (Table~\ref{tab:heldout}).

\begin{table}[ht]
\centering
\caption{The structural ladder on PATTERN: class-weighted accuracy pooled over each set, with 95\% graph-bootstrap intervals. Each rung adds statistics to the one above it. The top-degree rule labels each graph's nodes of highest degree, as many as the training split's share of pattern nodes gives; logistic regression and boosted trees are trained on the official training split, with thresholds chosen on the official validation split. The last row is the pattern-blind predictor of Section~\ref{sec:blind}.}
\label{tab:structural}
\small
\resizebox{\linewidth}{!}{\begin{tabular}{llccc}
\toprule
Predictor & Node statistics & Official test split & Familiar set & Fresh set \\
\midrule
Top-degree rule & degree & \ladderTopDegreeRealPooled{} (\ladderTopDegreeRealLo{} to \ladderTopDegreeRealHi{}) & \ladderTopDegreeFamiliarPooled{} (\ladderTopDegreeFamiliarLo{} to \ladderTopDegreeFamiliarHi{}) & \ladderTopDegreeFreshPooled{} (\ladderTopDegreeFreshLo{} to \ladderTopDegreeFreshHi{}) \\
\addlinespace
Logistic regression & degree & \ladderLogisticDegreeRealPooled{} (\ladderLogisticDegreeRealLo{} to \ladderLogisticDegreeRealHi{}) & \ladderLogisticDegreeFamiliarPooled{} (\ladderLogisticDegreeFamiliarLo{} to \ladderLogisticDegreeFamiliarHi{}) & \ladderLogisticDegreeFreshPooled{} (\ladderLogisticDegreeFreshLo{} to \ladderLogisticDegreeFreshHi{}) \\
 & degree and its rank in the graph & \ladderLogisticDegreeRankRealPooled{} (\ladderLogisticDegreeRankRealLo{} to \ladderLogisticDegreeRankRealHi{}) & \ladderLogisticDegreeRankFamiliarPooled{} (\ladderLogisticDegreeRankFamiliarLo{} to \ladderLogisticDegreeRankFamiliarHi{}) & \ladderLogisticDegreeRankFreshPooled{} (\ladderLogisticDegreeRankFreshLo{} to \ladderLogisticDegreeRankFreshHi{}) \\
 & one hop & \ladderLogisticOneHopRealPooled{} (\ladderLogisticOneHopRealLo{} to \ladderLogisticOneHopRealHi{}) & \ladderLogisticOneHopFamiliarPooled{} (\ladderLogisticOneHopFamiliarLo{} to \ladderLogisticOneHopFamiliarHi{}) & \ladderLogisticOneHopFreshPooled{} (\ladderLogisticOneHopFreshLo{} to \ladderLogisticOneHopFreshHi{}) \\
 & two hops & \ladderLogisticTwoHopRealPooled{} (\ladderLogisticTwoHopRealLo{} to \ladderLogisticTwoHopRealHi{}) & \ladderLogisticTwoHopFamiliarPooled{} (\ladderLogisticTwoHopFamiliarLo{} to \ladderLogisticTwoHopFamiliarHi{}) & \ladderLogisticTwoHopFreshPooled{} (\ladderLogisticTwoHopFreshLo{} to \ladderLogisticTwoHopFreshHi{}) \\
 & two hops and node features & \ladderLogisticTwoHopFeaturesRealPooled{} (\ladderLogisticTwoHopFeaturesRealLo{} to \ladderLogisticTwoHopFeaturesRealHi{}) & \ladderLogisticTwoHopFeaturesFamiliarPooled{} (\ladderLogisticTwoHopFeaturesFamiliarLo{} to \ladderLogisticTwoHopFeaturesFamiliarHi{}) & \ladderLogisticTwoHopFeaturesFreshPooled{} (\ladderLogisticTwoHopFeaturesFreshLo{} to \ladderLogisticTwoHopFeaturesFreshHi{}) \\
\addlinespace
Boosted trees & degree & \ladderBoostedDegreeRealPooled{} (\ladderBoostedDegreeRealLo{} to \ladderBoostedDegreeRealHi{}) & \ladderBoostedDegreeFamiliarPooled{} (\ladderBoostedDegreeFamiliarLo{} to \ladderBoostedDegreeFamiliarHi{}) & \ladderBoostedDegreeFreshPooled{} (\ladderBoostedDegreeFreshLo{} to \ladderBoostedDegreeFreshHi{}) \\
 & degree and its rank in the graph & \ladderBoostedDegreeRankRealPooled{} (\ladderBoostedDegreeRankRealLo{} to \ladderBoostedDegreeRankRealHi{}) & \ladderBoostedDegreeRankFamiliarPooled{} (\ladderBoostedDegreeRankFamiliarLo{} to \ladderBoostedDegreeRankFamiliarHi{}) & \ladderBoostedDegreeRankFreshPooled{} (\ladderBoostedDegreeRankFreshLo{} to \ladderBoostedDegreeRankFreshHi{}) \\
 & one hop & \ladderBoostedOneHopRealPooled{} (\ladderBoostedOneHopRealLo{} to \ladderBoostedOneHopRealHi{}) & \ladderBoostedOneHopFamiliarPooled{} (\ladderBoostedOneHopFamiliarLo{} to \ladderBoostedOneHopFamiliarHi{}) & \ladderBoostedOneHopFreshPooled{} (\ladderBoostedOneHopFreshLo{} to \ladderBoostedOneHopFreshHi{}) \\
 & two hops & \ladderBoostedTwoHopRealPooled{} (\ladderBoostedTwoHopRealLo{} to \ladderBoostedTwoHopRealHi{}) & \ladderBoostedTwoHopFamiliarPooled{} (\ladderBoostedTwoHopFamiliarLo{} to \ladderBoostedTwoHopFamiliarHi{}) & \ladderBoostedTwoHopFreshPooled{} (\ladderBoostedTwoHopFreshLo{} to \ladderBoostedTwoHopFreshHi{}) \\
 & two hops and node features & \ladderBoostedTwoHopFeaturesRealPooled{} (\ladderBoostedTwoHopFeaturesRealLo{} to \ladderBoostedTwoHopFeaturesRealHi{}) & \ladderBoostedTwoHopFeaturesFamiliarPooled{} (\ladderBoostedTwoHopFeaturesFamiliarLo{} to \ladderBoostedTwoHopFeaturesFamiliarHi{}) & \ladderBoostedTwoHopFeaturesFreshPooled{} (\ladderBoostedTwoHopFeaturesFreshLo{} to \ladderBoostedTwoHopFeaturesFreshHi{}) \\
\midrule
Pattern-blind predictor & & \realDensityPooled{} (\realDensityLo{} to \realDensityHi{}) & \familiarDensityPooled{} (\familiarDensityLo{} to \familiarDensityHi{}) & \freshDensityPooled{} (\freshDensityLo{} to \freshDensityHi{}) \\
\bottomrule
\end{tabular}
}
\end{table}

\paragraph{Values estimated from the training split.} We reran the samplers of PATTERN and CLUSTER, and Minesweeper's belief propagation, with every value they take from the generator's code replaced by an estimate from the training split (Table~\ref{tab:estimated}). These runs keep the form of each generator and estimate only its values. For PATTERN, the pattern's internal density and its rate of edges to the host are pooled over the training graphs, and the pattern sizes and the share of pattern nodes are read off the labels. A block model with one edge probability within blocks and one across is fitted by mean field to each training host, with the number of blocks chosen for each host by the integrated classification likelihood \citep{daudin2008mixture}. The host's blocks cannot be recovered from graphs this small: on \estPatternBlocksTwoGraphs{} of the \estPatternGraphs{} training hosts the estimator chooses the fewest blocks it allows, and it fits nearly the same probability within blocks as across them. With these values the sampler treats the host as one random graph, and its score differs from the run with the code's values by \estRealDensityMinusKnownTwo{} points on the official test split and by \estFreshDensityMinusKnownTwo{} on the fresh set. A second run that fixes the code's \genPatternHostBlocks{} host blocks gives similar differences. For CLUSTER, the edge probabilities, the number of blocks, and the range of block sizes estimated from the labeled training graphs agree with the code's values, and the sampler's score differs from the run with the code's values by \estClusterMinusCeilingTwo{} points. For Minesweeper, the mine rate estimated from each mask's training labels equals the code's \minesPrior{} on every mask, and belief propagation rerun with it reproduces both of its scores. Given the generator's form, the samplers' reference scores therefore do not depend on knowing its values. The likelihood matcher was not rerun with estimated values. Recall of the training patterns without the generator's values is measured by the lookup at the start of this appendix.

\begin{table}[ht]
\centering
\caption{The values each sampler takes from the generator's code and as estimated from the training split, with the samplers' scores under each. For PATTERN's host, the middle column uses the number of blocks the estimator chooses for most training hosts, the fewest it allows, and the last column fixes the code's number. Most hosts fitted with \estPatternHostBlocks{} blocks leave one of them empty, so the middle column's block sizes are close to the sizes of whole hosts. The sizes the sampler's prior allows span both the fitted host blocks and the patterns. The threshold was already the training split's share of pattern nodes. Scores are pooled class-weighted accuracy with 95\% graph-bootstrap intervals, and differences from the run with the code's values are paired by graph. Minesweeper's estimated mine rate equals the code's, so the table has no rows for it.}
\label{tab:estimated}
\small
\resizebox{\linewidth}{!}{\begin{tabular}{llccc}
\toprule
 & & & \multicolumn{2}{c}{Estimated from the training split} \\
\cmidrule(lr){4-5}
Benchmark & Value & From the code & As chosen & At five host blocks \\
\midrule
PATTERN & Host blocks & \genPatternHostBlocks{} & \estPatternHostBlocks{} & \genPatternHostBlocks{} \\
 & Edge probability in a host block & \genPatternPIn{} & \estPatternPIn{} & \estPatternFiveBlocksPIn{} \\
 & Edge probability across host blocks & \genPatternPOut{} & \estPatternPOut{} & \estPatternFiveBlocksPOut{} \\
 & Host block size & \genPatternBlockMin{} to \genPatternBlockMax{} & \estPatternBlockMin{} to \estPatternBlockMax{} & \estPatternFiveBlocksBlockMin{} to \estPatternFiveBlocksBlockMax{} \\
 & Edge probability, pattern to host & \genPatternPHost{} & \estPatternPHost{} & \estPatternPHost{} \\
 & Edge probability in the pattern & \genPatternPInternal{} & \estPatternPInternal{} & \estPatternPInternal{} \\
 & Pattern size & \genPatternSizeMin{} to \genPatternSizeMax{} & \estPatternSizeMin{} to \estPatternSizeMax{} & \estPatternSizeMin{} to \estPatternSizeMax{} \\
 & Sizes the sampler's prior allows & \blindSizeMin{} to \blindSizeMax{} & \estPatternPriorMin{} to \estPatternPriorMax{} & \estPatternPriorMin{} to \estPatternPriorMax{} \\
 & Threshold (pattern share of nodes) & \blindThresholdRound{} & \estPatternClassShareRound{} & \estPatternClassShareRound{} \\
\addlinespace
 & Sampler, official test split & \realDensityPooled{} (\realDensityLo{} to \realDensityHi{}) & \estRealDensityPooled{} (\estRealDensityLo{} to \estRealDensityHi{}) & \estFiveBlocksRealDensityPooled{} (\estFiveBlocksRealDensityLo{} to \estFiveBlocksRealDensityHi{}) \\
 & Sampler, fresh patterns & \freshDensityPooled{} (\freshDensityLo{} to \freshDensityHi{}) & \estFreshDensityPooled{} (\estFreshDensityLo{} to \estFreshDensityHi{}) & \estFiveBlocksFreshDensityPooled{} (\estFiveBlocksFreshDensityLo{} to \estFiveBlocksFreshDensityHi{}) \\
 & Minus the code's values, official, paired &  & \estRealDensityMinusKnownTwo{} (\estRealDensityMinusKnownLoTwo{} to \estRealDensityMinusKnownHiTwo{}) & \estFiveBlocksRealDensityMinusKnownTwo{} (\estFiveBlocksRealDensityMinusKnownLoTwo{} to \estFiveBlocksRealDensityMinusKnownHiTwo{}) \\
 & Minus the code's values, fresh, paired &  & \estFreshDensityMinusKnownTwo{} (\estFreshDensityMinusKnownLoTwo{} to \estFreshDensityMinusKnownHiTwo{}) & \estFiveBlocksFreshDensityMinusKnownTwo{} (\estFiveBlocksFreshDensityMinusKnownLoTwo{} to \estFiveBlocksFreshDensityMinusKnownHiTwo{}) \\
\midrule
CLUSTER & Blocks & \genClusterBlocks{} & \estClusterBlocks{} &  \\
 & Edge probability in a block & \genClusterPIn{} & \estClusterPIn{} &  \\
 & Edge probability across blocks & \genClusterPOut{} & \estClusterPOut{} &  \\
 & Block size & \genClusterBlockMin{} to \genClusterBlockMax{} & \estClusterBlockMin{} to \estClusterBlockMax{} &  \\
\addlinespace
 & Sampler, test split & \clusterCeilingPooled{} (\clusterCeilingLo{} to \clusterCeilingHi{}) & \estClusterPooled{} (\estClusterLo{} to \estClusterHi{}) &  \\
 & Minus the code's values, paired &  & \estClusterMinusCeilingTwo{} (\estClusterMinusCeilingLoTwo{} to \estClusterMinusCeilingHiTwo{}) &  \\
\bottomrule
\end{tabular}
}
\end{table}

\section{What each method costs}
\label{app:cost}

The trained networks label PATTERN's test split in seconds on a GPU and in minutes on a CPU. The likelihood matcher, the pattern-blind predictor, and the lookup of the training patterns each take hours of CPU time, and so do the samplers of CLUSTER and Minesweeper; belief propagation on Minesweeper takes minutes (Table~\ref{tab:cost}). We did not time GatedGCN-PE, GRASS or PPGT. The methods built from the generator are expensive at inference; they serve as reference scores.

\begin{table}[ht]
\centering
\caption{What each method needs at test time, and how long it took: above the rule on PATTERN's official test split of \recTestGraphs{} graphs, below it on CLUSTER's test split of \clusterGraphs{} graphs and on Minesweeper's board. GPU times are on one NVIDIA T4, and CPU times on machines with four CPU threads. The lookup's time is the sum of its per-graph times, its time on one process. Every other time is the wall-clock time with the processes listed running in parallel, added over the runs that divided a split, so its CPU time is about this time multiplied by the number of processes. The time per graph is the split's time divided by its number of graphs. The networks' times run from loading the graphs to their predictions and are medians over repeated passes on the GPU; GRIT's CPU time was measured on \costGritCpuGraphs{} graphs and scaled to the split. CKGConv's two recipes share the architecture and took the same time on the GPU; its CPU time is the released recipe's. The matcher's time includes the sampler that builds its shortlist. For two-hop logistic regression, the test times cover the node statistics, and the training time covers the training split's statistics and the fit. Minesweeper's board is scored on \minesSplits{} masks, so it has no time per graph; the strip sampler ran for a time fixed in advance (Appendix~\ref{app:mines}), and the time of belief propagation covers both of its runs.}
\label{tab:cost}
\footnotesize
\setlength{\tabcolsep}{4pt}
\begin{tabular}{>{\raggedright\arraybackslash}p{0.16\linewidth}>{\raggedright\arraybackslash}p{0.17\linewidth}>{\raggedright\arraybackslash}p{0.17\linewidth}rr>{\raggedright\arraybackslash}p{0.14\linewidth}}
\toprule
Method & Needs at test time & Hardware & Per graph & Test split & Training \\
\midrule
CKGConv & a trained model & one T4 GPU, batches of \costGpuBatch{} & \costCkgGpuGraph{} ms & \costCkgGpuSplit{} s & \costCkgTrainLo{} to \costCkgTrainHi{} h per seed on a T4 \\
 &  & CPU, \costCpuThreads{} threads & \costCkgCpuGraph{} ms & \costCkgCpuSplit{} s &  \\
\addlinespace
GRIT & a trained model & one T4 GPU, batches of \costGpuBatch{} & \costGritGpuGraph{} ms & \costGritGpuSplit{} s & \costGritTrain{} h per seed on a T4 \\
 &  & CPU, \costCpuThreads{} threads & \costGritCpuGraph{} ms & \costGritCpuSplit{} s &  \\
\addlinespace
Two-hop logistic regression & a trained model & CPU, \costLadderProcs{} processes & \costLadderGraph{} ms & \costLadderSplit{} s & \costLadderTrain{} min \\
\addlinespace
Lookup of the training patterns & the training patterns & CPU, one process & \costLookupGraph{} s & \costLookupSplit{} h & none \\
\addlinespace
Likelihood matcher & the training patterns and the generator's values & CPU, \matcherOfficialProcs{} processes & \costMatcherGraph{} s & \costMatcherSplit{} h & none \\
\addlinespace
Pattern-blind predictor & the generator's values & CPU, \costDensityProcs{} processes & \costDensityGraph{} s & \costDensitySplit{} h & none \\
\midrule
CLUSTER sampler & the generator's values & CPU, \costClusterProcs{} processes & \costClusterGraph{} s & \costClusterSplit{} h & none \\
\addlinespace
Minesweeper strip sampler & the generator's values & CPU, \costMinesExactProcs{} processes &  & \costMinesExactSplit{} h & none \\
\addlinespace
Minesweeper belief propagation & the generator's values & CPU, one process &  & \costMinesBpSplit{} min & none \\
\bottomrule
\end{tabular}

\end{table}

\section{Expectations and outcomes}
\label{app:registrations}

We wrote down what we expected of each experiment and check, and what would refute it, in a dated log before running it, with the exceptions below; Table~\ref{tab:registrations} gives both, with what happened, in the order of the paper. The expectations for the two lookups were written after the largest-match rule's official score and a first run of the lookup on a sample of test graphs were known. The ranges for Minesweeper from the features alone were set after an early run of belief propagation had been seen. We wrote no expectation for the size-matched values of GatedGCN-PE, GRASS and PPGT, or for the class recalls of Appendix~\ref{app:fresh}. The table marks the rows these exceptions touch.

Most expectations held, and the table reports each miss. Two misses changed what the paper reports. The prior on block sizes moved the sampler's score by more than we had allowed, so the paper reports the sampler with the prior as the pattern-blind predictor, as a rule written before the run required. At the new-pattern models' checkpoint epochs, the expected gap to the reuse models appeared in one seed of three, so Section~\ref{sec:reuse} treats that comparison as confounded with training length (Appendix~\ref{app:traininglength}). One wording fixed in advance also changed: the familiar set's small advantage over the size-matched set is not detected, so Section~\ref{sec:limitations} states only that no effect is detected.

{\setlength{\LTcapwidth}{\linewidth}
\renewcommand{\arraystretch}{1.0}
\setlength{\tabcolsep}{4pt}
\begin{longtable}{>{\footnotesize\raggedright\arraybackslash}p{0.15\linewidth}>{\footnotesize\raggedright\arraybackslash}p{0.3\linewidth}>{\footnotesize\raggedright\arraybackslash}p{0.18\linewidth}>{\footnotesize\raggedright\arraybackslash}p{0.29\linewidth}}
\caption{What we expected from each experiment and check, in the order of the paper, and what happened. Batch mean and pooled are the two ways of computing class-weighted accuracy (Appendix~\ref{app:metric}); where a prediction named the batch mean, its row says so. AUC is in percent. $^{*}$An exception listed in the text of Appendix~\ref{app:registrations}.}
\label{tab:registrations}\\
\toprule
Experiment or check & What was predicted & What would have refuted it & What happened \\
\midrule
\endfirsthead
\multicolumn{4}{l}{\footnotesize Table~\ref{tab:registrations}, continued} \\
\toprule
Experiment or check & What was predicted & What would have refuted it & What happened \\
\midrule
\endhead
\bottomrule
\endfoot
\bottomrule
\endlastfoot
\multicolumn{4}{l}{\footnotesize\itshape Section~\ref{sec:recall}: PATTERN's test split} \\*
\addlinespace
Recurrence in the released files & about \regPilotPatterns{} distinct training patterns (\regPilotPatternsLo{} to \regPilotPatternsHi{}), each in about \regPilotPerPattern{} training graphs; at least \regPilotRecur{}\% of validation and test graphs carry one & fewer than \regPilotKillRecur{}\% of test graphs carrying a training pattern & held: exactly \recDistinctPatterns{} patterns, each in \recGraphsPerPattern{} training graphs; \recTestRecur{}\% of validation and test graphs carry one \\
\addlinespace
The largest-match rule on \fastGraphs{} random test graphs, then on all \realGraphs{} & \regPilotMatcherLo{} to \regPilotMatcherHi{} (batch mean); for the whole split, written after the sample, \regLargestLo{} to \regLargestHi{} & a score below \regPilotKillMatcher{} & \fastMatcherBatch{} on the sample (pooled \fastMatcherPooled{}) and \convLargestBatch{} on the whole split (pooled \realLargestPooled{}), both under the first range; the whole split's score lies inside the second; not refuted \\
\addlinespace
The likelihood matcher on all test graphs & \regMatcherPoint{} (\regMatcherLo{} to \regMatcherHi{}), batch mean; exact recovery at least \regExactAll{}\% (largest match about \regExactLargest{}\%), and at least \regExactSmall{}\% for patterns under \matcherBig{} nodes & the matcher below \regKillMatcher{} & held on the scores: \realMatcherBatch{} (pooled \realMatcherPooled{}); exact recovery \realMatcherExactPct{}\% (largest match \realLargestExactPct{}\%); missed for small patterns, at \realMatcherExactPctSmall{}\% \\
\addlinespace
The lookup and the lookup with a degree shortlist on the official test split$^{*}$ & the lookup \regLookupLo{} to \regLookupHi{}; the lookup with a degree shortlist within \regLookupDegreeWithin{} points of the largest-match rule's \regLookupRule{}, paired. On graphs with a pattern of \matcherBig{} nodes or more the lookups make the rule's predictions by construction, and as a check of the code, both should make them wherever such a pattern is found & none; a difference in the check would have stopped the analysis as a code error & held: \lookupRealPooled{}; the lookup with a degree shortlist \lookupDegreeRealMinusLargestTwo{} points from the rule (\lookupDegreeRealMinusLargestLoTwo{} to \lookupDegreeRealMinusLargestHiTwo{}); the same predictions on every graph where a large pattern is found, on the official test split and the familiar set \\
\addlinespace
\multicolumn{4}{l}{\footnotesize\itshape Section~\ref{sec:measured}: the pattern-blind predictor and patterns no model has seen} \\*
\addlinespace
The pattern-blind predictor on the official test split & \regDensityPoint{} (\regDensityLo{} to \regDensityHi{}), batch mean. Separately, that the best model published up to 2023, GRIT at \regDensityClauseAnchor{}, lies within \regDensityClauseWithin{} point of this score, so that the score lies from \regDensityClauseLo{} to \regDensityClauseHi{} & for the second statement only: a score below \regKillDensity{} & held: \firstBlindRealBatch{} (pooled \firstBlindRealPooled{}), inside both ranges; the sampler with the size prior, which the paper reports, scores \convDensityBatch{} (pooled \realDensityPooled{}) \\
\addlinespace
The three predictors on the \freshGraphs{} graphs of the fresh set & the matcher \regFreshMatcherLo{} to \regFreshMatcherHi{}, below the pattern-blind predictor; the largest-match rule \regFreshLargestLo{} to \regFreshLargestHi{}; the pattern-blind predictor within \regFreshDensityWithin{} points of its test-split value & the matcher within \regFreshKill{} points of its test-split score & held: \freshMatcherPooled{}, \freshLargestPooled{}, and, for the sampler without the size prior, \firstBlindFreshPooled{}, \firstBlindRealMinusFresh{} below its test-split value \\
\addlinespace
The released patterns against the law of fresh patterns & sizes consistent with the uniform law ($p$ above \regTwoBOneP{}), with a mean within \regTwoBOneMeanWithin{} of \regTwoBOneMean{}; internal edge density within \regTwoBTwoWithin{} of \regTwoBTwoDensity{}, features consistent with uniform ($p$ above \regTwoBTwoP{}), and pattern-to-host edges at a rate within \regTwoBTwoHostWithin{} of \regTwoBTwoHost{} & none; a difference would have been reported with its possible effect on the fall & held: $p = \releasedSizesChiP{}$ and mean \releasedSizesMean{}; density \releasedDensity{}, features $p = \releasedFeatureChiP{}$, pattern-to-host rate \releasedHostRate{} \\
\addlinespace
The matcher and the pattern-blind predictor on the familiar set and on the validation split & each within \regTwoCOneWithin{} point of its official score (\regTwoCOneValue{} and \regTwoCTwoValue{}) on both sets & none; a matcher far below its official score would have pointed to our regenerated hosts, or to settings fitted to the test split & held: the matcher \familiarMatcherPooled{} and \valMatcherPooled{}, the sampler without the size prior \firstBlindFamiliarPooled{} and \firstBlindValPooled{} \\
\addlinespace
The two lookups on the familiar and fresh sets$^{*}$ & both \regLookupFreshLo{} to \regLookupFreshHi{} on the fresh set, and within \regLookupFamiliarWithin{} point of their official scores on the familiar set & none & held: \lookupFreshPooled{} and \lookupDegreeFreshPooled{} on the fresh set; on the familiar set \lookupFamiliarMinusRealTwo{} and \lookupDegreeFamiliarMinusRealTwo{} points above the official scores \\
\addlinespace
CKGConv with its released recipe, two seeds & reproduction within \regCkgReproduce{} point of the published \pubPatternCKGConv{}; on the fresh set between \regCkgBandLo{} and \regCkgBandHi{}, around the pattern-blind predictor. Required before the fresh scores were read: the round trip within \regCkgRoundtrip{}, the reproduction within \regCkgGate{} points of the published score, and the familiar set within \regCkgFamiliar{} of the official score & a fresh-pattern score within \regCkgKill{} point of the model's own official score & checks passed: \gpuOriginalPooled{} and \gpuSeedBOriginalPooled{} on the official test split, the round trip identical, the familiar set \gpuFamiliarMinusOriginal{} and \gpuSeedBFamiliarMinusOriginal{} from it. The fresh scores, \gpuFreshPooled{} and \gpuSeedBFreshPooled{}, fall below the band, beyond the prediction; not refuted \\
\addlinespace
CKGConv without its threshold, two seeds & on the fresh set, at the single best threshold, below the pattern-blind predictor's \regTwoAOne{}; a ROC AUC on the fresh set at least \regTwoATwo{} point below that on the familiar set; both in both seeds & none; had the first failed, the fall below the pattern-blind predictor would have been reported as an effect of the threshold & held: \ckgFreshOracle{} and \ckgSeedBFreshOracle{}; AUC \ckgFreshAuc{} and \ckgSeedBFreshAuc{} against \ckgFamiliarAuc{} and \ckgSeedBFamiliarAuc{} (Table~\ref{tab:auc}) \\
\addlinespace
CKGConv trained without \regTwoDHeld{} of the \recDistinctPatterns{} patterns, two seeds & on official test graphs whose pattern was held out, at least \regTwoDOne{} points below those whose pattern was kept, and within \regTwoDTwo{} point of the pattern-blind predictor or below it; on the kept graphs within \regTwoDThree{} point of CKGConv trained on all the patterns & the held-out graphs within \regTwoDKill{} point of the kept ones in both seeds & held: \heldH{} and \heldSeedBH{} on the held-out graphs against \heldKept{} and \heldSeedBKept{} on the kept ones, and \firstBlindHeldBelow{} and \firstBlindHeldSeedBBelow{} points below the sampler without the size prior (\heldBelowBlind{} and \heldSeedBBelowBlind{} below the pattern-blind predictor); on the kept graphs \heldKeptMinusFull{} and \heldSeedBKeptMinusFull{} points above CKGConv trained on all the patterns \\
\addlinespace
The held-out graphs against CKGConv trained on all the patterns & the paired difference below zero, its interval included, in both seeds & none & held: \heldHMinusFull{} (\heldHMinusFullLo{} to \heldHMinusFullHi{}) and \heldSeedBHMinusFull{} (\heldSeedBHMinusFullLo{} to \heldSeedBHMinusFullHi{}) \\
\addlinespace
CKGConv trained without a second draw of \regTwoDHeld{} patterns, two seeds & as for the first draw: on official test graphs whose pattern was held out, at least \regTwoDOne{} points below those whose pattern was kept, and within \regTwoDTwo{} point of the sampler without the size prior or below it; on the kept graphs within \regTwoDThree{} point of CKGConv trained on all the patterns & the held-out graphs within \regTwoDKill{} point of the kept ones in both seeds & held: \heldSecondH{} and \heldSecondSeedBH{} on the \heldSecondGraphs{} held-out graphs against \heldSecondKept{} and \heldSecondSeedBKept{} on the kept ones, and \heldSecondBelowCue{} and \heldSecondSeedBBelowCue{} points below the sampler without the size prior (\heldSecondBelowDensity{} and \heldSecondSeedBBelowDensity{} below the pattern-blind predictor); on the kept graphs \heldSecondKeptMinusFull{} and \heldSecondSeedBKeptMinusFull{} points above CKGConv trained on all the patterns. The held-out graphs lay at least \regTwoDOne{} points below the kept ones in both seeds of both draws \\
\addlinespace
The sampler with the prior on block sizes and \blindSweepsRatioWord{} times the sweeps & within \regBOneWithin{} points of the sampler without them, at \regBOneAnchorReal{} on the official test split and \regBOneAnchorFresh{} on the fresh set & none; a larger move would make the sampler with the prior the paper's estimate, with every comparison recomputed & missed: \blindMoveOfficialThree{} and \blindMoveFreshThree{} points above, so the paper reports the sampler with the prior as the pattern-blind predictor \\
\addlinespace
The sampler without the size prior, started at the planted pattern & within \regBThreeWithin{} points of \regBThreeAnchor{} & none; a score more than that above would have meant that the chains do not mix and that the pattern-blind level may lie well above the sampler's score & held: \blindTrueStartPooled{} \\
\addlinespace
The pattern-blind predictor's ROC AUC on the fresh set & above CKGConv's \regBFourSeedA{} and \regBFourSeedB{} on the same graphs & none & held: \firstBlindAucFresh{} for the sampler without the size prior, and \blindAucFresh{} for the pattern-blind predictor (Table~\ref{tab:auc}) \\
\addlinespace
Calibration of the pattern-blind predictor on the official test split and the familiar and fresh sets & on each set, the class-weighted accuracy that the posterior expects (Equation~\eqref{eq:bayes}) within \regCalibWithin{} point of the realized score, and an expected calibration error over ten bins of at most \regCalibEce{} & none; a posterior more confident than its realized accuracy would not be the generator's posterior, and the paper would say so & held on every set: expected minus realized \calibRealExpectedMinusRealizedTwo{}, \calibFamiliarExpectedMinusRealizedTwo{} and \calibFreshExpectedMinusRealizedTwo{}; calibration error \calibRealEce{}, \calibFamiliarEce{} and \calibFreshEce{} \\
\addlinespace
The structural ladder: logistic regression and boosted trees on statistics computed from the edges alone & logistic regression on degree alone \regLadderDegreeLo{} to \regLadderDegreeHi{} on the official test split; on the fresh set, the best predictor at most \regLadderAboveSampler{} point above the pattern-blind predictor, and boosted trees on two-hop statistics at least \regLadderBoostedMin{}; official minus fresh within \regLadderShift{} point for every predictor; boosted trees retrained without the held-out patterns, the test graphs that carry them minus the others, within \regLadderHeldWithin{} point. If logistic regression on degree alone (official test split) or the best predictor without node features (fresh set) reached \regLadderRemedy{} or more, detection was to be read as resting mostly on degree, and a change to the generator proposed & none; a best predictor whose paired interval lay above the pattern-blind predictor would have replaced it as the reference score, and a larger shift or held-out difference would have meant that node statistics carry pattern identity & held: \ladderLogisticDegreeRealPooled{}; the best on the fresh set \ladderBestFreshMinusDensityTwo{} points from the pattern-blind predictor (\ladderBestFreshMinusDensityLoTwo{} to \ladderBestFreshMinusDensityHiTwo{}); boosted trees \ladderBoostedTwoHopFreshPooled{}; the largest shift \ladderLargestRealMinusFreshTwo{}; held-out minus kept \ladderHeldBoostedTwoHopHMinusKeptTwo{}. The best predictor without node features, two-hop logistic regression, reaches \ladderBestFreshPooled{} on the fresh set (degree alone \ladderLogisticDegreeFreshPooled{}), so Section~\ref{sec:practices} proposes a change to the generator \\
\addlinespace
The same boosted trees with node features (exploratory) & official minus fresh at least \regLadderFeaturesGap{} point & none & missed: \ladderBoostedTwoHopFeaturesRealMinusFreshTwo{} \\
\addlinespace
The pattern-blind predictor with every value estimated from the training split & the pattern's internal density and rate of edges to the host within \regEstRateWithin{} of \regEstRate{}; at \regEstBlocks{} host blocks, the host's edge probabilities within \regEstPInWithin{} of \regEstPIn{} inside blocks and within \regEstPOutWithin{} of \regEstPOut{} across them; \regEstBlocks{} blocks chosen; the sampler within \regEstSamplerWithin{} point of the run with the code's values on the official test split and the fresh set, paired & the sampler more than \regEstSamplerKill{} point below the run with the code's values; the paper would then have said that the predictor depends on the exact values & the rates held, at \estPatternPInternal{} and \estPatternPHost{}. The host's values missed, as checks on synthetic hosts, made after the prediction and before the run, had shown they would: the estimator chooses \estPatternHostBlocks{} blocks on \estPatternBlocksTwoGraphs{} of the \estPatternGraphs{} training hosts, and at \regEstBlocks{} blocks it fits \estPatternFiveBlocksPIn{} inside and \estPatternFiveBlocksPOut{} across. The score held: \estRealDensityMinusKnownTwo{} and \estFreshDensityMinusKnownTwo{} points from the run with the code's values \\
\addlinespace
GRIT on the fresh set, two seeds & official score within \hvProtocolGritReproductionTolerance{} points of the published score, and the familiar set within \hvProtocolGritFamiliarTolerance{} point of it; fresh accuracy at least \hvProtocolGritAccuracyGap{} point below official, and its paired difference from the pattern-blind predictor wholly below zero; familiar minus fresh AUC at least \hvProtocolGritAucGap{} point & fresh accuracy within \hvProtocolGritAccuracyGap{} point of official in both seeds would have contradicted the accuracy-drop expectation; each AUC-gap check is reported separately & validity checks passed. Fresh accuracy \hvCompletedGritSeedAOfficialMinusFresh{} and \hvCompletedGritSeedBOfficialMinusFresh{} points below official; paired intervals from the pattern-blind predictor below zero in both seeds (Table~\ref{tab:gritcompleted}). AUC expectation mixed: gaps \hvCompletedGritSeedAFamiliarMinusFreshAuc{} and \hvCompletedGritSeedBFamiliarMinusFreshAuc{}, failing in the first seed and holding in the second \\
\addlinespace
CKGConv and GRIT on the size-matched set, two seeds each & familiar minus size-matched, paired by graph, at least \regSizeCkg{} points for CKGConv and at least \regSizeGrit{} for GRIT in both seeds; the pattern-blind predictor on the size-matched set within \regSizeDensity{} point of its score on the fresh set & none; a difference under \regSizeKill{} point for CKGConv in both seeds would have meant that pattern size explains much of the fall & held: CKGConv \sizematchCkgSeedAFamiliarMinusMatchedTwo{} and \sizematchCkgSeedBFamiliarMinusMatchedTwo{}, GRIT \sizematchGritSeedAFamiliarMinusMatchedTwo{} and \sizematchGritSeedBFamiliarMinusMatchedTwo{}; the pattern-blind predictor \sizematchDensityMatched{} against \freshDensityPooled{} \\
\addlinespace
GatedGCN-PE on the fresh set, seeds \gatedSeedASeed{} and \gatedSeedBSeed{} & in each seed: the official batch mean within \regGatedWithin{} point of the published \pubPatternGatedGCNPETwo{}; familiar minus fresh at most \regGatedDrop{} point; fresh minus the pattern-blind predictor, paired, between \regGatedSamplerLo{} and \regGatedSamplerHi{} & none; the second in both seeds was to be read as no sizable dependence on the training patterns in a model that scores as the pattern-blind predictor does & held in both seeds: official batch means \gatedSeedAOfficialBatchTwo{} and \gatedSeedBOfficialBatchTwo{}; familiar minus fresh and fresh minus the pattern-blind predictor, with its paired interval, in Table~\ref{tab:fivemodels_seeds} \\
\addlinespace
GRASS on the fresh set, two seeds & in each seed: the fresh score at least \regGrassFall{} points below the official score; fresh minus the sampler without the size prior, paired, no more than \regGrassNear{} point above zero; familiar minus fresh AUC at least \regGrassAuc{} point. Required before the fresh scores were read: the round trip within \regGrassRoundtrip{}, GRASS's own mean over \fiveGrassPasses{} passes on the official test split within \regGrassGate{} points of the published \regGrassPublished{}, and the familiar set within \regGrassFamiliar{} of the official score & in either seed, a fresh-pattern score within \regGrassKill{} point of the model's own official score & checks passed: GRASS's own means \grassSeedAGate{} and \grassSeedBGate{}, the round trip \grassSeedARoundtrip{} and \grassSeedBRoundtrip{} points with identical inputs, the familiar set \grassSeedAFamiliarMinusOfficial{} and \grassSeedBFamiliarMinusOfficial{} from the official score. Held in both seeds: the fresh score \grassSeedAOfficialMinusFresh{} and \grassSeedBOfficialMinusFresh{} points below the official one; fresh minus the sampler without the size prior \grassSeedAFreshMinusCue{} and \grassSeedBFreshMinusCue{} (Table~\ref{tab:fivemodels_seeds} gives the difference from the pattern-blind predictor); AUC gaps \grassSeedAAucGap{} and \grassSeedBAucGap{} \\
\addlinespace
PPGT on the fresh set, two seeds & in each seed: the fresh score at least \regPpgtFall{} points below the official score; fresh minus the pattern-blind predictor, paired, no more than \regPpgtNear{} point above zero; familiar minus fresh AUC at least \regPpgtAuc{} point. Required before the fresh scores were read: the round trip within \regPpgtRoundtrip{}, the official score within \regPpgtGate{} points of the published \regPpgtPublished{}, and the familiar set within \regPpgtFamiliar{} of the official score & in either seed, a fresh-pattern score within \regPpgtKill{} point of the model's own official score & checks passed: the round trip identical, the familiar set \ppgtSeedAFamiliarMinusOfficial{} and \ppgtSeedBFamiliarMinusOfficial{} from the official score. Held in both seeds: the fresh score \ppgtSeedAOfficialMinusFresh{} and \ppgtSeedBOfficialMinusFresh{} points below the official one; AUC gaps \ppgtSeedAAucGap{} and \ppgtSeedBAucGap{}; official scores and fresh minus the pattern-blind predictor in Table~\ref{tab:fivemodels_seeds} \\
\addlinespace
The five models, on each model's first two seeds & GRASS and PPGT each lose more from the familiar to the fresh set than GatedGCN-PE and than GRIT; over the five models, a Spearman correlation of at least \fiveSpearmanAtLeast{} between the drop and the excess, the official score minus the pattern-blind predictor's \fiveSampler{} & none; had the first failed, the paper would not have claimed that dependence on the training patterns grows with a model's score & held: drops \fiveGatedFirstTwoDropTwo{} for GatedGCN-PE, \fiveGritFirstTwoDropTwo{} for GRIT, \fiveCkgFirstTwoDropTwo{} for CKGConv, \fiveGrassFirstTwoDropTwo{} for GRASS and \fivePpgtFirstTwoDropTwo{} for PPGT; Spearman correlation \fiveSpearman{}, from the first two seeds of each model in Table~\ref{tab:fivemodels_seeds} \\
\addlinespace
Added seeds: GRIT's seed \fiveGritSeedCNumber{}, GatedGCN-PE's seeds \fiveGatedSeedCNumber{} and \fiveGatedSeedDNumber{}, CKGConv's seed \fiveCkgSeedCNumber{} with the released patterns and with new patterns, GRASS's seed \fiveGrassSeedCNumber{}, and PPGT's seeds \fivePpgtSeedCNumber{} and \fivePpgtSeedDNumber{} & each new seed's familiar minus fresh within \regToneWithin{} point of the mean of the model's first two seeds; the model's expectations above unchanged for its new seeds$^{*}$ & none; a model that failed the first would have been named, and its drop reported with its spread & by model in the six rows below; Table~\ref{tab:fivemodels_seeds} gives every seed's accuracies and Table~\ref{tab:auc} its AUC \\
\addlinespace
\quad GRIT, seed \fiveGritSeedCNumber{} & as above, with GRIT's expectations, including the size-matched one & & held: the drop lies within \regToneWithin{} point of the first two seeds' mean, and GRIT's expectations held, with an AUC gap of \gritSeedCAucGap{} and familiar minus size-matched \sizematchGritSeedCFamiliarMinusMatchedTwo{} (paired interval \sizematchGritSeedCFamiliarMinusMatchedLoTwo{} to \sizematchGritSeedCFamiliarMinusMatchedHiTwo{}) against the required \regSizeGrit{} \\
\addlinespace
\quad GatedGCN-PE, seeds \fiveGatedSeedCNumber{} and \fiveGatedSeedDNumber{} & as above, with GatedGCN-PE's expectations & & held in both seeds: the drops lie within \regToneWithin{} point of the first two seeds' mean, and GatedGCN-PE's expectations held, with official batch means \gatedSeedCOfficialBatchTwo{} and \gatedSeedDOfficialBatchTwo{} \\
\addlinespace
\quad CKGConv, released recipe, seed \fiveCkgSeedCNumber{} & as above, with CKGConv's expectations, including the size-matched one & & the drop lies within \regToneWithin{} point of the first two seeds' mean. CKGConv's expectations held except the fresh-set band, which this seed misses as the first two did; familiar minus size-matched \ckgSeedCSizeMatchedDropTwo{} (paired interval \ckgSeedCSizeMatchedDropLoTwo{} to \ckgSeedCSizeMatchedDropHiTwo{}) against the required \regSizeCkg{} \\
\addlinespace
\quad CKGConv, new training and validation patterns, seed \fiveCkgSeedCNumber{} & as above, with the expectations for this recipe & & held: the drop lies within \regToneWithin{} point of the first two seeds' mean, the gain over the released recipe has its paired interval above zero, the difference from the pattern-blind predictor lies inside the proximity range, and the official score within its tolerance (Table~\ref{tab:ckgfreshcompleted}) \\
\addlinespace
\quad GRASS, seed \fiveGrassSeedCNumber{} & as above, with GRASS's checks and expectations & & checks passed and expectations held: the drop lies within \regToneWithin{} point of the first two seeds' mean; GRASS's own mean \grassSeedCGate{}, the round trip \grassSeedCRoundtrip{} points with identical inputs, and fresh minus the sampler without the size prior \grassSeedCFreshMinusCue{} \\
\addlinespace
\quad PPGT, seeds \fivePpgtSeedCNumber{} and \fivePpgtSeedDNumber{} & as above, with PPGT's expectations & & the first missed: the new seeds lost \ppgtSeedCToneOverTwo{} and \ppgtSeedDToneOverTwo{} points more than the first two seeds' mean, so PPGT's drop is reported with its range (Section~\ref{sec:fresh}). PPGT's own expectations held in both new seeds \\
\addlinespace
The five models, on every seed & from each model's mean over all its seeds, the same conclusions as on its first two seeds for both expectations above & none & held, with the mean drops of Table~\ref{tab:fivemodels} and a Spearman correlation of \fiveSpearmanAllSeeds{} \\
\addlinespace
CKGConv with new training and validation patterns, two seeds & paired improvement over the original training recipe on the same fresh test graphs, with its interval wholly above zero; fresh minus the pattern-blind predictor, as a point value, between \hvProtocolCkgProximityLower{} and \hvProtocolCkgProximityUpper{} points; official score within \hvProtocolCkgOfficialTolerance{} point of the pattern-blind predictor's official score. For the fresh-set difference we expected this proximity, with the large deficit of the next row as the competing outcome & an improvement interval not wholly above zero would have failed the improvement check; a point difference outside the specified range would have failed the proximity check & held in both seeds (Table~\ref{tab:ckgfreshcompleted}): the gains' paired intervals lie above zero, the differences from the pattern-blind predictor inside the specified range, and the official scores within the tolerance. Proximity is a criterion on the point difference alone \\
\addlinespace
The competing outcome: a large remaining deficit after training on new patterns & a fresh score more than \hvProtocolCkgLargeDeficit{} points below the pattern-blind predictor in both seeds, the alternative to proximity & a seed whose fresh score is not that far below & not met in either seed (Table~\ref{tab:ckgfreshcompleted}) \\
\addlinespace
CKGConv trained on our graphs that reuse a library of \reuseLibraryPatterns{} new patterns, three libraries with one seed each & in each run: library set minus fresh set at least \regReuseGap{} points; on the fresh set, below CKGConv trained with new patterns and the same seed, with the paired interval wholly below zero; on the official test split, at least \regReuseOfficialBelow{} points below CKGConv trained on the released patterns and the same seed, paired, and at or below the pattern-blind predictor's \fiveSampler{}. The first two in at least two runs were to be read as reuse alone producing the gap and the loss on new patterns, the last two as the official advantage coming from training on the released patterns & library minus fresh under \regReuseKill{} point in all three runs, which would have made the difference between our generator and the original a candidate cause & held in all three runs (Tables~\ref{tab:reusedetail} and~\ref{tab:libref}): library minus fresh above \regReuseGap{} points, and both comparisons with paired intervals below zero; the official score minus \fiveSampler{}: \reuseRunOneOfficialMinusSamplerTwo{}, \reuseRunTwoOfficialMinusSamplerTwo{} and \reuseRunThreeOfficialMinusSamplerTwo{}. The check below qualifies the reading of the second, and the paper states the first more narrowly: reuse, with the released rule for choosing the checkpoint, is enough to produce the gap \\
\addlinespace
The reuse control at the new-pattern models' epochs, read from the saved training histories & in each seed, the new-pattern model's official score at least \regMatchedGap{} point above the reuse model's logged official accuracy at the new-pattern model's epoch, and that accuracy below \fiveSampler{}; for seed 2, the new-pattern model's logged official accuracy at the reuse model's epoch, and its mean over the second half of training, within \regMatchedWithin{} point of its official score; the reuse model's highest logged official accuracy after warm-up below the new-pattern model's official score (descriptive) & the first in fewer than two seeds, to be read as the comparison with the new-pattern models being confounded with training length & the gap held in seed 0 only (\reuseRunOneNewPatternMinusMatchedTwo{}, \reuseRunTwoNewPatternMinusMatchedTwo{} and \reuseRunThreeNewPatternMinusMatchedTwo{}), so the comparison is confounded with training length (Section~\ref{sec:reuse}); the logged accuracy lies below \fiveSampler{} in all three seeds (Table~\ref{tab:reusedetail}); for seed 2, missed: \matchedNewPatternAtReuseEpochTwo{} at the reuse model's epoch and \matchedNewPatternLateMeanTwo{} over the second half, against \hvCompletedCkgSeedCOfficialAccuracy{}; the highest logged accuracy after warm-up, \reuseRunOneMaxAfterWarmupTwo{}, \reuseRunTwoMaxAfterWarmupTwo{} and \reuseRunThreeMaxAfterWarmupTwo{}, below the new-pattern model's score in seeds 0 and 2 (Table~\ref{tab:reusedetail}) \\
\addlinespace
A reference on the reuse control's library sets, and the reuse models on the size-matched set & on each library set, the pattern-blind predictor within \regLibrefWithin{} point of its score on the fresh set (\librefBlindFreshTwo{}); each reuse model at least \regLibrefAbove{} point above the predictor and above the new-pattern model of the same seed, paired, with the intervals above zero; the new-pattern model within \regLibrefBand{} point of the predictor, paired; and each reuse model's familiar minus size-matched interval, paired, containing zero in at least two runs. If the reuse models' two comparisons held, Section~\ref{sec:reuse} was to state the gap between seen and new patterns against the predictor and the new-pattern model on the same graphs; if the last held, Section~\ref{sec:limitations} was to say that the familiar set's small advantage follows the patterns' sizes & a library set on which the predictor lies further from its fresh score, to be read as graphs of different difficulty, with the gap then stated against the predictor alone; a reuse model less than \regLibrefAbove{} point above the predictor, to be read as its library advantage not established; intervals above zero in two runs, to be read as a fixed set of test patterns raising the score & held in every library and run (Tables~\ref{tab:reuse}, \ref{tab:reusedetail} and~\ref{tab:libref}): the predictor \librefOneBlindLibraryMinusFresh{}, \librefTwoBlindLibraryMinusFresh{} and \librefThreeBlindLibraryMinusFresh{} points above its fresh score; the reuse models \librefMinReuseMinusBlind{} to \librefMaxReuseMinusBlind{} points above the predictor and \librefMinReuseMinusNew{} to \librefMaxReuseMinusNew{} above the new-pattern models; the new-pattern models within about a tenth of a point of the predictor; every familiar minus size-matched interval contains zero. In the second library the predictor lies at the edge of its tolerance of \regLibrefWithin{}: inside it against the unrounded fresh score, the reference fixed before the runs, and just outside it against the printed \librefBlindFreshTwo{}. Section~\ref{sec:limitations} states only that no effect is detected, in place of the wording fixed for it \\
\addlinespace
\multicolumn{4}{l}{\footnotesize\itshape Section~\ref{sec:ceilings}: CLUSTER and Minesweeper} \\*
\addlinespace
CLUSTER's reference score on all \clusterGraphs{} test graphs, the sampler without the revealed nodes' factor & \regClusterPoint{} (\regClusterLo{} to \regClusterHi{}), batch mean; median chain agreement at least \regClusterAgreement{}; called saturated if the best model lies within \regClusterSaturationWord{} of its reported standard deviations of the estimate & none; reported either way, as a lower estimate if the chains disagree & in range at its lower edge and below the point prediction: \clusterFirstBatch{} (pooled \clusterFirstPooled{}), median agreement \clusterFirstAgreementMedian{}. The best model lies inside the estimate's interval, so we cannot tell whether headroom remains; it lies \clusterFirstGapToBestTwo{} points, about \clusterFirstGapInBestSds{} of its standard deviations, below the estimate, so the rule for calling CLUSTER saturated is not met. The rule compares the gap with the model's standard deviation over its seeds, a different yardstick from the estimate's interval \\
\addlinespace
CLUSTER with the revealed nodes' factor & \regAOneLo{} to \regAOneHi{} points above the sampler without it, with the same chains and sweeps & none & missed: \clusterFactorDiff{} (\clusterFactorDiffLo{} to \clusterFactorDiffHi{}) \\
\addlinespace
CLUSTER with the factor, \clusterChains{} chains and twice the sweeps & at least the score with the factor alone; its 95\% interval wholly above the best published \regAThreeBest{} & none; if the interval reached \regAThreeBest{}, whether headroom remains would be open & the first held (\clusterLongerDiff{} higher); the second missed, with the interval from \clusterCeilingLo{} to \clusterCeilingHi{} \\
\addlinespace
Minesweeper by belief propagation & from the features alone$^{*}$ \regMinesFeaturesLo{} to \regMinesFeaturesHi{}; with the training labels \regMinesLabelsLo{} to \regMinesLabelsHi{} & none; a features-only value under the best model would mean that belief propagation is not tight & outside both ranges: \minesFeaturesAuc{} from the features alone, under the best model and not converged; \minesLabelsAuc{} with the labels \\
\addlinespace
Minesweeper from the features alone, the strip sampler$^{*}$ & between \regMinesStripLo{} and \regMinesStripHi{} (point \regMinesStripPoint{}), above the best model; convergence expected to fail & a value at or under the \pubMinesRegisteredBest{} of \citet{hu2026graphtarif} & held: \minesExactAuc{}; the criteria were not met, and each of the \minesExactRanWord{} chains scores above that value \\
\addlinespace
CLUSTER's sampler with every value estimated from the training split & each estimate within \regEstClusterWithin{} of the code's value, and the sampler within \regEstClusterSampler{} point of its \regEstClusterAnchor{} with the code's values, paired & none & held: within and across blocks \estClusterPIn{} and \estClusterPOut{}, \estClusterBlocks{} blocks of \estClusterBlockMin{} to \estClusterBlockMax{} nodes; the sampler \estClusterMinusCeilingTwo{} points from the run with the code's values (\estClusterMinusCeilingLoTwo{} to \estClusterMinusCeilingHiTwo{}) \\
\addlinespace
Minesweeper's belief propagation with each mask's mine rate estimated from its training labels & AUC within \regEstMinesWithin{} points of the run with the code's rate, from the features alone and with the training labels & none & held: the estimated rate is the code's \estMinesRate{} on every mask, and the AUC changes by \estMinesFeaturesChange{} and \estMinesLabelsChange{} \\
\end{longtable}}

\end{document}